\documentclass[preprint,12pt,authoryear]{elsarticle}

\usepackage{amsmath,amssymb,mathtools}
\usepackage{graphicx}
\usepackage{booktabs}
\usepackage{multirow}
\usepackage{makecell}
\usepackage{array}
\usepackage{tabularx}
\usepackage{arydshln}
\usepackage{adjustbox}
\usepackage{subcaption}
\usepackage{algorithm}
\usepackage{algpseudocode}
\usepackage{siunitx}
\usepackage{xcolor}
\usepackage{pgfplots}
\usepackage{url}
\usepackage[hidelinks]{hyperref}
\usepackage{lineno}

\definecolor{darkblue}{RGB}{31,120,180}
\definecolor{indianred}{rgb}{0.8,0.36,0.36}
\definecolor{mediumseagreen}{rgb}{0.24,0.7,0.44}
\pgfplotsset{
  compat=1.18,
  umapAxis/.style={
    width=\linewidth,
    height=0.72\linewidth,
    scale only axis,
    xlabel={UMAP Dimension 1},
    ylabel={UMAP Dimension 2},
    label style={font=\scriptsize},
    ticklabel style={font=\scriptsize},
    legend style={font=\scriptsize, draw=none, fill=none},
    grid=major,
    grid style={line width=0.25pt, draw=gray, opacity=0.25},
    axis line style={line width=0.3pt},
    tick style={line width=0.3pt},
  },
}

\journal{Engineering Applications of Artificial Intelligence}

\begin{document}

\begin{frontmatter}

\title{Mitigating Illumination-Induced Domain Shift in Night-Time Pedestrian Detection for Intelligent Vehicles using Annotation-Preserving Diffusion Augmentation }

\author[hull]{Franky George\corref{cor1}}
\ead{f.george-2021@hull.ac.uk}

\author[hull]{Muhammad Khalid}
\ead{m.khalid@hull.ac.uk}

\author[hull]{Adil Khan}
\ead{a.m.khan@hull.ac.uk}

\author[hull]{Koorosh Aslansefat}
\ead{k.aslansefat@hull.ac.uk}

\cortext[cor1]{Corresponding author}
\affiliation[hull]{organization={School of Digital and Physical Sciences, University of Hull},
            city={Hull},
            country={United Kingdom}}

\begin{abstract}

Reliable night-time pedestrian detection remains challenging for vehicle-mounted vision systems because labelled night-time data are limited and the illumination shift between day and night reduces the effectiveness of detectors trained on daytime images. To address this challenge, we propose Contrastive Stable Diffusion XL (Contrastive-SDXL), an annotation-preserving diffusion-based augmentation framework that generates labelled night-time training data for vehicle-mounted pedestrian detection. Built on Stable Diffusion XL (SDXL)-Turbo and adapted using Low-Rank Adaptation, the framework employs patch-wise semantic contrastive supervision using features from a pre-trained DINOv2 encoder to maintain semantic correspondence. A detector-guided object consistency loss preserves pedestrians and the reliability of inherited annotations. Adversarial target-domain alignment encourages generated images to match night-time characteristics, while identity regularisation limits unnecessary changes to target-domain content. On the EuroCity Persons dataset, Contrastive-SDXL achieves a Fréchet Inception Distance of 22.5 against the real night-time distribution. Since visual similarity alone does not establish the utility of synthetic data for intelligent-vehicle perception, we further evaluate the synthetic images using Pedestron and You Only Look Once (YOLO). In the synthetic-only setting, the log-average miss rate decreases from 20.0\% to 16.0\% for Pedestron and from 19.0\% to 18.0\% for YOLO. When combined with limited labelled real night-time data, the synthetic images reduce the miss rate to 14.0\% for Pedestron, matching full real-night images training, and to 17.0\% for YOLO, compared with 16.0\% under full real-night training. These results show that preserving detector-relevant content and inherited annotations during diffusion-based augmentation can improve night-time pedestrian detection while reducing reliance on labelled night-time data.

\end{abstract}

\end{frontmatter}

\section{Introduction}
\label{sec:introduction}

Reliable pedestrian perception at night remains a persistent challenge for autonomous vehicles and advanced driver-assistance systems (ADAS). Although pedestrian detectors have improved substantially under daytime and favourable illumination conditions, their performance often degrades at night because reduced visibility, glare, sensor noise, and large differences in illumination alter the appearance of pedestrians and their surroundings~\citep{neumann2019nightowls,li2024light}. This illumination-induced domain shift is consequential for vehicle-mounted perception systems, in which reliable pedestrian detection is required for safety-critical functions such as automatic emergency braking and collision avoidance. Collecting and annotating representative night-time data across locations, traffic scenes, camera configurations, and illumination conditions can help mitigate this problem. However, doing so is costly and difficult to scale~\citep{braun2018eurocity,neumann2019nightowls}.

Day-to-night domain translation using generative models provides one approach to increasing the availability of night-time training data without requiring additional night-time images to be collected and manually annotated. For data augmentation in this setting, however, the objective is not simply to make daytime images appear visually similar to night-time scenes. A translated image is useful for downstream detector training only if it reflects the target operating conditions while preserving the source content and pedestrian structure sufficiently for the inherited annotations to remain valid. The challenge of achieving target-domain adaptation while preserving source content has been recognised in image translation for autonomous-driving applications. Day-to-night translation studies have reported semantic inconsistencies between source and translated driving scenes~\citep{lee2023daytonight}, while other work has shown that object instances may be distorted or insufficiently preserved during cross-domain image translation~\citep{lin2023cycleobject}. More recent work on adverse-condition adaptation similarly treats preservation of scene semantics as an important requirement when modifying the visual domain of labelled driving images~\citep{cha2025styleprompt}. These studies show that matching the target-domain appearance and preserving task-relevant source content are separate requirements, and achieving one does not necessarily guarantee the other.

To address this preservation requirement, existing image-translation approaches have introduced several mechanisms for retaining source content during appearance changes. For example, CUT uses patch-wise contrastive learning to encourage correspondence between spatial regions of the source and translated images~\citep{park2020contrastive}. InstaFormer incorporates instance representations and an instance-level contrastive objective to improve content preservation within object regions~\citep{kim2022instaformer}. Diffusion-based methods such as Palette and SDEdit demonstrate that strong generative priors can support substantial image transformation while retaining elements of the source content~\citep{saharia2022palette,meng2021sdedit}. LightDiff further shows that diffusion-based processing can incorporate perception-oriented constraints for low-light autonomous-driving applications~\citep{li2024light}. However, these methods address different aspects of the problem rather than jointly targeting night-time domain alignment, semantic preservation, and bounding-box consistency for pedestrian-detection augmentation. In this setting, useful augmentation requires the generated image to align with the night-time target distribution while retaining local and global scene semantics, preserving pedestrian structure, and maintaining sufficient spatial correspondence for the inherited bounding boxes to remain usable. Moreover, perceptual quality or distributional similarity metrics alone cannot establish these properties or determine whether the generated data improve a downstream detector. Ultimately, the value of the generated data must be demonstrated by their ability to improve downstream detector performance under the target operating conditions.

Consequently, an effective diffusion-based day-to-night augmentation framework must align generated images with the night-time domain while limiting changes that could invalidate detector-relevant source information and its associated annotations. To this end, we propose Contrastive-SDXL, an annotation-preserving day-to-night augmentation framework designed for night-time pedestrian detection. Built on SDXL-Turbo~\citep{sauer2024adversarial} and fine-tuned using Low-Rank Adaptation (LoRA)~\citep{hu2021lora}, the framework uses patch-wise semantic contrastive objectives to preserve correspondence between source and translated images. Unlike approaches that compute these objectives using generator encoder features, Contrastive-SDXL derives its semantic representations from selected layers of a frozen pre-trained DINOv2 encoder~\citep{oquab2023dinov2}. These multi-level representations provide external semantic supervision for preserving local structure and broader scene content during translation~\citep{jung2022exploring}. A detector-guided object consistency loss further encourages pedestrians to retain their structure and spatial locations, supporting the reliability of inherited bounding-box annotations. Adversarial target-domain alignment encourages the translated images to align with the visual characteristics of real night-time scenes, while identity regularisation encourages the model to preserve content that is already in the target domain.

Experiments on EuroCity Persons demonstrate that these design choices translate into improvements in the intended downstream task rather than only resulting in improved visual appearance. In the synthetic-only setting, the log-average miss rate for the \emph{All} pedestrian subset decreases from 20.0\% to 16.0\% for Pedestron~\citep{hasan2021generalizable} and from 19.0\% to 18.0\% for YOLO~\citep{yolo26_ultralytics}. Adding limited labelled real night-time data yields further gains. Pedestron reaches the performance obtained with the full real night-time training set \emph{All} and \emph{Small} pedestrian subset when the number of added real images corresponds to 20\% of the synthetic-set size, while YOLO approaches full real-night performance when the corresponding ratio is only 5\%. These results indicate that annotation-preserving diffusion augmentation can improve night-time pedestrian detection and reduce reliance on separately labelled night-time data in the evaluated setting.

The main contributions of this work are as follows:

\begin{itemize}

\item We propose Contrastive-SDXL, an annotation-preserving diffusion-based augmentation framework that adapts SDXL-Turbo to generate night-time pedestrian-detection training data from labelled daytime images.

\item We introduce DINOv2-guided patch-wise semantic contrastive objectives that use features from a frozen external semantic encoder, rather than the generator encoder, to preserve source-scene content and semantic correspondence during day-to-night translation.

\item We introduce a detector-guided object consistency loss that promotes pedestrian preservation and supports the reliability of inherited bounding-box annotations by penalising changes that distort or remove pedestrian instances.

\item We demonstrate the downstream utility of Contrastive-SDXL on the public EuroCity Persons dataset using two detector architectures, Pedestron and YOLO, in both synthetic-only training and settings where synthetic data are combined with limited labelled real night-time images.

\end{itemize}

\section{Related Work}
\label{sec:related_work}

\subsection{Night-time Pedestrian Augmentation}

Pedestrian detection is a fundamental component of vehicle-environment perception for autonomous vehicles and advanced driver-assistance systems (ADAS), but its reliability remains sensitive to illumination, visibility, pedestrian scale, and occlusion. Benchmarks such as Caltech, CityPersons~\citep{zhang2017citypersons}, and EuroCity Persons (ECP)~\citep{braun2018eurocity} have driven substantial progress with dense pedestrian annotations in urban traffic scenes. However, these datasets are still dominated by daytime or favourable lighting conditions. In ECP, labelled daytime images are considerably more numerous than labelled night-time images. This imbalance limits the range of night-time appearances available for detector training and contributes to an illumination-induced domain shift between daytime training data and night-time operation.

Existing approaches to night-time perception improvements can be broadly divided into low-light enhancement and day-to-night translation. Low-light enhancement improves images that have already been captured under night-time conditions. LightDiff~\citep{li2024light}, for example, uses a multi-condition diffusion framework and perception-specific guidance to enhance low-light driving images and improve downstream detection performance. Enhancement methods can increase the usability of existing night-time images, but they do not generate additional night-time training scenes from labelled daytime data. Day-to-night translation instead changes the illumination condition of a daytime image, creating the possibility of reusing its annotations if the underlying scene geometry and objects remain sufficiently consistent.

Early unpaired translation methods were predominantly GAN-based. CycleGAN~\citep{zhu2017unpaired} established cycle-consistent translation without paired examples. \citet{lin2021gan} later proposed AugGAN, a structure-aware unpaired day-to-night augmentation framework that transforms daytime images into night-time scenes and uses the generated datasets to train Faster R-CNN and YOLO. Other translation methods introduced additional structural constraints. \citet{s24041339}, for example, use semantic-segmentation and depth-estimation guidance to preserve semantic and geometric scene information during day-to-night translation. These works demonstrate the importance of retaining driving-scene structure and detector-relevant objects when changing illumination.

Recent diffusion-based methods provide alternative mechanisms for unpaired translation. CycleGAN-Turbo~\citep{parmar2024one} adapts a pre-trained text-to-image model for efficient one-step translation. Seed-to-Seed Translation~\citep{greenberg2024seed} learns an unpaired mapping between inverted diffusion seeds and uses structural conditioning to preserve the layout of complex driving scenes. These approaches improve translation efficiency and source-scene correspondence, but do not address whether small and occluded pedestrians remain sufficiently preserved and spatially aligned for their original bounding boxes to remain usable.

Our work addresses this gap by formulating day-to-night translation as an annotation-preservation problem for pedestrian-detector training rather than solely as a target-domain alignment problem. Contrastive-SDXL combines target-domain realism with semantic and object-level preservation using patch-wise correspondence in the feature space of a pre-trained DINOv2 encoder. A detector-guided object consistency loss is employed to preserve pedestrian consistency. Furthermore, we evaluate whether the generated images improve downstream night-time pedestrian detection compared with real night-time data and other day-to-night translation methods, providing empirical evidence of the usefulness of Contrastive-SDXL. This distinguishes the proposed framework from methods evaluated primarily as image-enhancement, scene-translation, or generic image-stylisation approaches.

\subsection{Consistency-preserving unpaired translation}

Preserving spatial and semantic correspondence between a source image and its translated output remains a central challenge in unpaired image-to-image translation. CycleGAN~\citep{zhu2017unpaired} learns forward and reverse mappings using adversarial objectives and a cycle-consistency constraint, which encourages the original input to be reconstructed after translation to the target domain and back. Although effective, cycle consistency does not explicitly guarantee that individual objects retain their original positions, boundaries, or semantic identities in the translated image. Consequently, small objects, such as distant pedestrians, may be distorted, displaced, removed, or semantically altered, thereby compromising the validity of source bounding boxes used to supervise downstream detector training~\citep{lin2021gan}.

Patch-wise contrastive learning offers a one-sided alternative to bidirectional cycle consistency. CUT~\citep{park2020contrastive} introduces PatchNCE, which maximises mutual information between corresponding regions of the input and translated images, encouraging local content preservation without requiring a reverse generator. Subsequent methods improve how spatial correspondence is modelled and how negative samples are selected. F/LSeSim~\citep{zheng2021spatially} preserves spatial self-similarity instead of enforcing direct agreement between source and target features. NEGCUT~\citep{wang2021instance} generates instance-specific hard negatives, PUT~\citep{lin2022exploring} investigates how unsuitable negatives can be identified and removed, and MoNCE~\citep{zhan2022modulated} adjusts the contribution of each negative sample. \citet{jung2022exploring} further introduce semantic relation consistency regularisation and hard-negative mining to preserve relationships among image patches under substantial appearance changes.

These formulations are particularly relevant to pedestrian augmentation because pedestrians in road scenes are often small or occluded, and appear against cluttered backgrounds. However, existing contrastive translation methods generally extract patch representations from the encoder of a GAN-based generator. These features are learned jointly with the translation task and can be extracted from corresponding encoder layers for both the source and translated images. In diffusion generators, by contrast, intermediate U-Net activations are produced during a stochastic denoising process, potentially making them less stable for semantic comparisons between source and translated images. Therefore, to decouple semantic contrastive supervision from the diffusion generator, Contrastive-SDXL uses a frozen DINOv2~\citep{oquab2023dinov2} encoder to extract patch features from both the daytime source image and its translated night-time counterpart. This choice is motivated by evidence that self-supervised vision transformers, such as DINOv2, learn spatially meaningful, object-sensitive representations without manual labels~\citep{caron2021emerging}. Semantic relation consistency (SRC) and decoupled contrastive learning with hard negatives (hDCE) objectives~\citep{jung2022exploring} are computed across selected DINOv2 layers to preserve local structure and higher-level semantic relationships. These general correspondence objectives are complemented by the detector-guided object consistency loss, which provides the explicit pedestrian-specific supervision required by the augmentation task.

\section{Problem Formulation and Design Objectives}
\label{sec:design_objectives}

This work formulates day-to-night image translation as an annotation-preserving cross-domain augmentation problem for vehicle-mounted pedestrian detection. We consider an unpaired setting in which the daytime source domain $S$ contains abundant labelled data, while the night-time target domain $T$ contains only a limited number of samples.

Let $X_S$ and $X_T$ denote the source and target image sets, respectively. Then the corresponding labelled datasets are:
\begin{equation}
D_S = \{(x_S^i, y_S^i)\}_{i=1}^N, \quad 
D_T = \{(x_T^j, y_T^j)\}_{j=1}^M, \quad N \gg M,
\end{equation}
where $x_S^i, x_T^j \in \mathbb{R}^{H \times W \times C}$ denote source and target images, and $y_S^i, y_T^j$ denote their pedestrian bounding-box annotations. 

The objective is to learn a mapping
\begin{equation}
G_{\theta}: S \rightarrow T,
\end{equation}

that translates daytime images $x_S \in X_S$ into the night-time domain while retaining the detector-relevant content of the source scene. For each source image, a synthetic night-time image is generated as

\begin{equation}
\tilde{x}_T^i = G_{\theta}(x_S^i), \quad i=1,\dots,N.
\end{equation}

The resulting augmented dataset is
\begin{equation}
\tilde{D}_T = \{(\tilde{x}_T^i, y_S^i)\}_{i=1}^N,
\end{equation}

where the source annotations $y_S^i$ are reused, provided that the pedestrians remain sufficiently preserved and spatially aligned after translation. At the distribution level, the generated images in $\tilde{X}_T$ should approximate the target-domain distribution:

\begin{equation}
p(\tilde{X}_T) \approx p(X_T).
\end{equation}

This condition alone is insufficient for detection-oriented augmentation. A translated image may resemble the night-time target distribution while containing distorted, displaced, or missing pedestrians that reduce the reliability of its inherited annotations. The problem therefore requires target-domain alignment and source-content preservation to be addressed jointly.

Based on these requirements, we define four design objectives:

\begin{enumerate}
    \item \textbf{Target-domain distribution alignment:} Generated images should match the visual characteristics of the target domain and reduce the distributional discrepancy between synthetic and target-domain data.
    \item \textbf{Semantic consistency:} Translation should preserve source-image semantics, including objects and their spatial relationships.
    \item \textbf{Pedestrian and annotation integrity:} Pedestrians should remain visible, structurally consistent, and sufficiently aligned with their original positions and extents.
    \item \textbf{Parameter-efficient adaptation:} The pre-trained generative model should be adapted efficiently to the day-to-night task while preserving its learned generative prior. 
\end{enumerate}

\section{Proposed Method}

\begin{figure*}[ht]
\centering
\includegraphics[width=\textwidth]{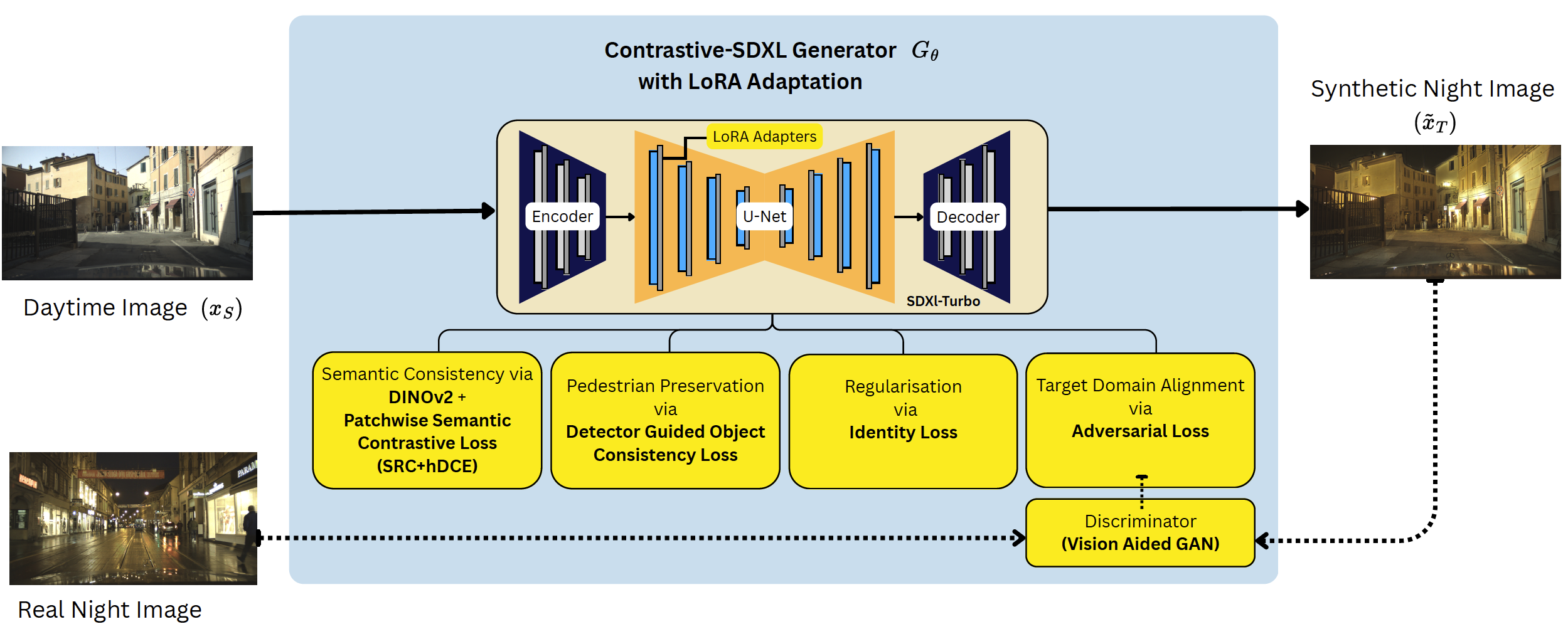}
\caption{Overview of Contrastive-SDXL. A daytime image $x_S$ is translated into a synthetic night-time image $\tilde{x}_T$ using an SDXL-Turbo generator $G_\theta$ with LoRA adaptation. The model is guided by patch-wise semantic contrastive loss, detector-guided object consistency, identity regularisation, and adversarial alignment.}
\label{fig:contrastive_sdxl}
\end{figure*}

To address the design objectives defined in Section~\ref{sec:design_objectives}, we propose \textbf{Contrastive-SDXL}, a diffusion-based cross-domain augmentation framework that translates labelled daytime images into the night-time domain while promoting semantic consistency, pedestrian preservation, and the reliability of inherited bounding-box annotations. The framework adapts the pre-trained generator to the target domain using Low-Rank Adaptation (LoRA). Its training combines DINOv2-guided patch-wise semantic contrastive objectives for general source-content preservation, a detector-guided object consistency loss for pedestrian-specific supervision, an adversarial loss for target-domain alignment, and identity regularisation to limit unnecessary content changes. Figure~\ref{fig:contrastive_sdxl} provides an overview of the complete framework.

\subsection{Generator Backbone and Efficient Adaptation}

To support \textbf{target-domain alignment}, Contrastive-SDXL uses SDXL-Turbo, an adversarially distilled latent diffusion model~\citep{sauer2024adversarial}, as its generator backbone. Designed for low-step image generation, SDXL-Turbo is configured as a one-step generator in our framework. Its pre-trained generative prior provides a strong basis for realistic image synthesis, while alignment with the specific night-time domain is learned through LoRA and the proposed training objectives, particularly the adversarial alignment described in Section~\ref{sec:adversarial_alignment}. Inspired by \citet{parmar2024one}, we also incorporate encoder--decoder skip connections into the generator to support the preservation of fine-grained spatial structure.

Given a daytime source image $x_S$, the modified variational autoencoder retains encoder activations at multiple spatial scales while encoding the image into its latent representation:
\begin{equation}
\left(z_S,\mathcal{H}_S\right)
=
\mathrm{VAE}_{\mathrm{enc}}(x_S),
\qquad
\mathcal{H}_S
=
\left\{h_S^{\ell}\right\}_{\ell=1}^{\mathcal{L}},
\end{equation}

where $z_S$ is the source latent and $\mathcal{H}_S$ contains the encoder activations extracted at $\mathcal{L}$ spatial resolutions. These activations are passed to corresponding decoder stages through trainable skip connections. The source latent is processed by the conditioned one-step denoising process and decoded using the retained encoder activations $\mathcal{H}_S$:
\begin{equation}
\tilde{x}_T
=
\mathrm{VAE}_{\mathrm{dec}}
\left(
\mathcal{D}_{\theta}(z_S,t,c);\mathcal{H}_S\right),
\end{equation}

where $\mathcal{D}_{\theta}$ denotes the one-step U-Net prediction and the scheduler update, while $c$ is the text prompt embedding.

To satisfy \textbf{efficient adaptation}, we use LoRA~\citep{hu2021lora} instead of fully fine-tuning SDXL-Turbo. Given a pre-trained weight matrix $W_0 \in \mathbb{R}^{d \times k}$, LoRA decomposes the update $\Delta W$ as
\begin{equation}
\Delta W = AB, \quad r \ll \min(d,k),
\end{equation}
where $A \in \mathbb{R}^{d \times r}$ and $B \in \mathbb{R}^{r \times k}$ are trainable, giving
\begin{equation}
W = W_0 + \Delta W.
\end{equation}

\subsection{Patch-Wise Semantic Contrastive Supervision with DINOv2}

To encourage \textbf{semantic consistency} between a daytime source image and its translated night-time counterpart, we build on the semantic relation consistency (SRC) and decoupled contrastive learning with hard negatives (hDCE) objectives proposed by \citet{jung2022exploring}. These objectives preserve relationships among corresponding image patches without requiring a reverse translation mapping. Unlike the original formulation, which computes these objectives using generator encoder features, our framework applies them to multi-level patch representations extracted from a frozen DINOv2 encoder. We first describe the adopted objectives and then explain their integration with DINOv2.

\paragraph{Semantic Relation Consistency Loss}
Let $z_k^l$ and $\tilde{z}_k^l$ denote the normalised projected features of patch $k$ extracted from the source feature maps $F_S^l$ and the translated feature maps $F_T^l$, respectively, at layer $l$. For every $l \in L$, SRC aims to ensure that the semantic relationships between $z_k^l$ and the other sampled patches $\{z_i^l\}_{i=1}^{N_p}$ are preserved between their corresponding translated patches $\tilde{z}_k^l$ and $\{\tilde{z}_i^l\}_{i=1}^{N_p}$. In other words, SRC encourages the patch-wise semantic relationships within the source image to be preserved among the corresponding patches in the translated image. These pairwise relationships define a similarity distribution over patches. For a source patch $z_k^l$ and translated patch $\tilde{z}_k^l$, the similarity relations with patches  $\{z_i^l\}$ and $\{\tilde{z}_i^l\}$ are defined as
\begin{equation}
\begin{aligned}
S_k^l(i)
&=
\frac{
    \exp\left((z_k^l)^{\top}z_i^l\right)
}{
    \sum_{\substack{j=1\\j\neq k}}^{N_p}
    \exp\left((z_k^l)^{\top}z_j^l\right)
}, \\[4pt]
T_k^l(i)
&=
\frac{
    \exp\left((\tilde{z}_k^l)^{\top}\tilde{z}_i^l\right)
}{
    \sum_{\substack{j=1\\j\neq k}}^{N_p}
    \exp\left((\tilde{z}_k^l)^{\top}\tilde{z}_j^l\right)
}.
\end{aligned}
\end{equation}

Here, $N_p$ denotes the number of sampled patches. These distributions describe how each query patch relates to all sampled regions within the same image.

The SRC loss at layer $l$ is computed as the Jensen--Shannon divergence between the corresponding similarity distributions:
\begin{equation}
L_{SRC}^l = \sum_{k=1}^{N_p} \mathrm{JSD}(S_k^l \parallel T_k^l).
\end{equation}

Minimising this loss encourages the translated image to retain the source image's patch-wise semantic structure despite the intended illumination change.

\paragraph{Hard Negative Contrastive Loss}
Complementing SRC, hDCE directly encourages correspondence between matching source and translated patches. For a translated query patch $\tilde{z}_k^l$, the positive is the source feature $z_k^l$ at the same spatial location while the remaining sampled source patches, $\{z_j^l\}_{j \neq k}$, form the negative set. The hDCE objective maximises the similarity of the positive pair $(z_k^l, \tilde{z}_k^l)$ while minimising similarity between  $\tilde{z}_k^l$ and hard negatives. These hard negatives are semantically similar patches from different spatial locations, thereby providing a stronger contrastive signal.

In practice, hard negatives are semantically weighted according to their similarity to the corresponding positive source patch $z_k^l$. Let $\hat{\mathbf{F}}_S^l$ denote the matrix whose rows are the normalised projected source-patch features:
\begin{equation}
\hat{\mathbf{F}}_S^l = \left[ z_1^l, z_2^l, \ldots, z_{N_p}^l\right]^\top \in \mathbb{R}^{N_p \times d}, 
\end{equation}
where $N_p$ is the number of sampled patches at layer $l$, and $d$ is the projected feature dimension. The intra-image similarity matrix is then computed as:
\begin{equation}
\mathbf{M}^l = \hat{\mathbf{F}}_S^l (\hat{\mathbf{F}}_S^l)^{\top}, \qquad \mathbf{M}^l \in \mathbb{R}^{N_p \times N_p},
\end{equation}
where $M_{ki}^l$ is the cosine similarity between the projected patches $z_k^l$ and $z_i^l$ at layer $l$. Assuming that the empirical base distribution is uniform over the
$N_p-1$ candidate negative patches, the hard-negative weights are derived as
\begin{equation}
W_{ki}^l = \frac{\exp(\gamma M_{ki}^l )}{\sum_{\substack{j=1\\j\neq k}}^{N_p} \exp(\gamma M_{kj}^l )}, \qquad i\neq k,
\end{equation}
with $W_{kk}^l=0$. Here, $\gamma$ controls the hardness of the negative distribution. Larger values of $\gamma$ concentrate the distribution on source patches that are more semantically similar to $z_k^l$, whereas $\gamma=0$ corresponds to uniform weighting over the negative patches.

For a source patch $z_k^l$ at layer $l$, let $q_{Z_s^-}^l(\cdot;z_k^l,\gamma)$ denote the discrete negative distribution defined over the negative patches $\{z_j^l\}_{j \neq k}$, where 
\begin{equation}
q_{Z_s^-}^l(z_i^l;z_k^l,\gamma) = W_{ki}^l.
\end{equation}

Then, for a positive pair $(\tilde{z}_k^l,z_k^l)$, the layer-wise hDCE objective is
\begin{equation}
\begin{aligned}
&L_{\mathrm{hDCE}}^{l}(\gamma,\tau)
\\
&=
-\frac{1}{N_p}
\sum_{k=1}^{N_p}
\log
\frac{
    \exp\!\left((\tilde{z}_k^l)^{\top} z_k^l/\tau\right)
}{
    N^-\,
    \mathbb{E}_{z_j^l \sim q_{Z_s^-}^l(\cdot;z_k^l,\gamma)}
    \left[
        \exp\!\left((\tilde{z}_k^l)^{\top} z_j^l/\tau\right)
    \right]
},
\end{aligned}
\end{equation}
where $N^-=N_p-1$, $\tau$ is the contrastive temperature, and $\gamma$ controls the hardness of negative sampling.

\paragraph{DINOv2-Based Multi-Level Feature Extraction}

\begin{figure}[tb]
    \centering

    \begin{subfigure}[t]{0.48\textwidth}
        \vspace{0pt}
        \centering
        \includegraphics[width=\linewidth]{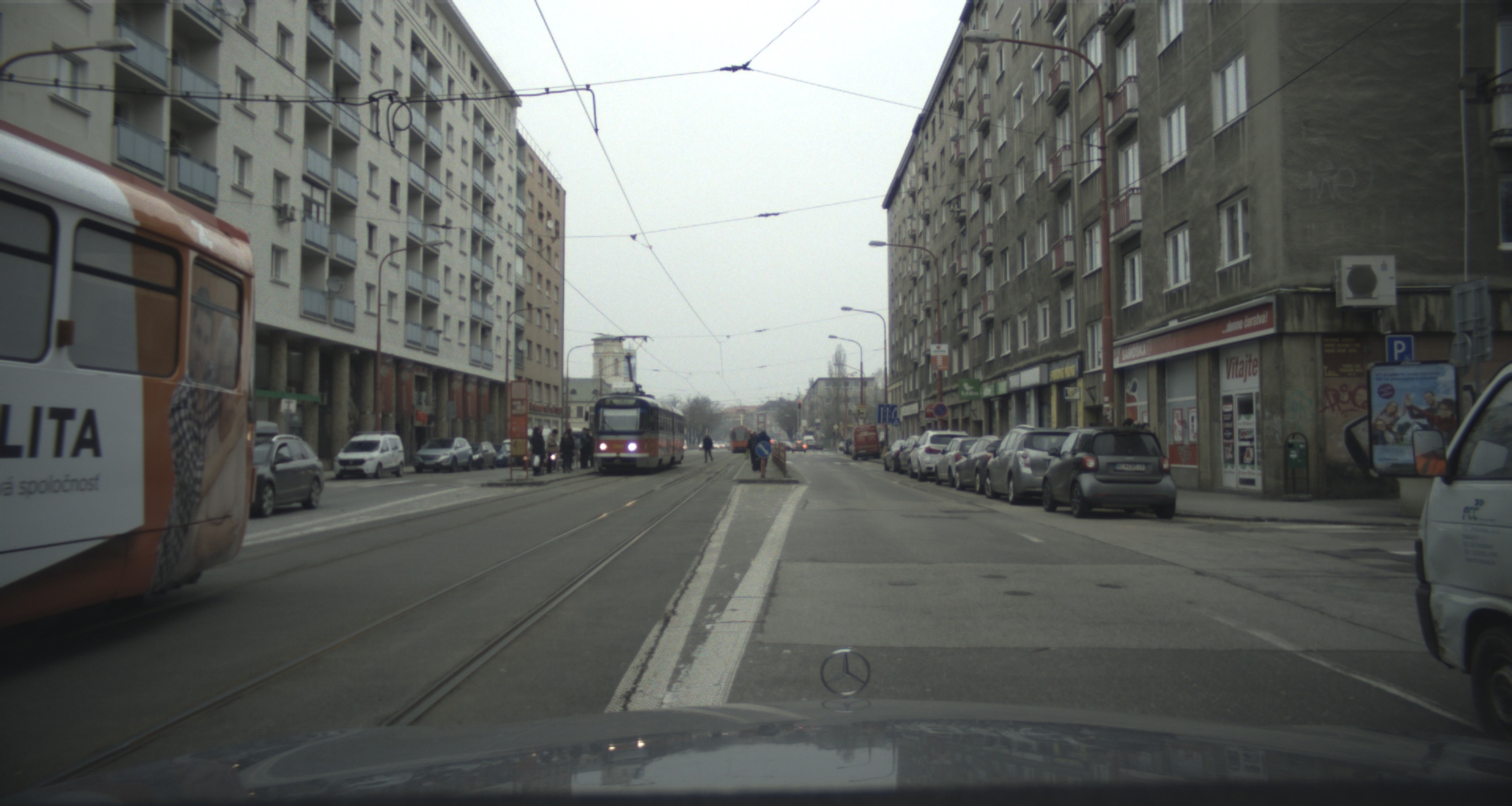}
        \caption{Real daytime image from the ECP dataset}
        \label{fig:sub1}
    \end{subfigure}
    \begin{subfigure}[t]{0.48\textwidth}
        \vspace{0pt}
        \centering
        \includegraphics[width=\linewidth]{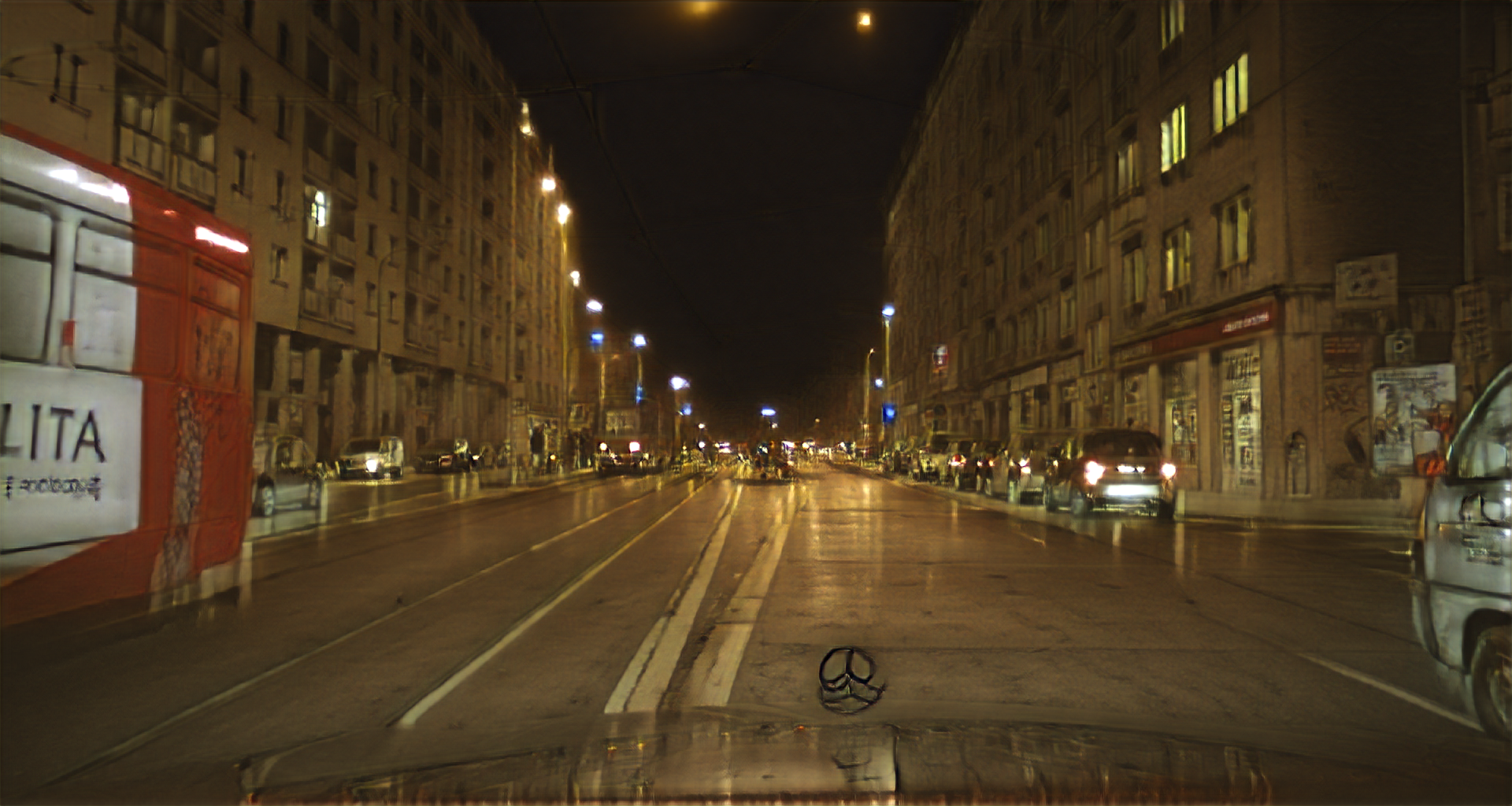}
        \caption{Night-time image generated using UNet encoder features}
        \label{fig:sub2}
    \end{subfigure}

    \caption{A comparison between a real daytime image and a night-time image generated by Contrastive-SDXL using UNet encoder features for patch-wise contrastive loss. The generated image exhibits significant artefacts (e.g., lane lines) and fails to preserve details (e.g., pedestrians, shop boards and the Mercedes logo), indicating challenges in providing cues to learn visual coherence during translation.}
    \label{fig:dinov2_comparison}
\end{figure}

A key modification in our framework is the source of patch features. In the original formulation, the contrastive objectives are computed using features extracted from the generator encoder. However, in our evaluation, we found this approach to be less effective for diffusion models, in which the intermediate U-Net activations are produced within a timestep-conditioned stochastic denoising process. In particular, the Contrastive-SDXL variants trained using U-Net features, shown in Figure~\ref{fig:dinov2_comparison} and~\ref{fig:unet_vs_dinov2}, fail to maintain semantic correspondence and often produce artefacts, distortions, and poor pedestrian preservation. We therefore decouple semantic supervision from the generator encoder and use a frozen, pre-trained DINOv2 semantic encoder~\citep{oquab2023dinov2} to extract semantic and context-aware patch features.

Let $\tilde{x}_T = G_\theta(x_S)$ denote the night-time translation of a daytime source image $x_S$. Both images are processed by the DINOv2 encoder $E_{\mathrm{DINO}}$. For each selected layer $l\in L$, the corresponding patch-token features are given by
  \begin{equation} 
    F_S^l = E_{\mathrm{DINO}}^l(x_S), \; F_T^l = E_{\mathrm{DINO}}^l(\widetilde{x}_T), \; F_S^l,F_T^l\in\mathbb{R}^{N_l'\times C_l}, 
  \end{equation}
where $N_l'$ denotes the number of patch tokens after discarding special tokens and $C_l$ is the dimensionality of each patch token. 

The same set of $N_p$ spatial indices, denoted by $\Omega_l = \{\omega_1,\ldots,\omega_{N_p}\} \subseteq \{1,\ldots,N_l'\}$, is sampled from the source and translated feature maps at layer $l$. A layer-specific two-layer projection head $\operatorname{MLP}_{\mathrm{proj}}^l$ maps the selected features into a shared $d$-dimensional embedding space: 
\begin{equation}
\begin{aligned}
z_k^l
    &= \operatorname{norm}( \operatorname{MLP}_{\mathrm{proj}}^l(F_S^l[\omega_k])), \\
\tilde{z}_k^l
    &= \operatorname{norm}(
        \operatorname{MLP}_{\mathrm{proj}}^l(F_T^l[\omega_k])),
       \qquad k = 1,\ldots,N_p,
\end{aligned}
\end{equation}
where $\operatorname{norm}(\cdot)$ denotes $\ell_2$ normalisation.

\begin{figure*}[tb]
    \centering
    \includegraphics[width=\textwidth]{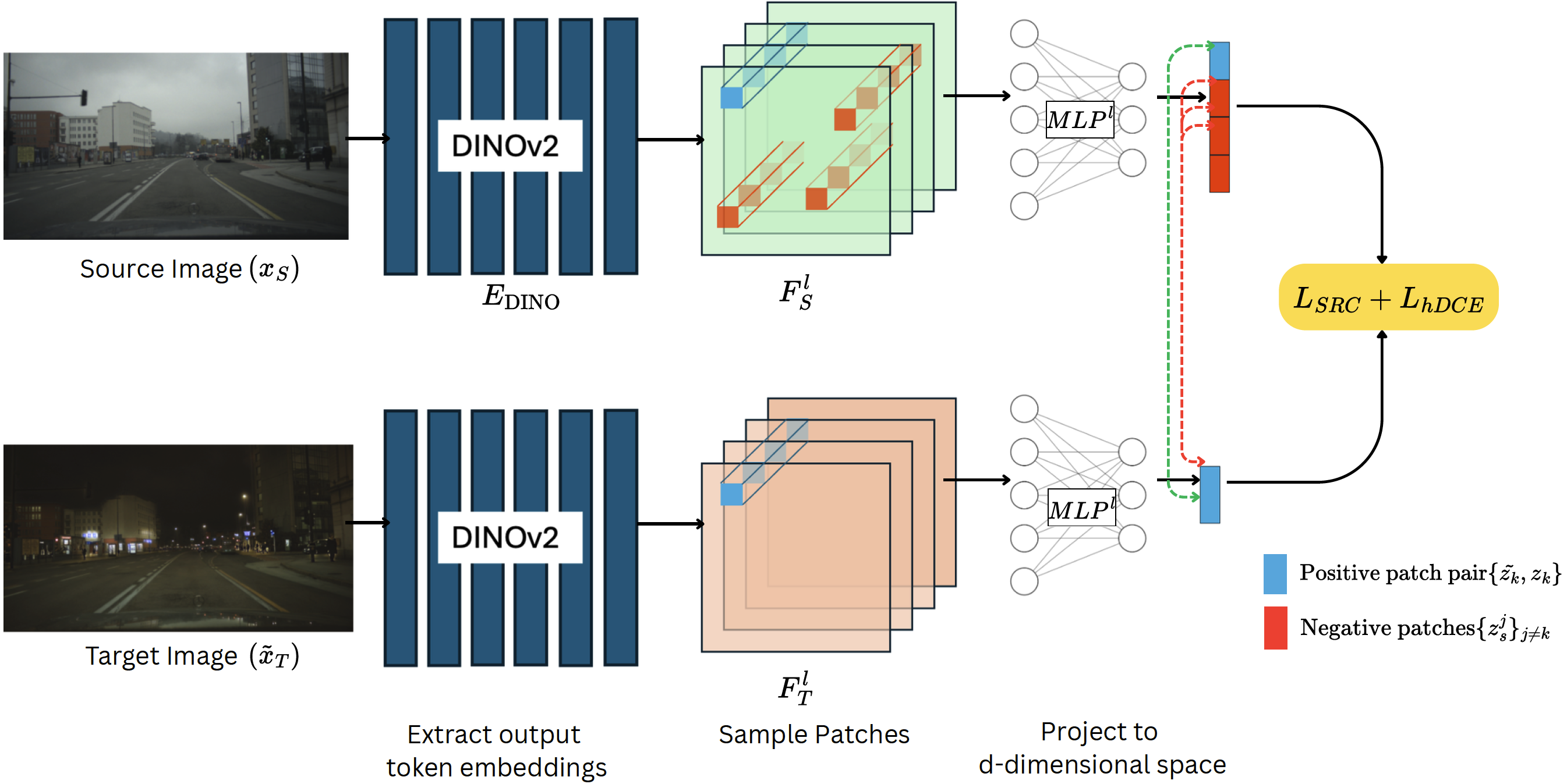}
    \caption{DINOv2 encoder used to extract multi-level semantic features for patch-wise contrastive loss computation.}
    \label{fig:dinov2}
\end{figure*}

Patch-wise contrastive losses are computed across all selected layers, treating corresponding spatial locations as positives and other patches as negatives, as illustrated in Figure~\ref{fig:dinov2}. The complete semantic contrastive objective is averaged across the selected DINOv2 layers:
\begin{equation}
L_{\mathrm{sem}}
=
\frac{1}{|L|}
\sum_{l\in L}
\left(
    \lambda_{\mathrm{SRC}}
    L_{\mathrm{SRC}}^l
    +
    \lambda_{\mathrm{hDCE}}
    L_{\mathrm{hDCE}}^l
\right).
\end{equation}
Extracting features at multiple representation depths allows the objective to constrain both local structural relationships and higher-level semantic correspondence, leading to night-time images that are more reliable for downstream pedestrian detector training.

\subsection{Detector-Guided Object Consistency Loss}

The DINOv2-guided contrastive objectives preserve general source-scene content, structure and semantic relationships, but do not explicitly guarantee that pedestrians remain detectable and spatially consistent after translation. To address the \textbf{pedestrian preservation and annotation reliability} objective, we introduce a detector-guided object consistency loss that encourages the generator to preserve pedestrian-specific visual cues using a pre-trained YOLO11 detector fine-tuned on EuroCity Persons (ECP)~\citep{braun2018eurocity}. The detector is kept fixed and used only to provide task-specific supervision to the generator. Its parameters remain frozen during Contrastive-SDXL training, while gradients with respect to its input are retained so that the detection loss can update the generator. 

Let $\tilde{x}_T = G_\theta(x_S)$ denote the translation of a source image $x_S$ and $y_S=\left\{(c_i,b_i)\right\}_{i=1}^{n_o}$ denote its inherited source annotations, where $c_i$ and $b_i$ are the pedestrian class and bounding box of object $i$, respectively, and $n_o$ is the number of annotated objects in the image. The translated image $\tilde{x}_T$ is passed through the frozen detector $\operatorname{D}_{\psi}$ to obtain its dense training predictions:
\begin{equation}
\mathbf{p}_T
=
\operatorname{D}_{\psi}\left(\tilde{x}_T\right).
\end{equation}

The detector-guided loss evaluates these predictions against the inherited source annotations:
\begingroup
\small
\begin{equation}
L_{\mathrm{det}}
=
\lambda_{\mathrm{box}} L_{\mathrm{box}}
\left(\mathbf{p}_T, y_S\right)
+
\lambda_{\mathrm{cls}} L_{\mathrm{cls}}
\left(\mathbf{p}_T, y_S\right)
+
\lambda_{\mathrm{dfl}} L_{\mathrm{dfl}}
\left(\mathbf{p}_T, y_S\right),
\end{equation}
\endgroup

where $L_{\text{box}}$, $L_{\text{cls}}$, and $L_{\text{dfl}}$ denote the bounding box regression, classification, and distribution focal losses, respectively. We use the default YOLO11 weights $\lambda_{\text{box}}=7.5$, $\lambda_{\text{cls}}=0.5$, and $\lambda_{\text{dfl}}=1.5$.

Here, $L_{\text{box}}$ penalises localisation errors between predicted and source bounding boxes using CIoU-style regression~\citep{zheng2020distance}. $L_{\mathrm{cls}}$ uses binary cross-entropy to encourage the detector's class predictions to remain consistent with the inherited pedestrian targets, while $L_{\mathrm{dfl}}$ refines localisation by modelling the distances from anchor points to the four bounding-box edges as discrete distributions~\citep{li2020generalized}. The loss is computed from the detector's raw training outputs before confidence thresholding and non-maximum suppression to preserve differentiability. During generator optimisation, gradients from $L_{\text{det}}$ are propagated through the frozen detector to the generator, thereby penalising translations in which pedestrians expected at inherited locations become inconsistent with the source classes or bounding boxes.

\subsection{Adversarial Target-Domain Alignment}
\label{sec:adversarial_alignment}

The pre-trained SDXL-Turbo backbone provides a broad image prior, but this alone does not guarantee alignment with the specific appearance of real ECP night-time scenes. To address the \textbf{target-domain distribution alignment} objective, we employ a Vision-Aided adversarial loss~\citep{kumari2022ensembling}, denoted by $L_{\mathrm{adv}}$, to encourage translated images to resemble the target night-time domain images.

Let $\Phi_{\mathrm{CLIP}}$ denote a frozen CLIP image encoder, and let $\{d_{\phi}^{m}\}_{m=1}^{K}$ denote $K$ trainable discriminator heads operating on features extracted from $K$ representation levels of the CLIP encoder. For an input image $x$, each discriminator head $d_{\phi}^{m}$ operates on the corresponding CLIP representation $\Phi_{\mathrm{CLIP}}^{m}(x)$, and the resulting loss terms are summed across all $K$ representation levels. The CLIP feature extractor remains fixed, while the discriminator heads are trained to distinguish real target-domain images from translated source images.

Using the multilevel sigmoid objective employed by the Vision-Aided discriminator, the discriminator loss is defined as
\begin{equation}
\begin{aligned}
L_{D}
={}&
\mathbb{E}_{x_T\sim X_T}
\left[
\sum_{m=1}^{K}
\ell_{\mathrm{sig}}
\left(
d_{\phi}^{m}
\left(\Phi_{\mathrm{CLIP}}^{m}(x_T)\right),
\alpha
\right)
\right]
\\
&+
\mathbb{E}_{x_S\sim X_S}
\left[
\sum_{m=1}^{K}
\ell_{\mathrm{sig}}
\left(
d_{\phi}^{m}
\left(\Phi_{\mathrm{CLIP}}^{m}
\left(G_{\theta}(x_S)\right)\right),
0
\right)
\right],
\end{aligned}
\end{equation}
where $\ell_{\mathrm{sig}}$ denotes the sigmoid classification loss, and $\alpha = 0.8$ denotes the smoothed label assigned to real target-domain images in the Vision-Aided discriminator. The fake-class label remains $0$.

The generator adversarial loss encourages translated images to be classified as belonging to the real target domain and is defined as
\begin{equation}
L_{\mathrm{adv}}
=
\mathbb{E}_{x_S\sim X_S}
\left[
\sum_{m=1}^{K}
\ell_{\mathrm{sig}}
\left(
d_{\phi}^{m}
\left(\Phi_{\mathrm{CLIP}}^{m}
\left(G_{\theta}(x_S)\right)\right),
\alpha
\right)
\right].
\end{equation}

Together with the pre-trained diffusion prior, $L_{\text{adv}}$ encourages target-domain realism and distribution-level alignment, while the contrastive and detector-guided losses preserve pedestrian-relevant structure.

\subsection{Identity Regularisation}

We apply identity regularisation to discourage unnecessary changes when the input already belongs to the target domain. Given a night-time image $x_T \sim X_T$, the generator is encouraged to reconstruct it with minimal alteration:
\begin{equation}
L_{\mathrm{idt}} =
\mathbb{E}_{x_T \sim X_T}\left[\|G_{\theta}(x_T) - x_T\|_1\right].
\end{equation}
Minimising $L_{\mathrm{idt}}$ penalises unnecessary pixel-level changes to images that already exhibit the target-domain appearance and it complements the contrastive, detector-guided, and adversarial objectives.

\subsection{Full Objective}

The overall objective of Contrastive-SDXL combines the patch-wise semantic contrastive losses, detector-guided object consistency, identity regularisation, and adversarial alignment:
\begin{equation}
L_{\text{total}} =
L_{\text{sem}} +
\lambda_{\text{det}} L_{\text{det}} +
\lambda_{\text{idt}} L_{\text{idt}} +
\lambda_{\text{adv}} L_{\text{adv}}.
\end{equation}

Here, $L_{\text{sem}}$ comprises the SRC and hDCE objectives weighted by $\lambda_{\text{SRC}}$ and $\lambda_{\text{hDCE}}$, respectively. These terms encourage preservation of patch-wise semantic relationships and similarity between corresponding source and translated regions. The coefficient $\lambda_{\text{det}}$ controls the contribution of the detector-guided loss $L_{\text{det}}$, which encourages preservation of pedestrians in the translated images. The identity loss $L_{\text{idt}}$ discourages unnecessary changes to target-domain inputs, while the generator-side adversarial loss $L_{\text{adv}}$ encourages translated images to resemble the real night-time domain.

By jointly optimising these objectives, Contrastive-SDXL balances target-domain realism with semantic consistency and pedestrian preservation, making the generated images more reliable for downstream night-time pedestrian detection.

\section{Experiments}

\subsection{Dataset and Translation Baselines}

We conduct our experiments on the EuroCity Persons (ECP) dataset~\citep{braun2018eurocity}, a large-scale benchmark for pedestrian detection in vehicle-mounted urban traffic scenes. ECP contains more than 238,200 annotated person instances in 47,300 images collected across 31 cities in 12 European countries. The dataset includes both daytime and night-time scenes, allowing day-to-night image translation to be studied within a common dataset and annotation framework. The labelled daytime images are used as the source domain for synthetic night-time image generation, while the comparatively limited set of real ECP night-time images is used as target-domain data. Downstream detector performance is evaluated on the real ECP night-time validation split. The exact composition of each subset is described in the corresponding experimental protocol.

NightOwls~\citep{neumann2019nightowls} also provides a benchmark for night-time pedestrian detection. However, because it contains only night-time images, it does not support the controlled day-to-night translation experiments considered in this work. Using ECP daytime images as the source domain and NightOwls images as the target domain would introduce cross-dataset differences in image characteristics and annotation protocols, beyond the intended change in illumination. We therefore use ECP to provide a controlled evaluation of the day-to-night augmentation framework.

We compare Contrastive-SDXL with two representative diffusion-based image-to-image translation baselines, InstructPix2Pix~\citep{brooks2023instructpix2pix} and CycleGAN-Turbo~\citep{parmar2024one}. InstructPix2Pix is a pre-trained diffusion-based image-editing model, which we use to assess whether text-guided diffusion editing can generate useful night-time images for data augmentation. CycleGAN-Turbo is an unpaired image translation model built on a text-to-image diffusion backbone and serves as a task-specific baseline for efficient day-to-night translation.

Both baselines are used with their default configurations. All methods are applied to the same set of daytime source images, and the generated images are evaluated using the same downstream detector-training and evaluation procedure. This comparison examines whether the different translation methods produce target-domain images that not only appear night-like but also improve the training of downstream pedestrian detectors.

\subsection{Pedestrian Detectors}

We evaluate the augmentation data using two downstream detectors with different architectural designs, a pedestrian-oriented two-stage Cascade Mask R-CNN with an HRNet backbone implemented in Pedestron~\citep{hasan2021generalizable} and YOLO26x, a general-purpose one-stage detector~\citep{yolo26_ultralytics}. Evaluating both detectors provides evidence as to whether the effects of synthetic night-time augmentation are consistent across two different detector architectures.

For Pedestron, we initialise the Cascade Mask R-CNN--HRNet model using the publicly available ECP daytime checkpoint. This checkpoint serves as the daytime-only baseline and is subsequently fine-tuned on the night-time adaptation sets, including synthetic-only, mixed real-synthetic, and full real-night training. Each configuration is therefore fine-tuned from the same daytime-trained model rather than trained from scratch.

For YOLO26, we train a COCO-pre-trained YOLO26x model on the ECP daytime training split at resolution of $1280 \times 1280$. This model serves as the corresponding YOLO baseline and provides the common initialisation for all subsequent adaptation experiments including fine-tuning on synthetic-only, mixed real-synthetic, and full real-night sets.  Fine-tuning is performed progressively by first freezing most of the backbone, then unfreezing intermediate layers, and finally training the full network with a lower learning rate and mosaic augmentation disabled.

For each detector, we consider four training configurations:

\begin{enumerate}
\item \textbf{Daytime-only baseline:} The detector is trained using labelled ECP daytime data without night-time adaptation. It is evaluated on both the daytime and night-time validation splits to quantify the illumination-induced domain gap.

\item \textbf{Synthetic-only adaptation:} The daytime detector is fine-tuned using the synthetic night-time images, with the inherited daytime bounding-box annotations to test whether generated data alone can support target-domain adaptation.

\item \textbf{Mixed real-synthetic adaptation:}  The daytime-trained detector is fine-tuned using the synthetic set together with limited quantities of labelled real night-time images. The quantity of added real night-time data is expressed using an injection rate $\rho$. An injection rate of $\rho\%$ means that the number of added real night-time images equals $\rho\%$ of the corresponding synthetic-set size:
\begin{equation}
    \left|X_{\mathrm{injected}}^{(\rho)}\right|
    =
    \frac{\rho}{100}
    \left|\tilde{X}_T\right|.
\end{equation}
This configuration represents settings in which only limited labelled target-domain data are available.

\item \textbf{Full real-night adaptation:} The daytime detector is fine-tuned using the complete real ECP night-time training set, which provides a fully supervised target-domain reference.

\end{enumerate}

All adaptation configurations are evaluated on the real ECP night-time validation split. The daytime-only baselines are additionally evaluated on the ECP daytime validation split to quantify the illumination-induced domain gap. The evaluation metrics and subsets are described in the following evaluation protocol.

\begin{figure}[tb]
      \centering

      \begin{minipage}[t]{0.32\textwidth}
          \centering
          {\footnotesize\textbf{Original Image}}\par\vspace{1mm}
          \includegraphics[width=\linewidth]{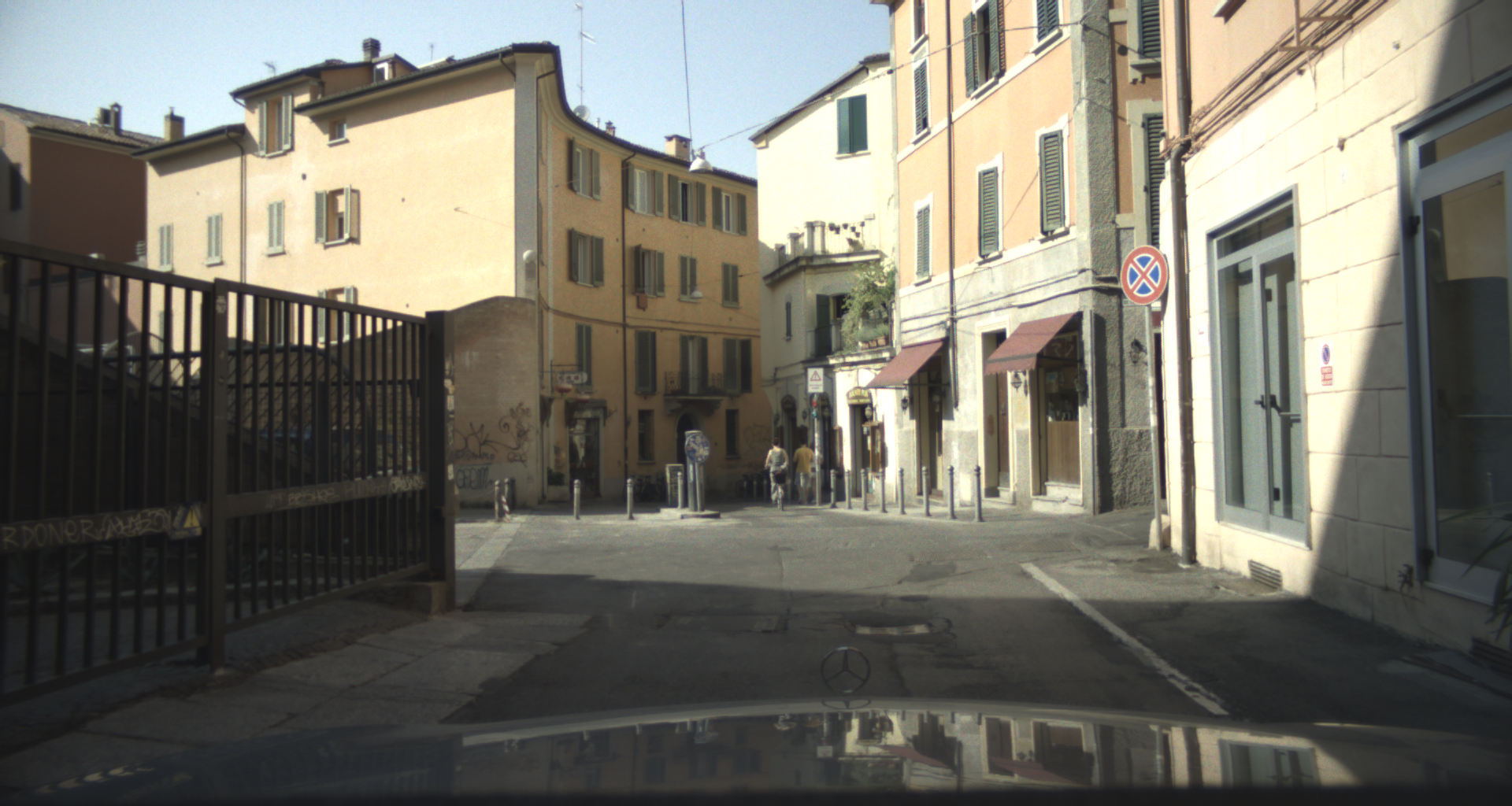}
      \end{minipage}
      \hspace{0.0002\textwidth}
      \begin{minipage}[t]{0.32\textwidth}
          \centering
          {\footnotesize\textbf{U-Net Encoder}}\par\vspace{1mm}
          \includegraphics[width=\linewidth]{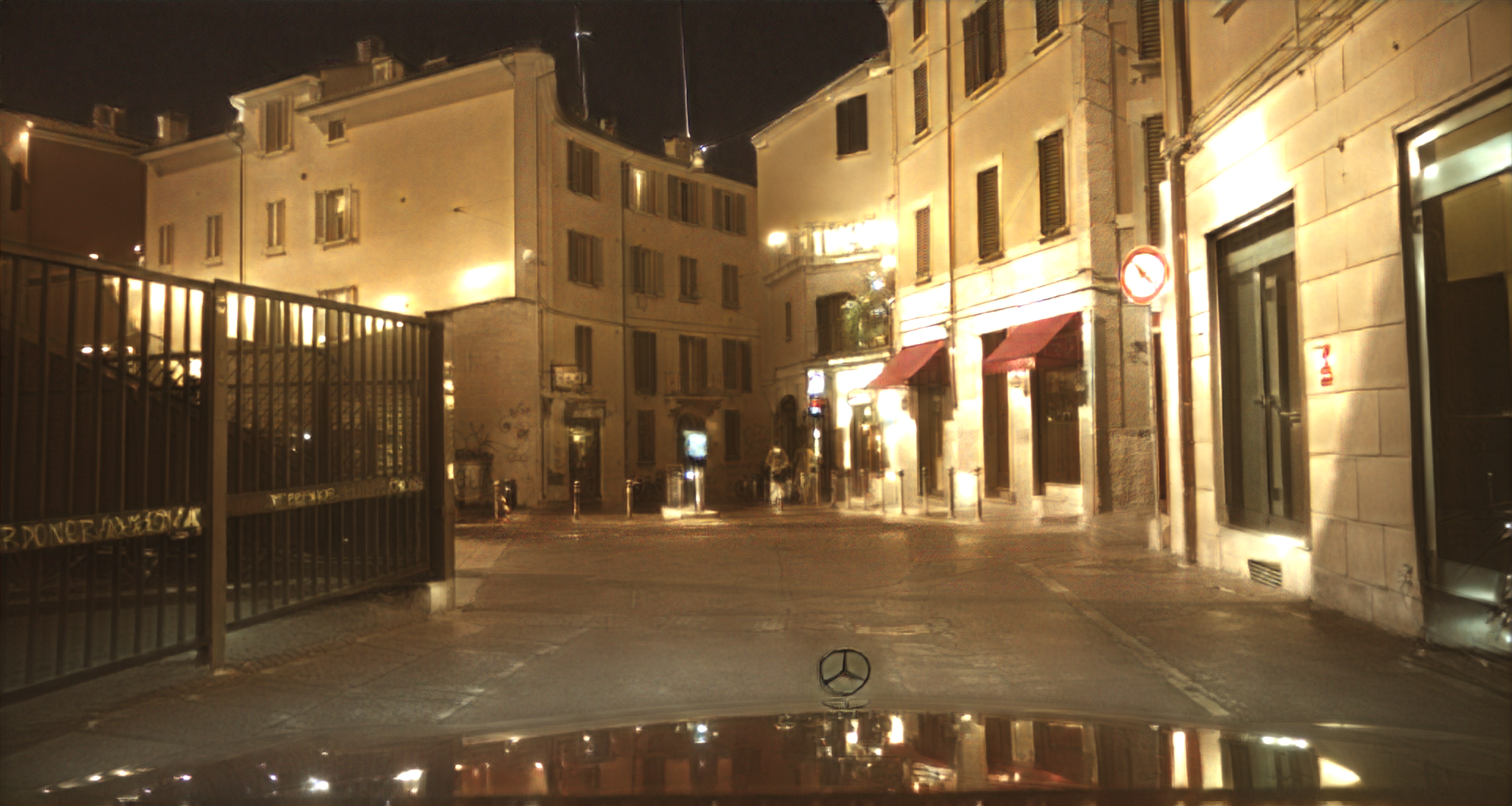}
      \end{minipage}
      \hspace{0.0002\textwidth}
      \begin{minipage}[t]{0.32\textwidth}
          \centering
          {\footnotesize\textbf{DINOv2 Encoder}}\par\vspace{1mm}
          \includegraphics[width=\linewidth]{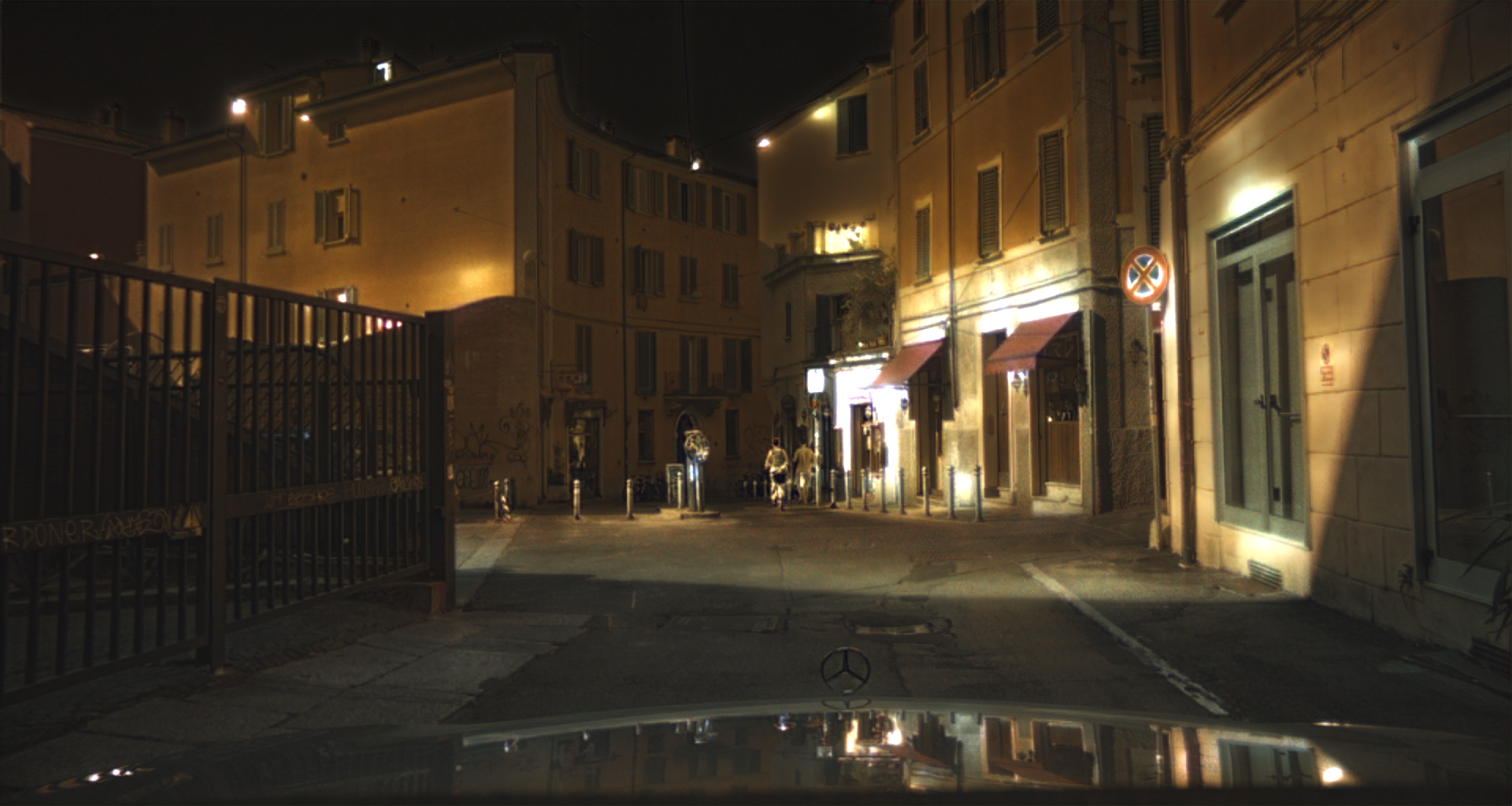}
      \end{minipage}

      \vspace{0.06cm}

      \begin{minipage}[t]{0.32\textwidth}
          \centering
          \includegraphics[width=\linewidth]{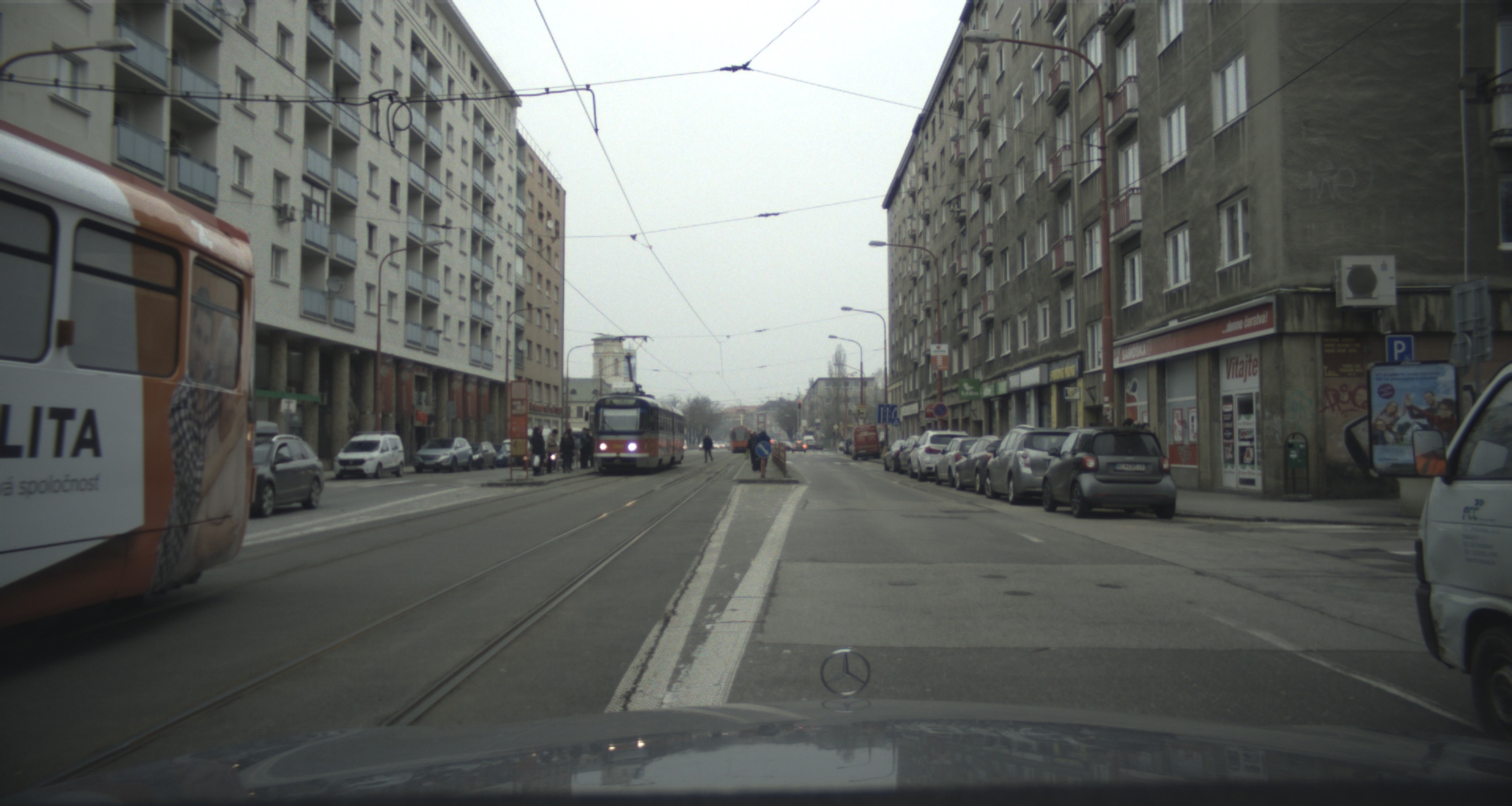}
      \end{minipage}
      \hspace{0.0002\textwidth}
      \begin{minipage}[t]{0.32\textwidth}
          \centering
          \includegraphics[width=\linewidth]{unet_sa.png}
      \end{minipage}
      \hspace{0.0002\textwidth}
      \begin{minipage}[t]{0.32\textwidth}
          \centering
          \includegraphics[width=\linewidth]{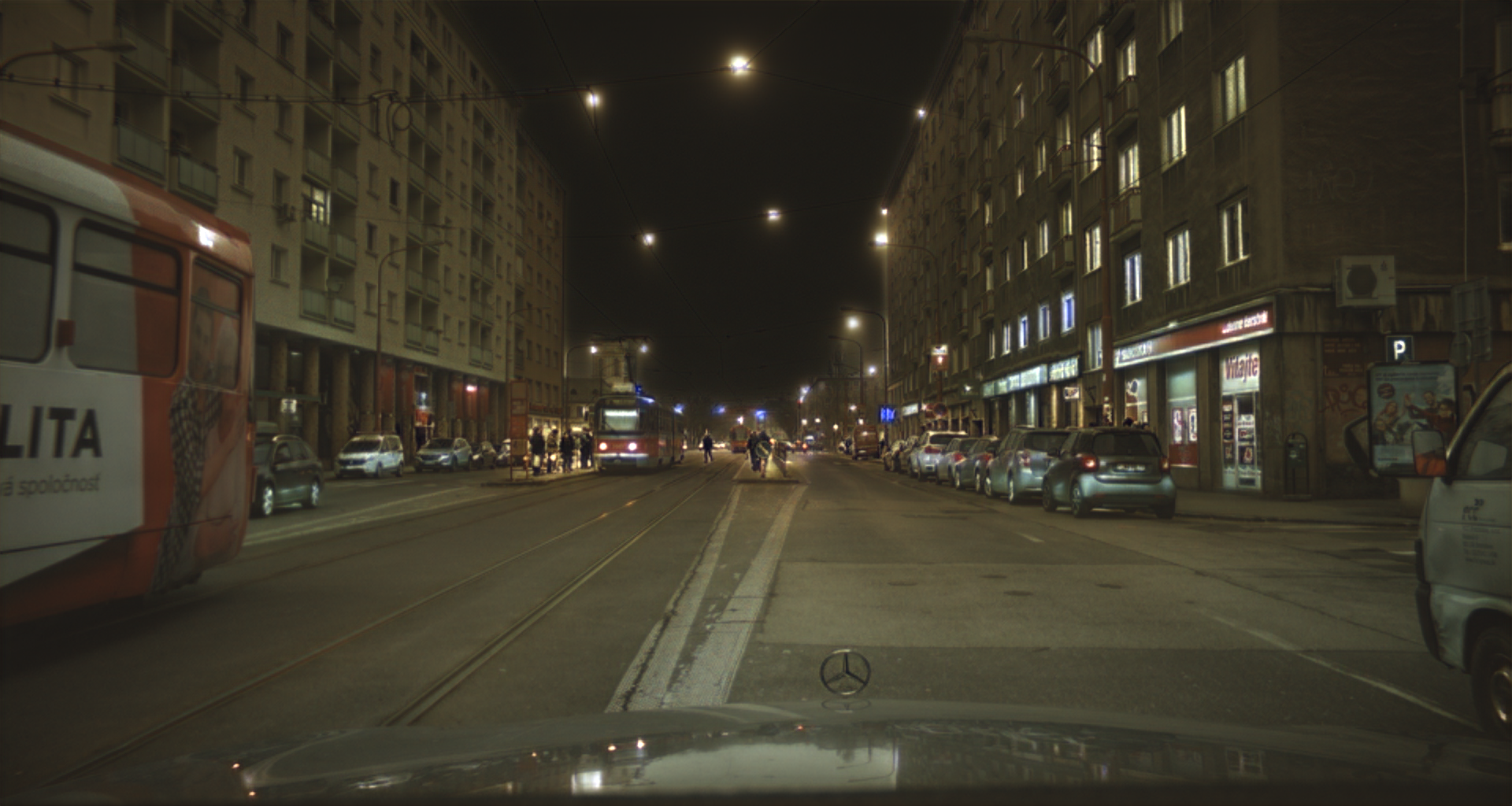}
      \end{minipage}

      \vspace{0.06cm}

      \begin{minipage}[t]{0.32\textwidth}
          \centering
          \includegraphics[width=\linewidth]{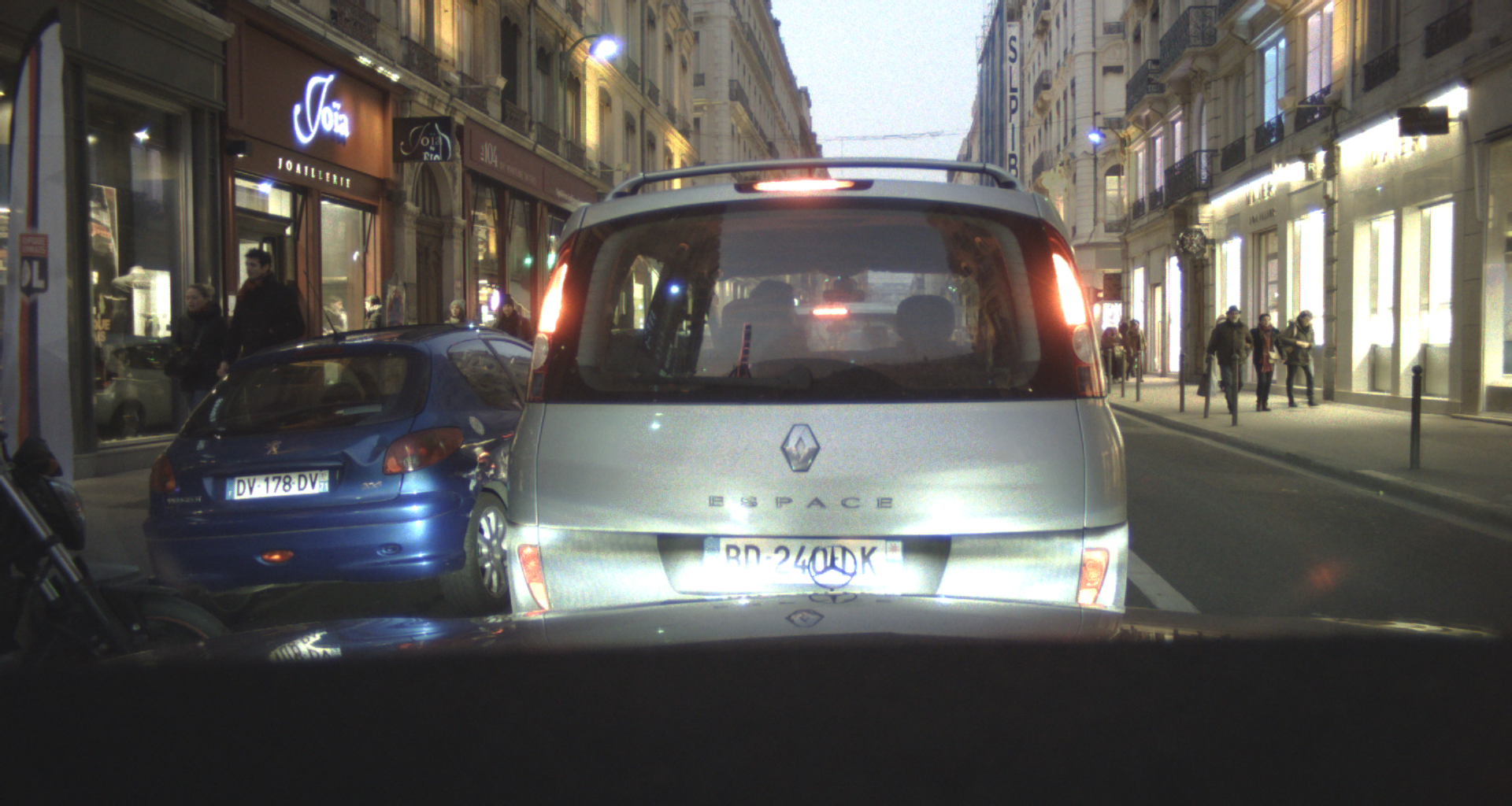}
      \end{minipage}
      \hspace{0.0002\textwidth}
      \begin{minipage}[t]{0.32\textwidth}
          \centering
          \includegraphics[width=\linewidth]{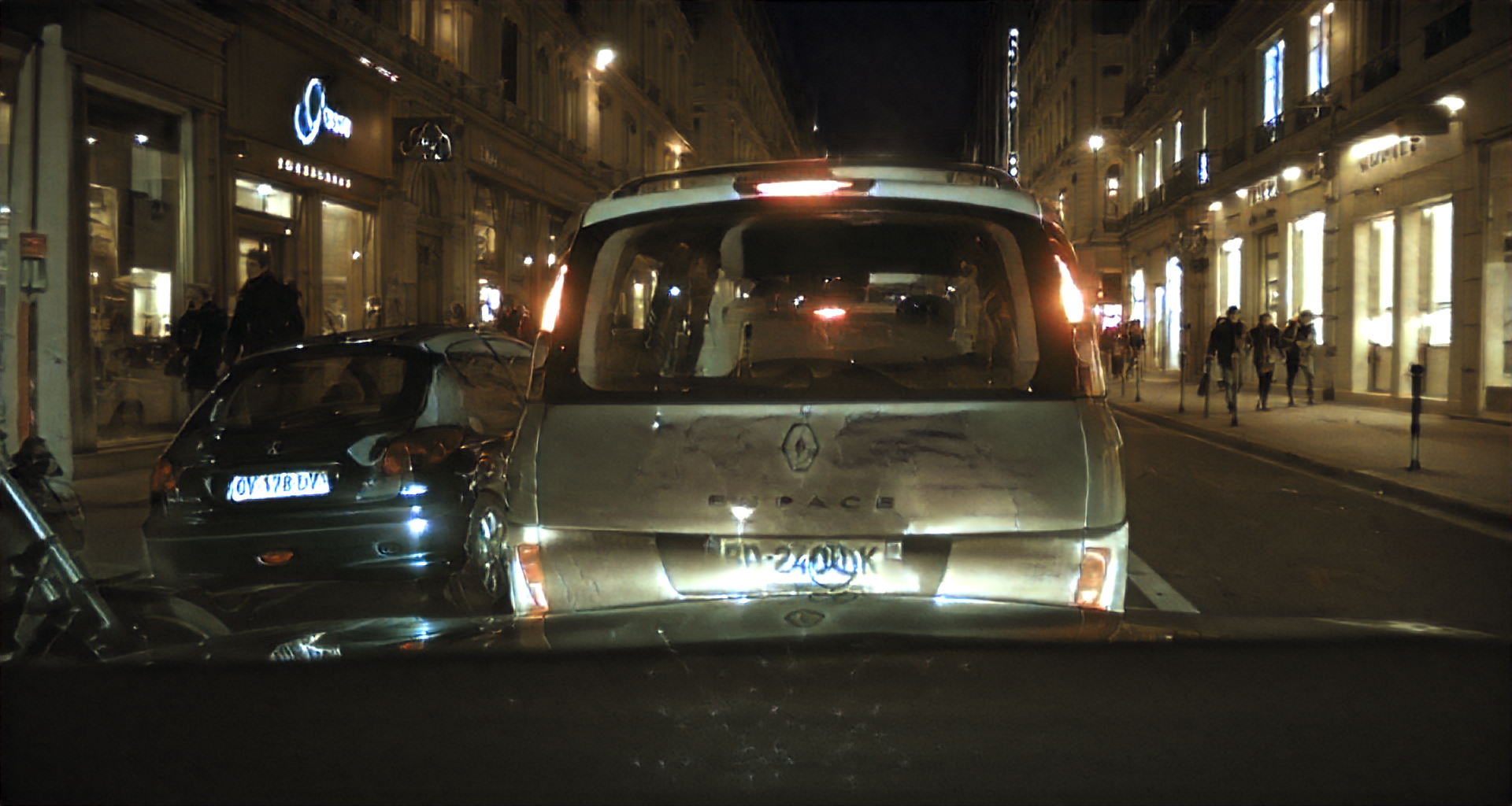}
      \end{minipage}
      \hspace{0.0002\textwidth}
      \begin{minipage}[t]{0.32\textwidth}
          \centering
          \includegraphics[width=\linewidth]{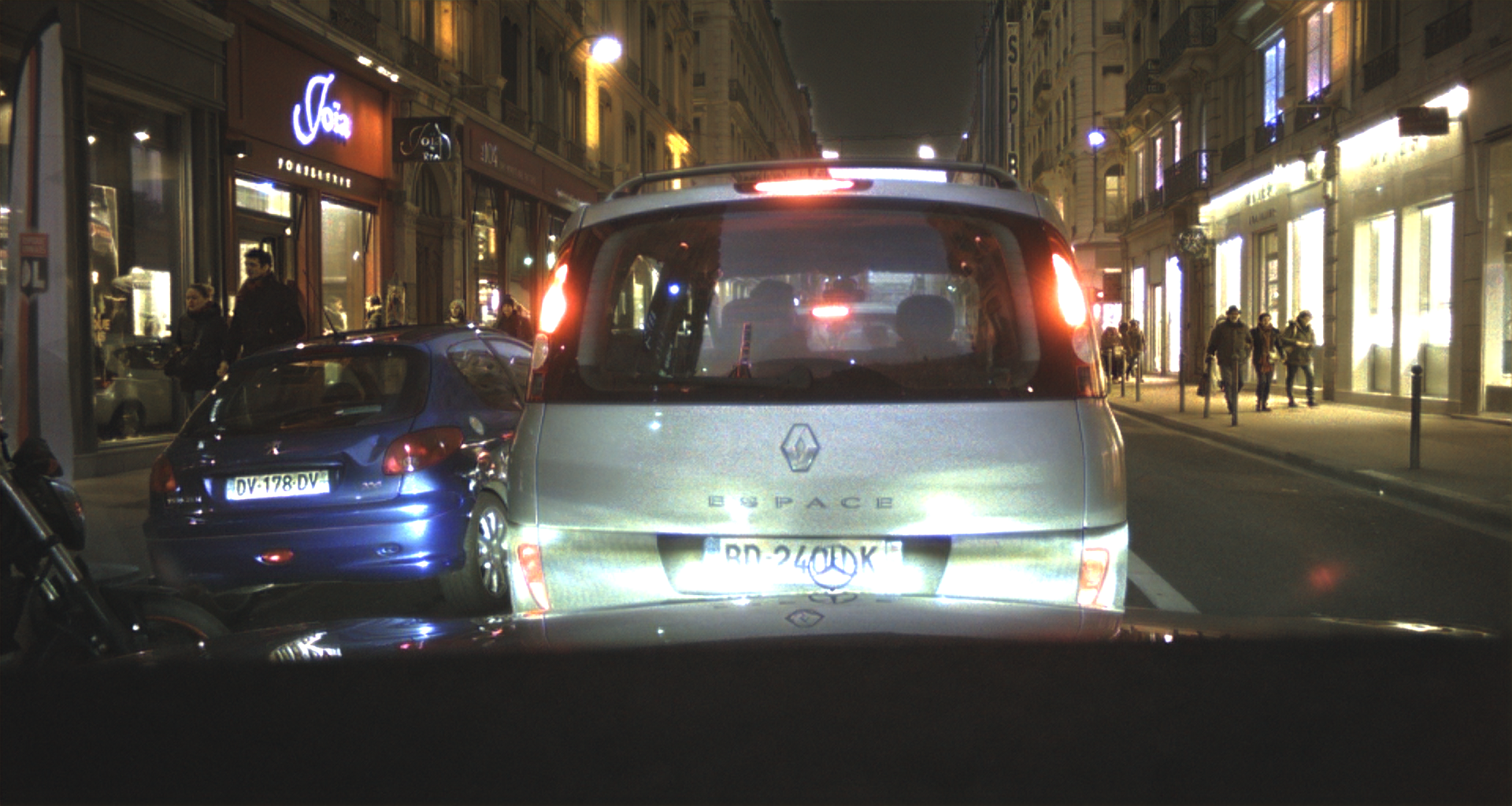}
      \end{minipage}

      \caption{Comparison of night-time images generated using Contrastive-SDXL with U-Net encoder features (middle column) versus DINOv2 features (right column). The left column shows the original daytime images. Using DINOv2 features leads to better preservation of pedestrian details and overall image quality.}
      \label{fig:unet_vs_dinov2}
    \end{figure}

\subsection{Contrastive-SDXL Training Details}
We follow a training setup similar to that of \citet{jung2022exploring}, with modifications for the diffusion-based generator, DINOv2-based patch-wise contrastive learning, and the detector-guided object consistency loss $L_{\text{det}}$. To adapt SDXL-Turbo to the night-time domain, LoRA modules are inserted into the U-Net cross-attention layers, transformer MLP projection layers, convolutional layers, and selected convolutional and attention layers throughout the VAE encoder and decoder.

We first tested patch-wise contrastive loss computation using spatially reshaped self-attention output features from every fourth U-Net encoder transformer layer, capturing intermediate token representations that encode both local and global context. However, using these features resulted in artefacts and weak preservation of fine details, as shown in  Figure~\ref{fig:unet_vs_dinov2}. We therefore use a frozen DINOv2 ViT-g/14 encoder for contrastive supervision. Forward hooks are attached to transformer blocks $34$--$38$, i.e., the five blocks immediately preceding the final block, to extract patch-token representations for the SRC and hDCE objectives. For each selected DINOv2 layer $l \in L$, the extracted hidden states have shape $[B,N_l,C_l]$, where $N_l=H_lW_l+1$ includes the $[\text{CLS}]$ token. We remove the $[\text{CLS}]$ token and reshape the remaining patch embeddings into feature maps $F^{l}\in \mathbb{R}^{B\times C_l\times H_l\times W_l}$.

For DINOv2 feature extraction, both $x_S$ and $\tilde{x}_T$ are resized to $504\times504$. For each selected layer $l\in L$, we sample 128 spatial indices from the translated feature map and use the same indices for the source feature map, ensuring spatially aligned positives. The sampled patch features at each layer are then passed through a layer-specific two-layer MLP projection head $\operatorname{MLP}_{\mathrm{proj}}^l$
\begin{equation}
  \begin{aligned}
f_S^{\,l}(p)&=\mathrm{norm}(\operatorname{MLP}_{\mathrm{proj}}^l(F_S^{\,l}[p])), \\ 
f_T^{\,l}(p)&=\mathrm{norm}(\operatorname{MLP}_{\mathrm{proj}}^l(F_T^{\,l}[p])),
  \end{aligned}
\end{equation}
producing 256-dimensional $\ell_2$-normalised features. Corresponding source and translated locations form positive pairs, while the other sampled source patches within the same image form the negative set. The SRC and hDCE losses are computed at each selected DINOv2 layer and averaged across the five layers. As shown in Figure~\ref{fig:unet_vs_dinov2}, DINOv2 features preserve semantic correspondence and pedestrian details more effectively than the tested U-Net encoder features.

For hard-negative sampling, the source patch self-similarity matrix underlying the SRC relation distribution is also used to weight and derive the negative patches in hDCE, thereby prioritising semantically similar negatives. For the reported configuration, the hard-negative concentration parameter is set to $\gamma=0.5$ and the contrastive temperature parameter to $\tau=0.07$.

For pedestrian consistency, we use a YOLO11 detector fine-tuned offline on a small annotated ECP subset disjoint from the daytime and night-time data used to train or validate the downstream detector. The detector is frozen during Contrastive-SDXL training, and $L_{\text{det}}$ is computed on dense predictions before confidence thresholding and non-maximum suppression, allowing gradients to propagate through the detector to the generator.

During training, the external DINOv2 encoder and the pre-trained U-Net and VAE weights remain frozen, while the U-Net and VAE LoRA adapters, skip connections, projection MLPs, and Vision-Aided discriminator branches are updated. The generator and discriminator are trained using AdamW with a learning rate $10^{-5}$ and a weight decay $0.01$ for 25,000 steps with a batch size 1, on an NVIDIA A100 GPU. Loss weights are set to $\lambda_{\mathrm{det}}=0.5$, $\lambda_{\mathrm{idt}}=0.1$, and $\lambda_{\mathrm{adv}}=0.01$. Weights of the SRC and hDCE losses, $\lambda_{\mathrm{SRC}}$ and $\lambda_{\mathrm{hDCE}}$, are increased linearly from 0.01 to 1 over the first 12,000 iterations to introduce semantic supervision progressively during training.

InstructPix2Pix and CycleGAN-Turbo are used in their default configurations without fine-tuning. Since InstructPix2Pix resizes ECP images from $1920\times1024$ to $576\times320$, bounding boxes are scaled by 0.3 in width and 0.3125 in height. In total, 4,266 ECP daytime validation images are translated by all models. We construct training sets using synthetic images only, and synthetic images mixed with real night-time images equal to $5\%$, $10\%$, and $20\%$ of the synthetic-set size.

The pre-trained Pedestron model is fine-tuned on each augmented set and, for reference, on the full ECP night-time training set, which contains 4,000 images. Fine-tuning is performed for 28 epochs with gradual layer unfreezing every fourth epoch, using SGD with learning rate 0.001, momentum 0.9, and weight decay 0.0001.

The YOLO26 model was fine-tuned using a three-stage strategy with gradual unfreezing. First, the initial 10 layers were frozen and the model was trained for 5 epochs using AdamW with a learning rate of 0.0003 and mosaic augmentation of 0.2. Next, only the first 5 layers were frozen for 15 epochs, with the learning rate reduced to 0.0001 and mosaic augmentation reduced to 0.1. Finally, the full network was unfrozen and trained for 20 epochs at a learning rate of $5\times10^{-5}$, with mosaic augmentation disabled. All stages used an image size of 1280, a batch size of 4 and a weight decay of 0.0005.

\subsection{Evaluation Protocols}

We evaluate the generated data from two complementary perspectives, night-domain distribution alignment and downstream pedestrian detection performance.

Night-domain alignment is quantified using Fréchet Inception Distance (FID)~\citep{heusel2017gans} and empirical 1-Wasserstein distance $W_1$ (WD), computed between synthetic night-time and real ECP night-time images. FID measures the discrepancy between their Inception feature distributions, while WD evaluates the similarity of their latent feature distributions extracted using a trained YOLOv8 model~\citep{yolov8_ultralytics}. Each image is represented by a YOLOv8 feature vector, and all samples are assigned equal weight when forming the two distributions. Pairwise Euclidean distances between real and synthetic feature vectors are used as the transport costs, and WD corresponds to the minimum cost required to align the two distributions. This provides a detection-oriented measure of distributional similarity, complementing the generic Inception representation used by FID. Lower FID and WD values indicate closer agreement with the real night-time distribution.

We additionally use UMAP (Uniform Manifold Approximation and Projection)~\citep{mcinnes2018umap} to visualise the relationship between real daytime, real night-time, and synthetic night-time samples. Images are resized to $240\times128$, represented as flattened RGB pixel vectors, and jointly projected into a two-dimensional embedding. 

Downstream performance is evaluated on the real ECP night-time validation split using the log-average miss rate (LAMR) protocol. Detections are matched to ground-truth annotations at an IoU threshold of $0.5$, and LAMR is computed as the geometric mean of the miss rates over nine false-positive-per-image (FPPI) points logarithmically spaced from $10^{-2}$ to $10^{0}$. Lower values indicate better performance. Results are reported on the ECP \emph{Reasonable}, \emph{Small}, \emph{Occluded}, and \emph{All} subsets. The daytime validation split is additionally used for the daytime-only baselines to quantify the illumination-induced domain gap.

\begin{figure}[!t]
\centering

\setlength{\tabcolsep}{2pt}
\renewcommand{\arraystretch}{1.0}

\begin{tabular}{cccc}

\hline
\scriptsize\textbf{Input} & 
\scriptsize\textbf{InstructPix2Pix} &
\scriptsize\textbf{CycleGAN-Turbo} &
\scriptsize\textbf{Contrastive-SDXL} \\
\hline
\noalign{\vskip 1.8mm} 

\includegraphics[width=0.235\textwidth]{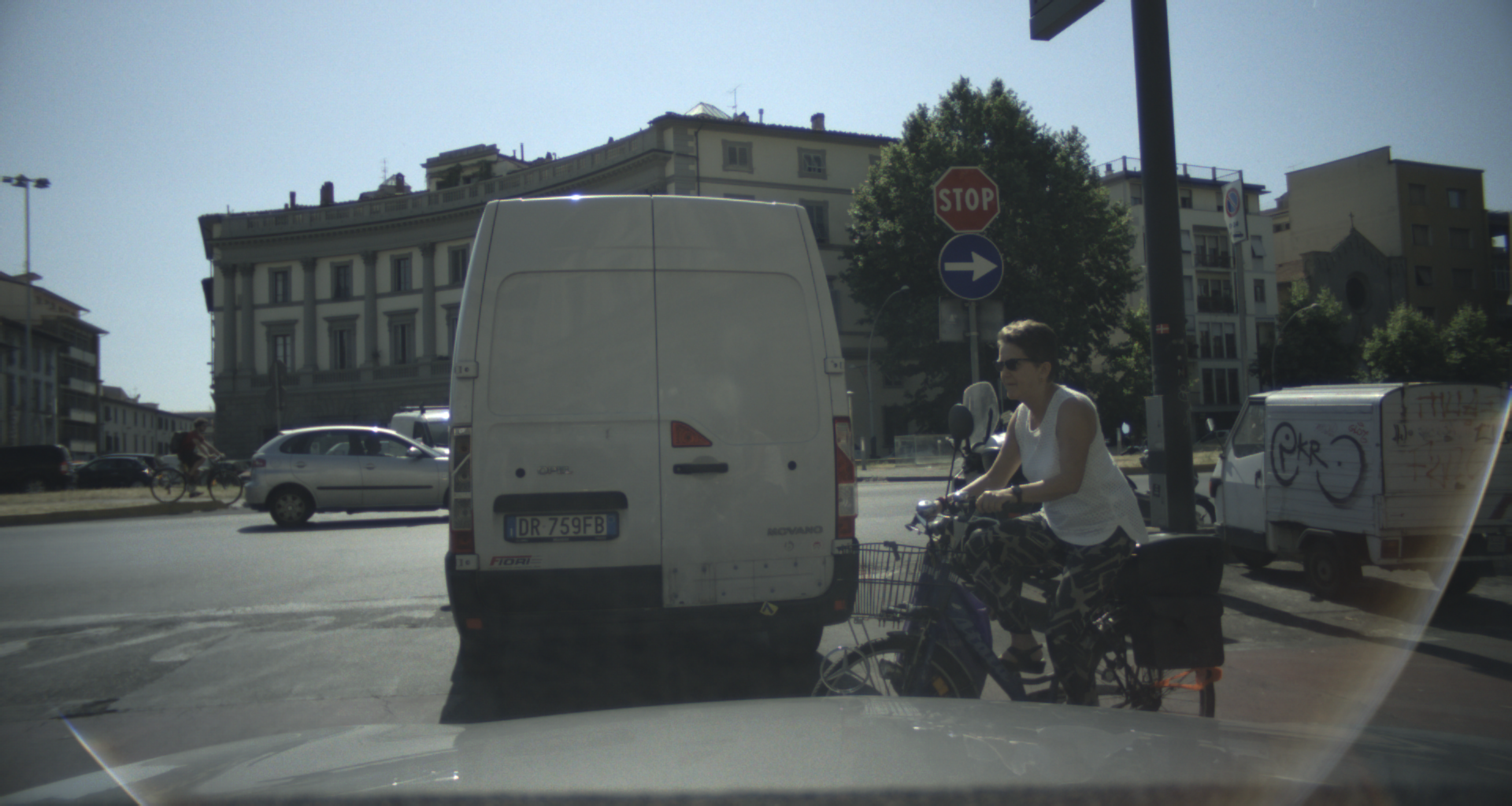} &
\includegraphics[width=0.235\textwidth]{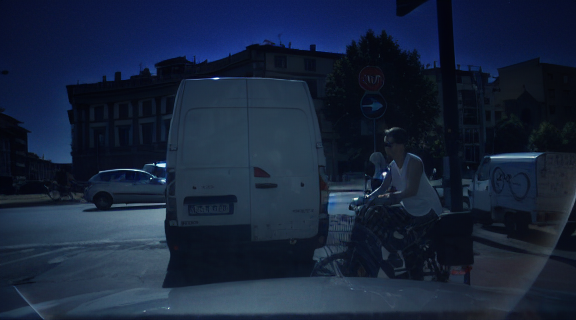} &
\includegraphics[width=0.235\textwidth]{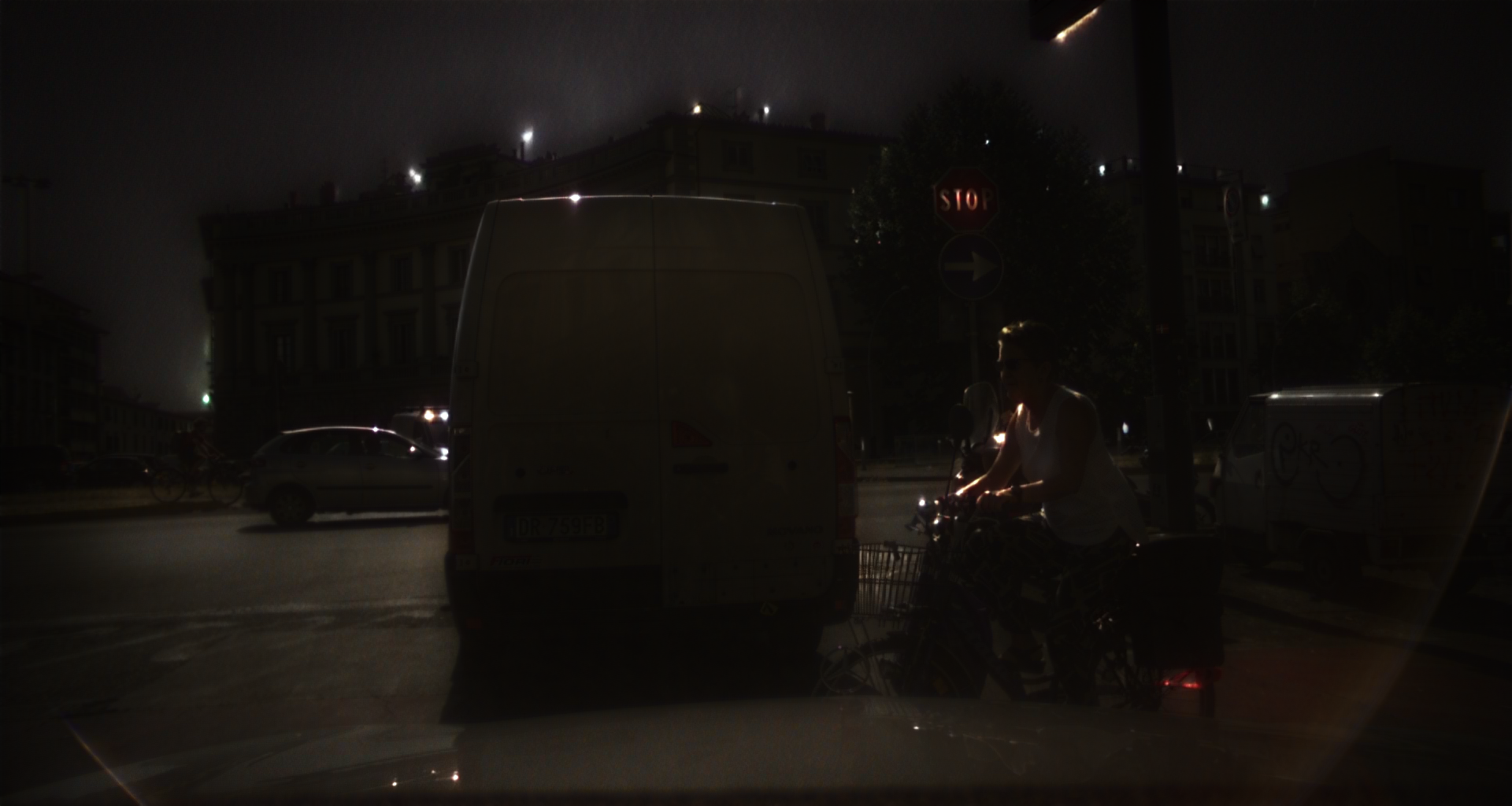} &
\includegraphics[width=0.235\textwidth]{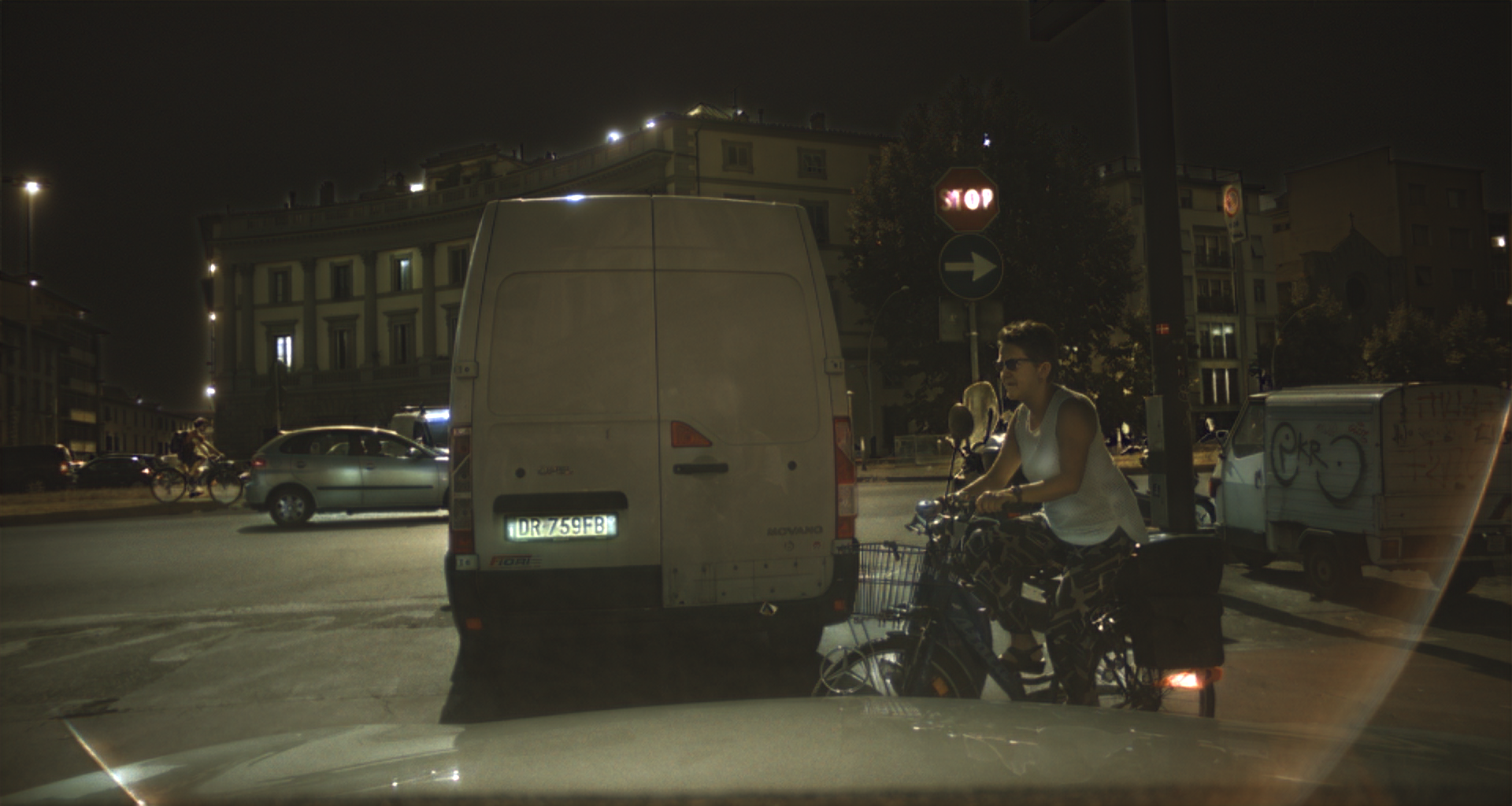} \\

\includegraphics[width=0.235\textwidth]{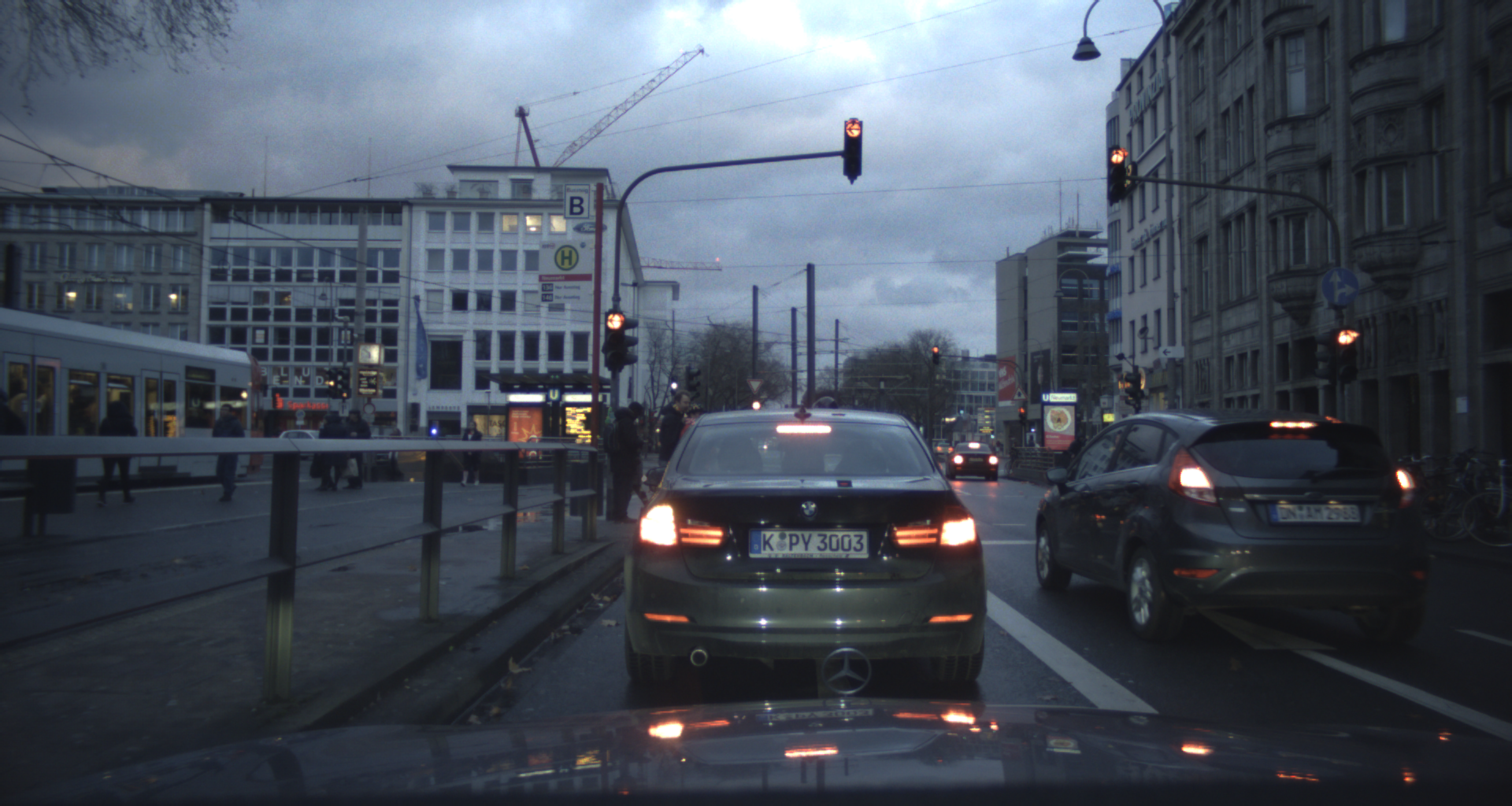} &
\includegraphics[width=0.235\textwidth]{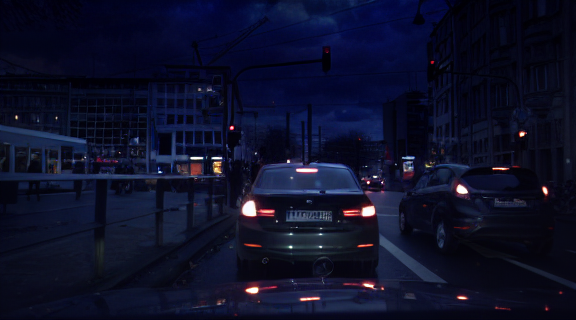} &
\includegraphics[width=0.235\textwidth]{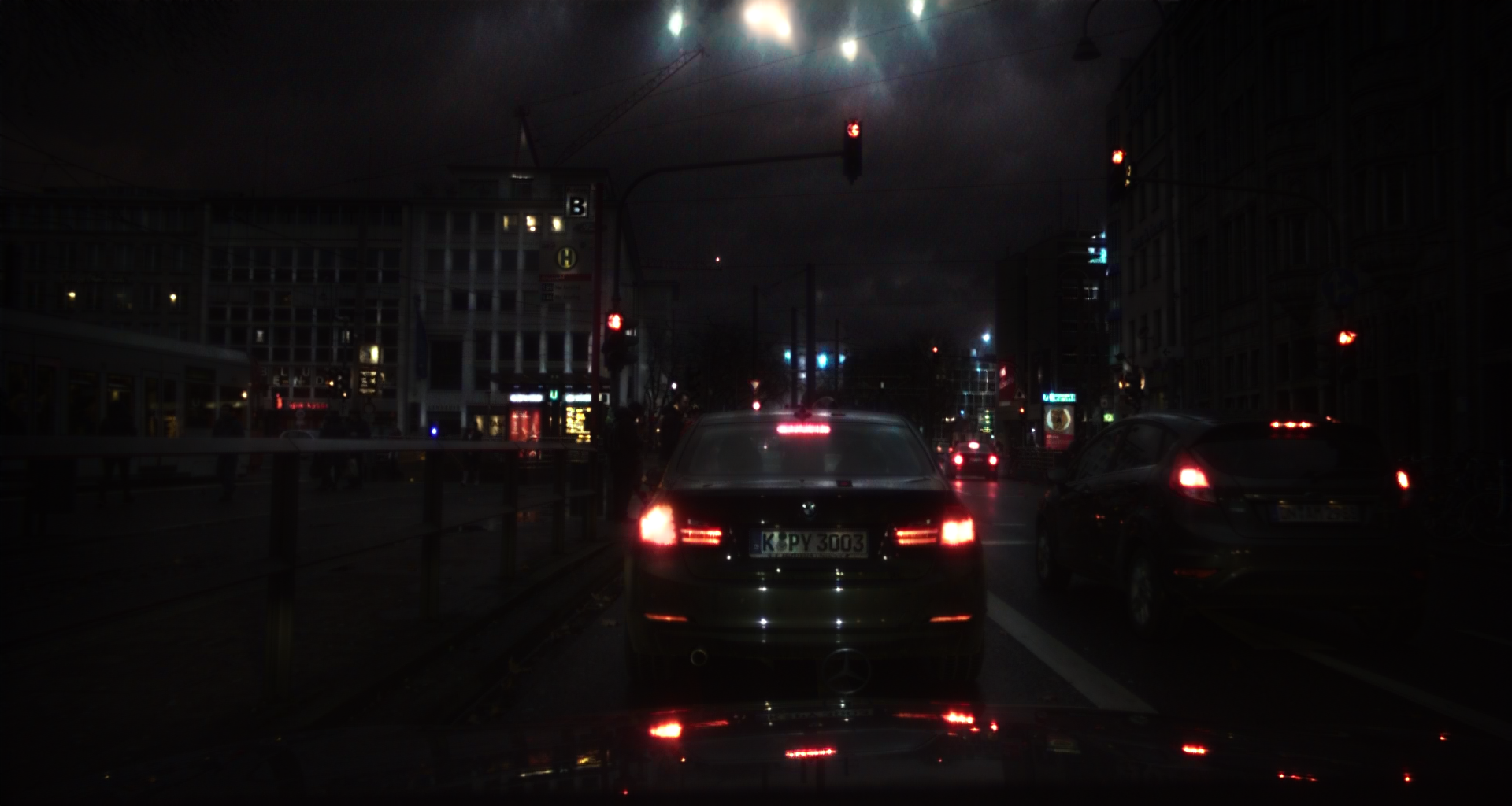} &
\includegraphics[width=0.235\textwidth]{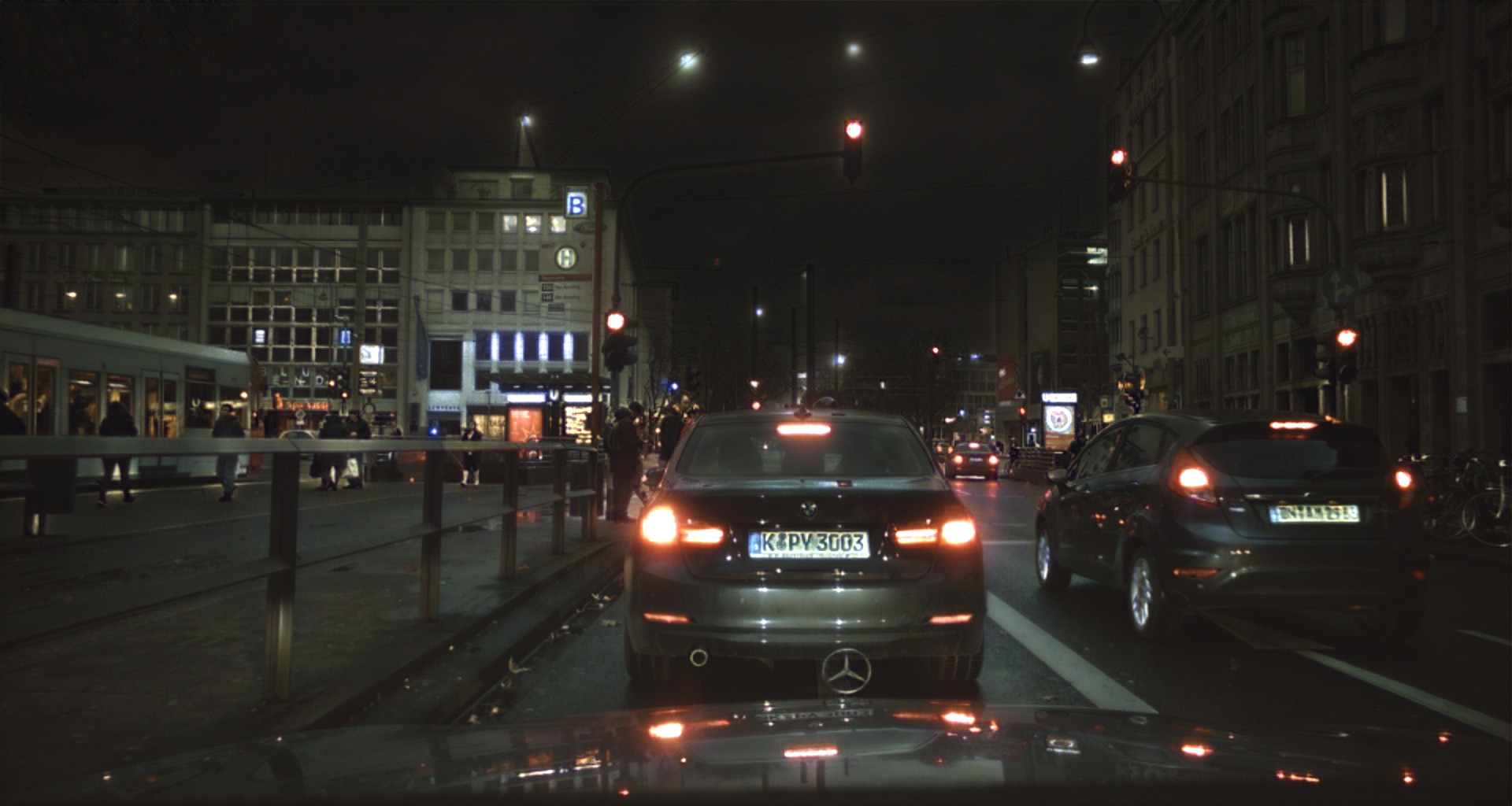} \\

\includegraphics[width=0.235\textwidth]{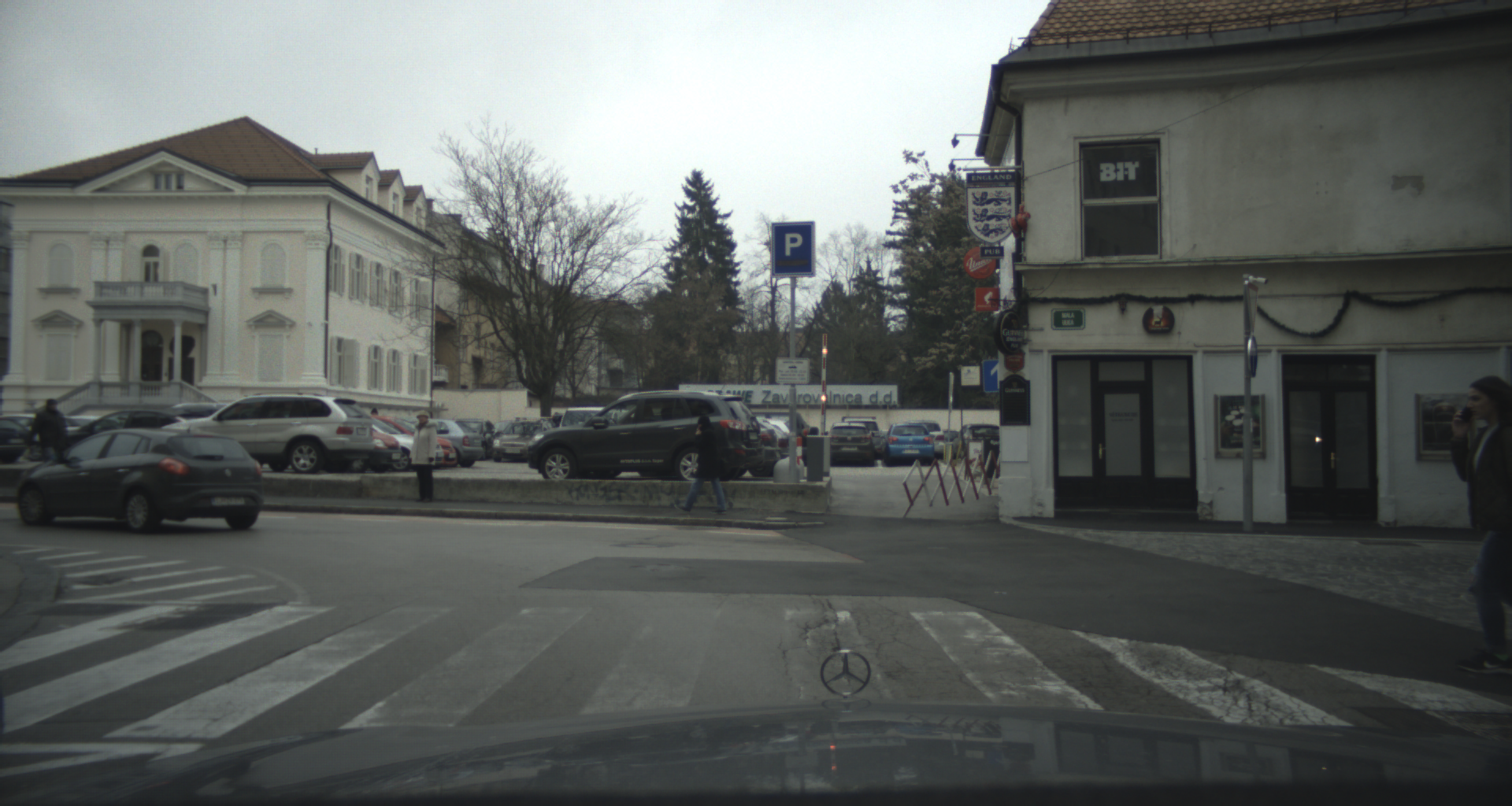} &
\includegraphics[width=0.235\textwidth]{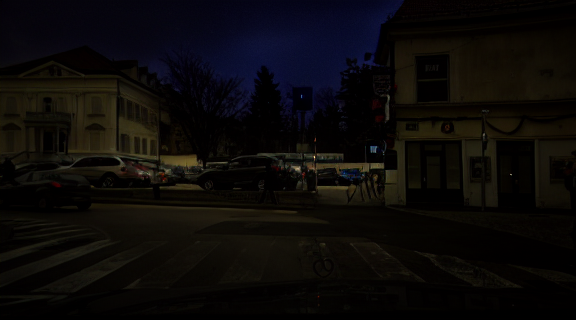} &
\includegraphics[width=0.235\textwidth]{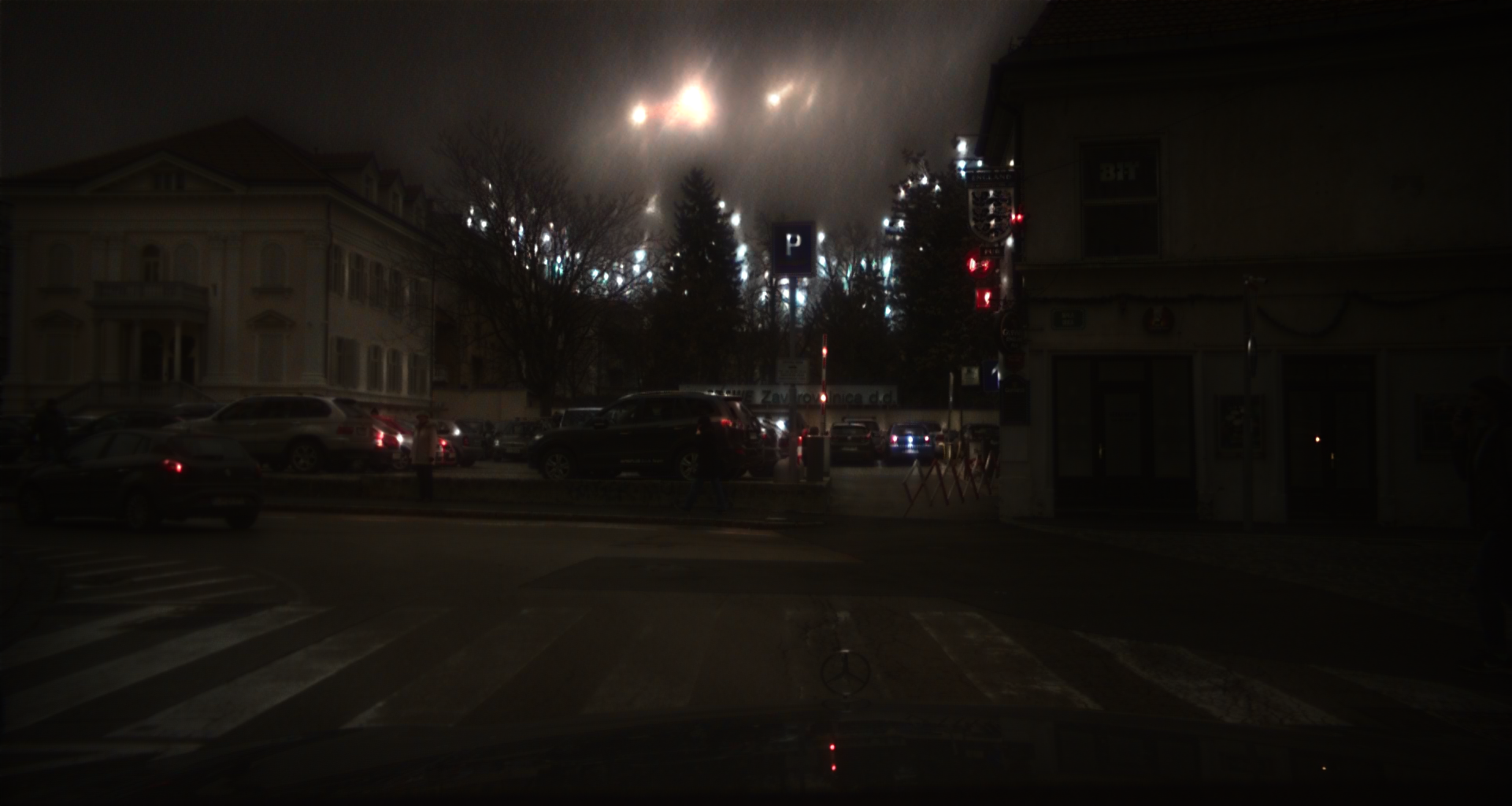} &
\includegraphics[width=0.235\textwidth]{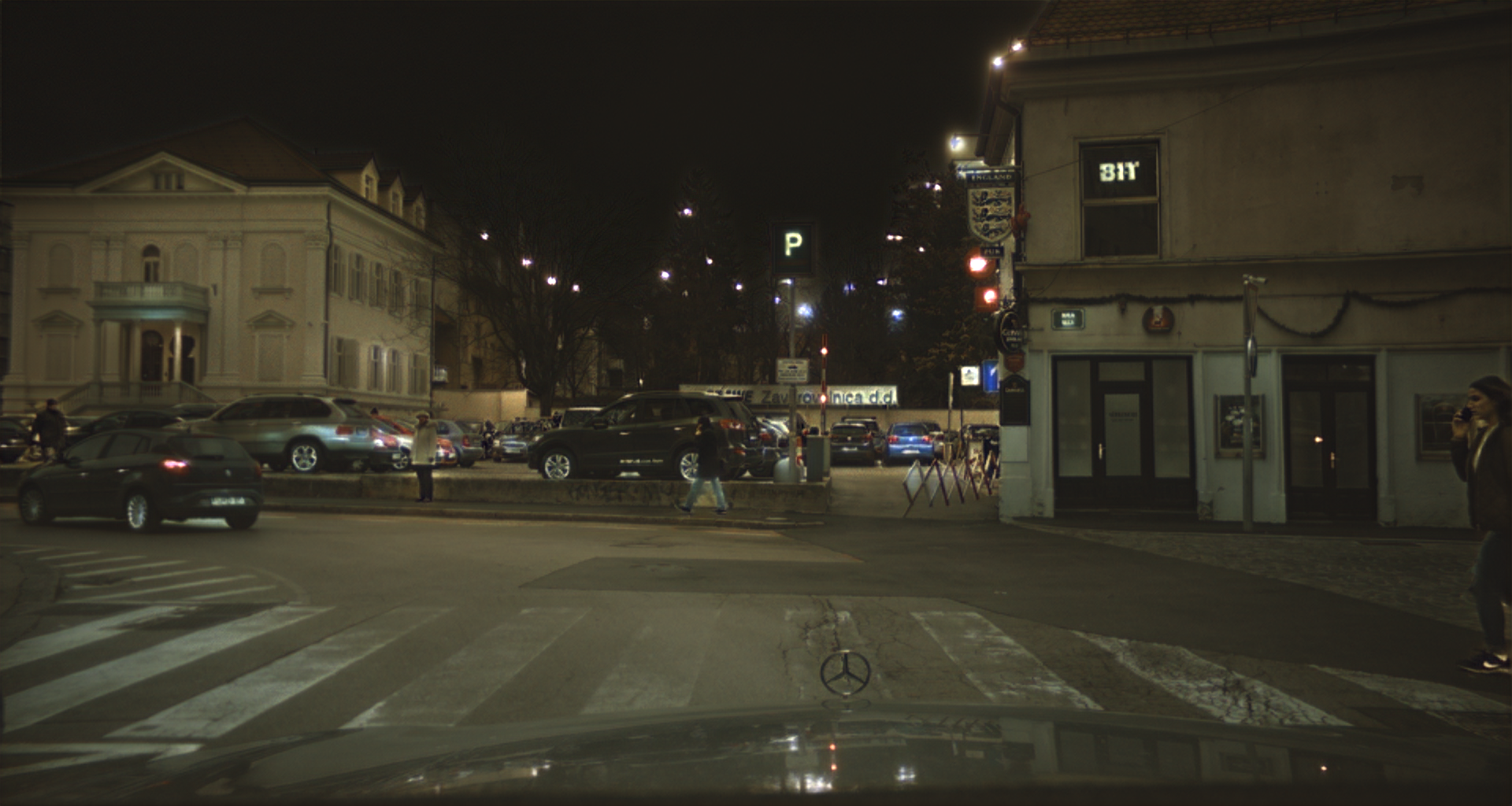} \\

\includegraphics[width=0.235\textwidth]{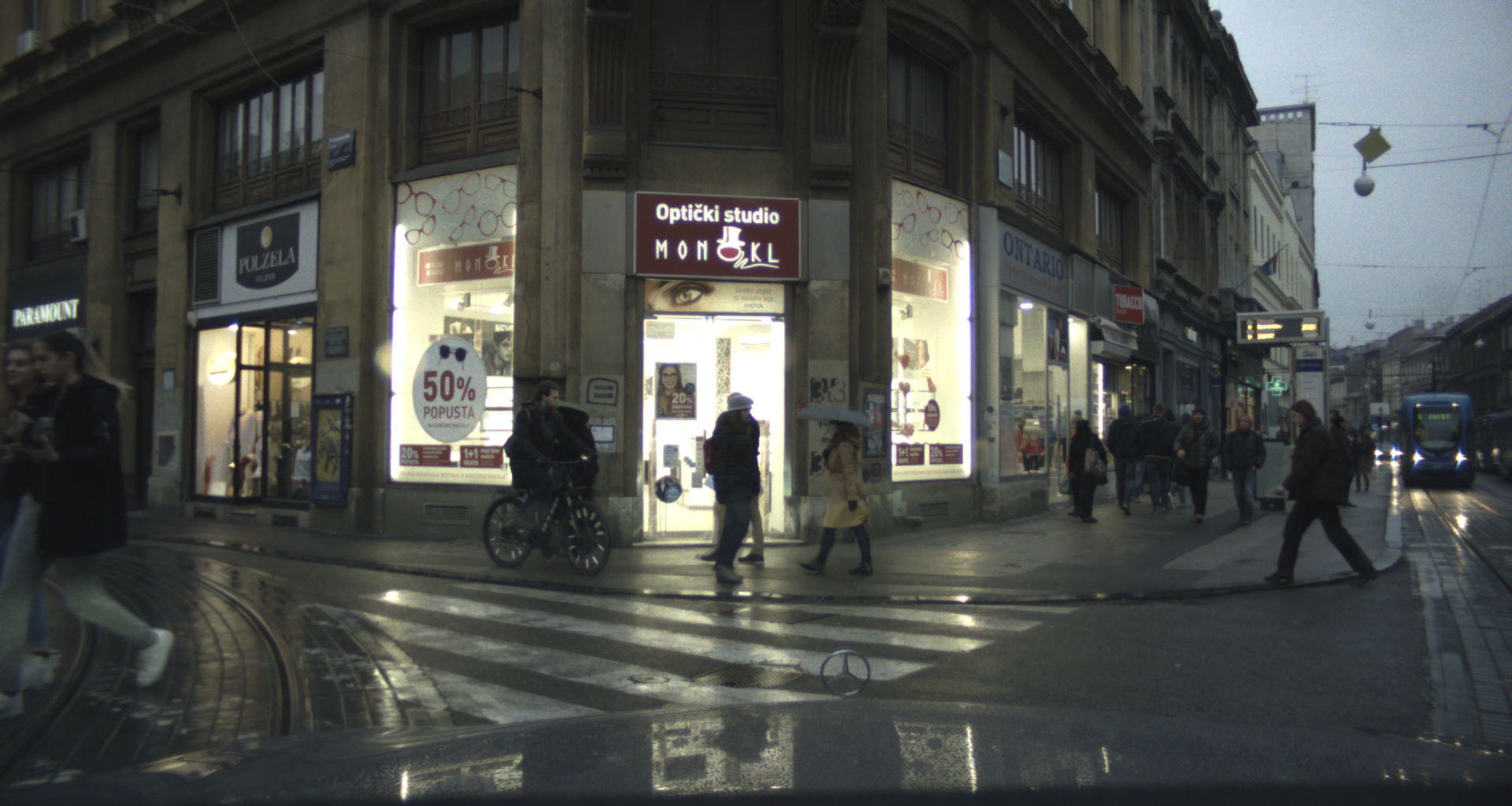} &
\includegraphics[width=0.235\textwidth]{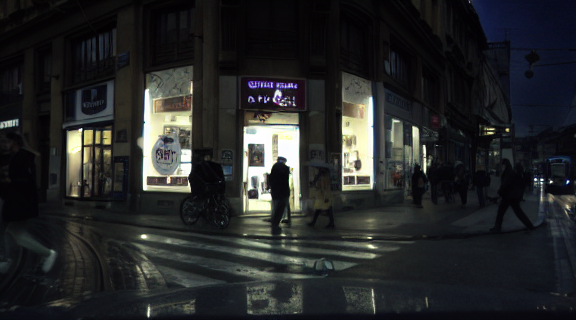} &
\includegraphics[width=0.235\textwidth]{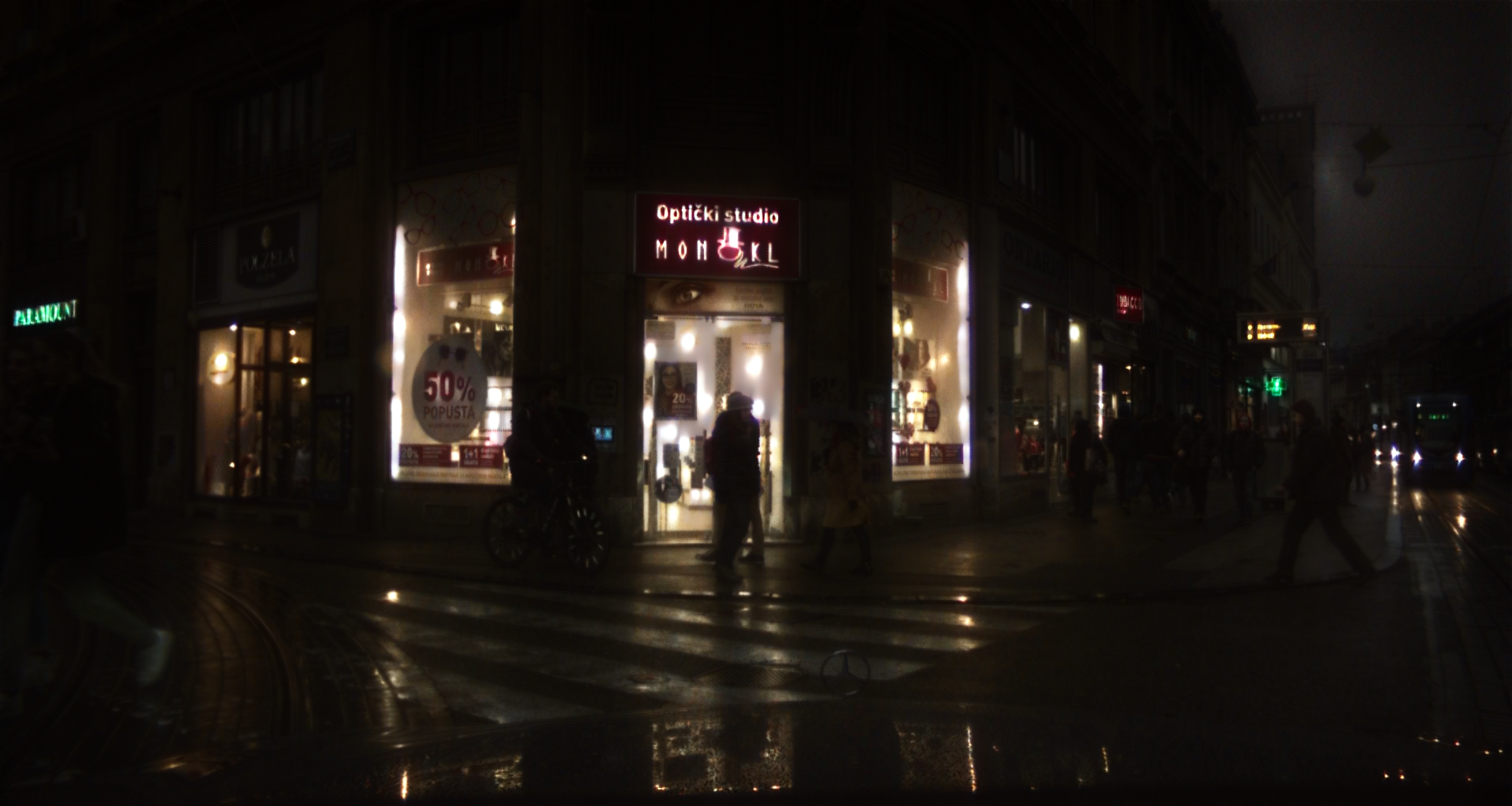} &
\includegraphics[width=0.235\textwidth]{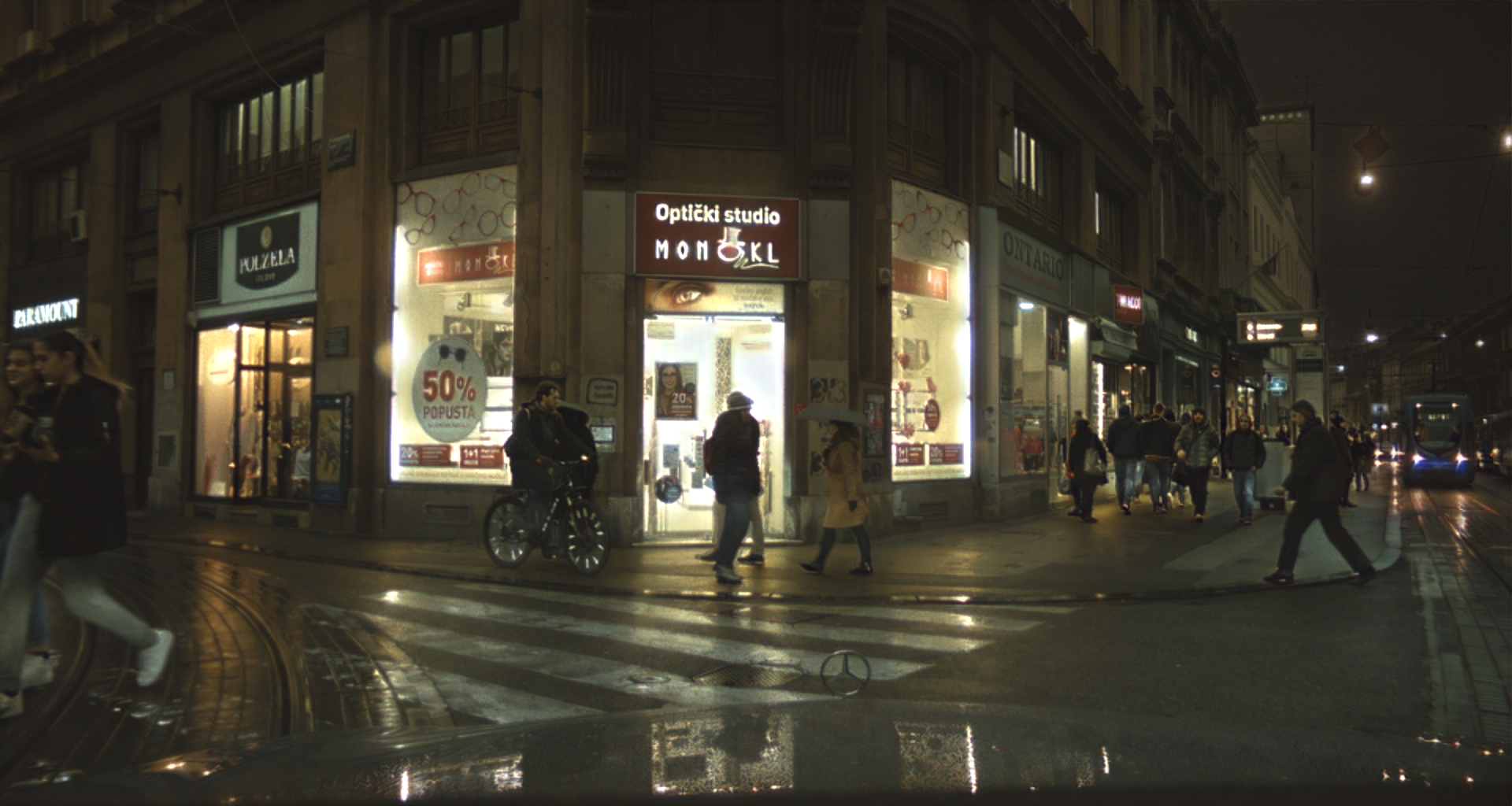} \\

\includegraphics[width=0.235\textwidth]{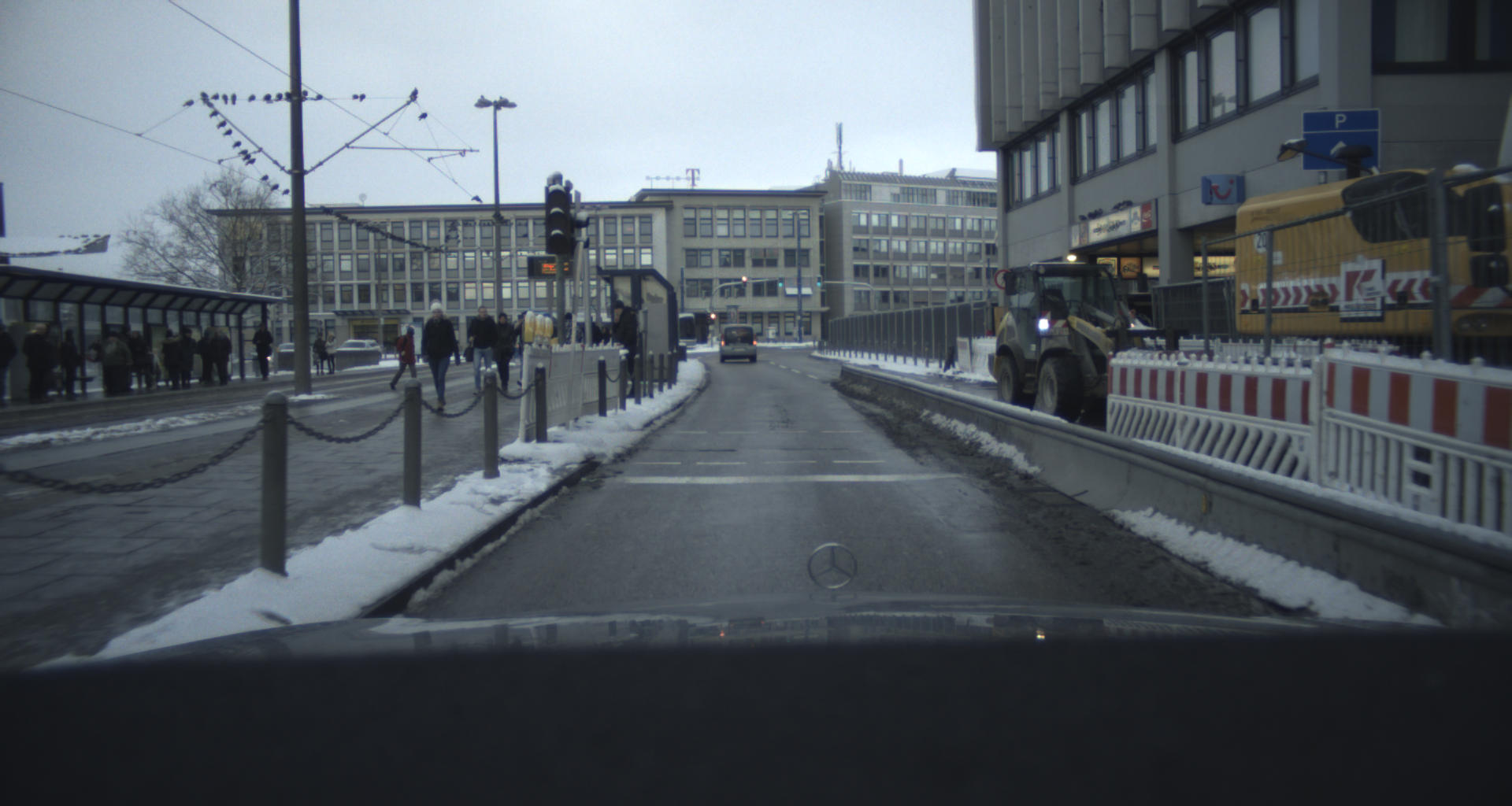} &
\includegraphics[width=0.235\textwidth]{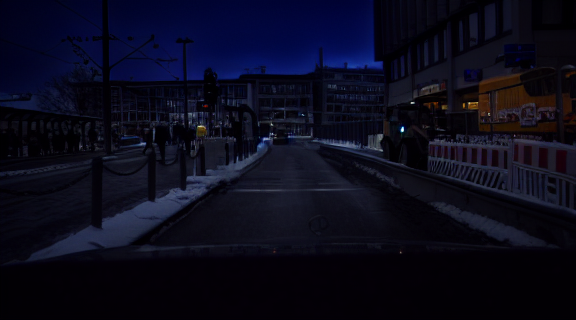} &
\includegraphics[width=0.235\textwidth]{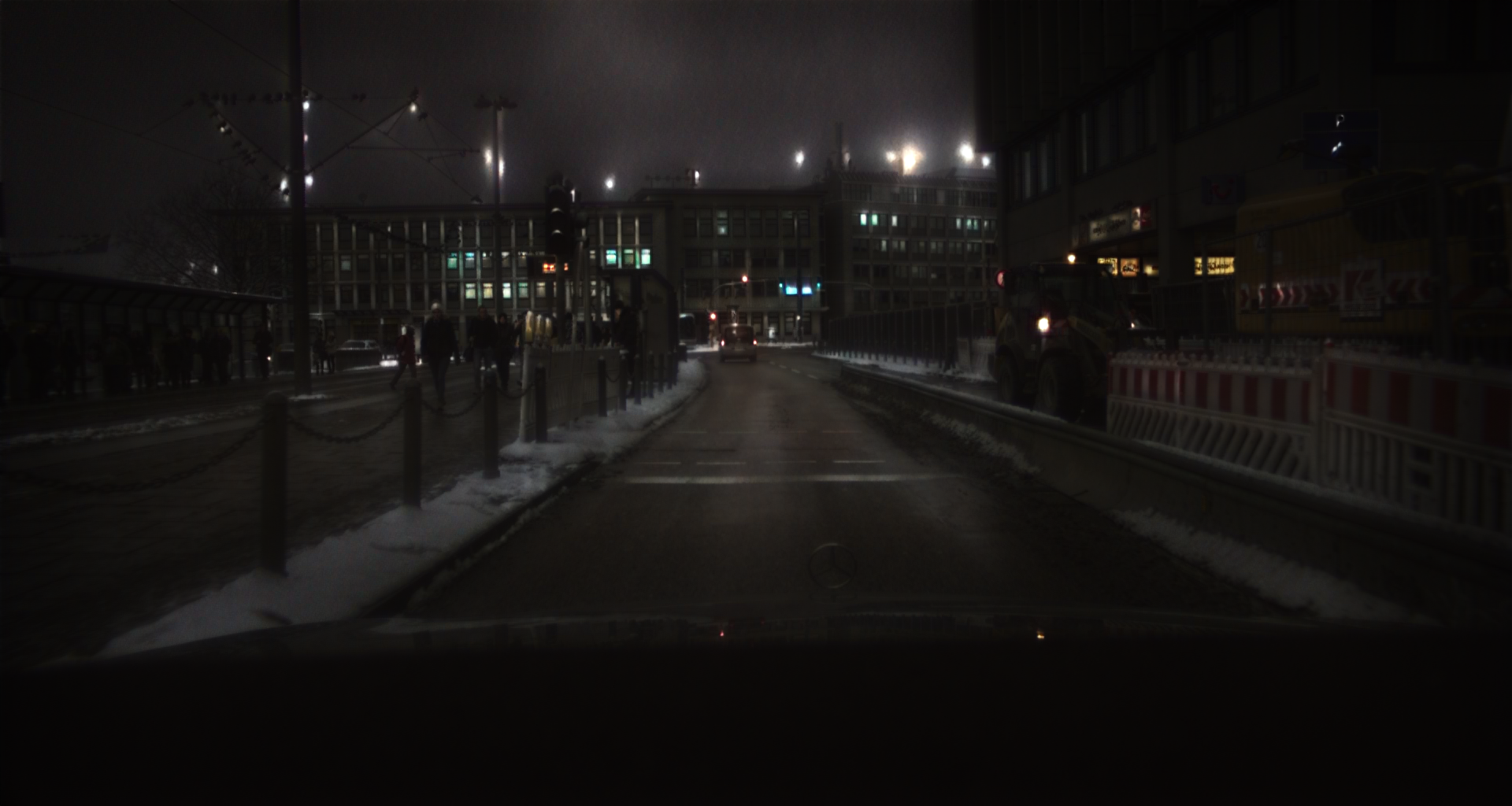} &
\includegraphics[width=0.235\textwidth]{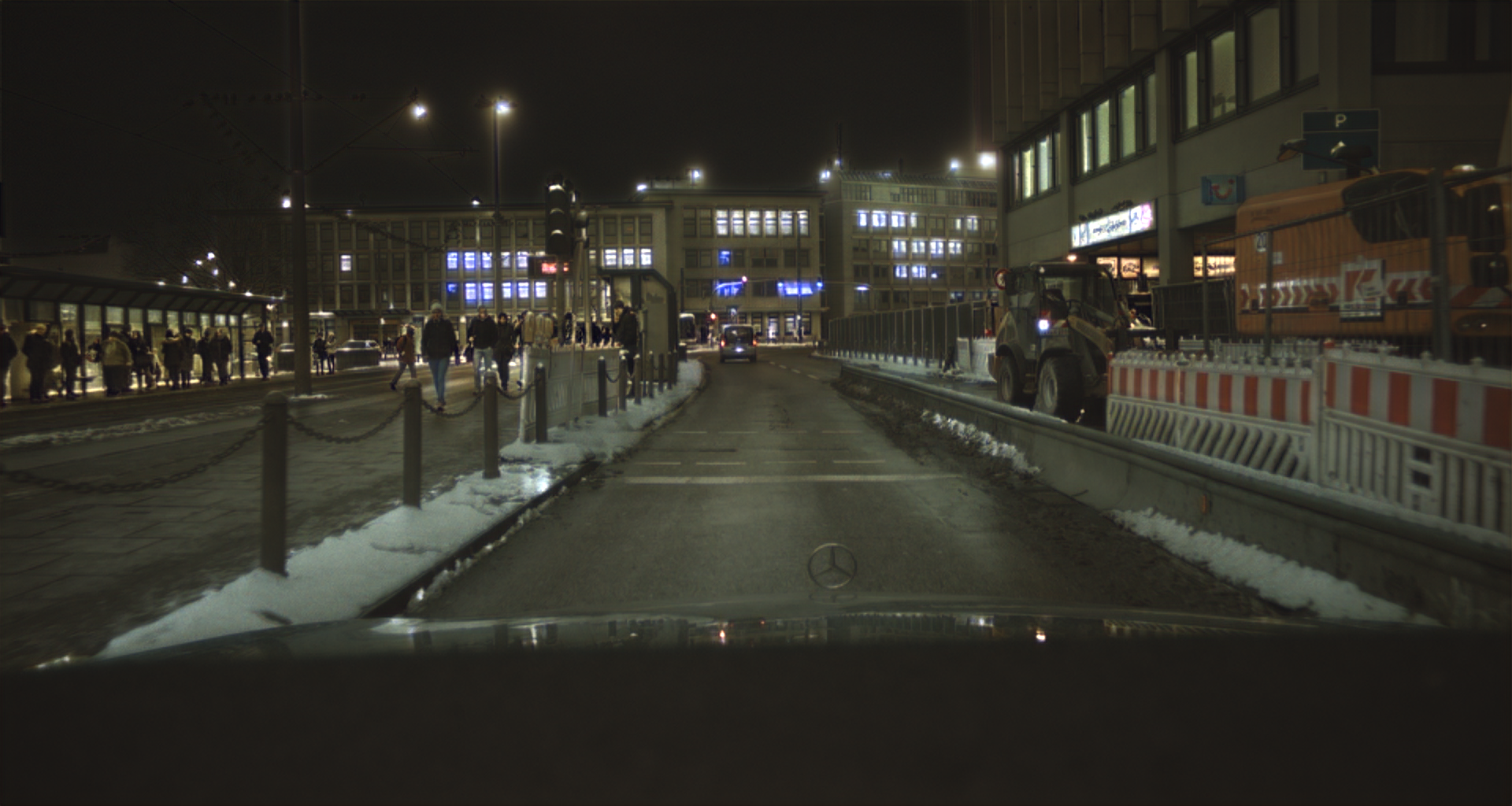} \\

\hline
\end{tabular}

\vspace{2mm}
\caption{Qualitative comparison of synthetic night-time images generated by InstructPix2Pix, CycleGAN-Turbo, and Contrastive-SDXL (Ours). The leftmost column shows the original daytime inputs. CycleGAN-Turbo produces night-like images that are often overly dark and unrealistic and lack pedestrian details, while InstructPix2Pix captures night-time lighting but introduces artefacts and inconsistent pedestrian preservation. In contrast, Contrastive-SDXL generates realistic night-time images while preserving critical pedestrian details.}
\label{fig:qualitative_comparison}

\end{figure}

\begin{table}[htp]
\centering
\caption{FID and Wasserstein distance (WD) between synthetic night-time images from different models and real ECP night-time images. Lower values indicate higher fidelity and closer distributional alignment. Real ECP subsets are included to contextualise the WD values, while FID is not reported for these reference comparisons. Best synthetic results are shown in bold.}

\label{tab:fid_wd}
\small
\setlength{\tabcolsep}{2pt}

\begin{tabular*}{\columnwidth}{@{\extracolsep{\fill}}l c c c@{}}
  \toprule
  \textbf{Model / Dataset}
  & \textbf{Reference Distribution}
  & \multicolumn{2}{c}{\textbf{Day $\rightarrow$ Night}} \\

  \cmidrule(l){3-4}
  & & FID $\downarrow$ & WD $\downarrow$ \\

  \midrule
  InstructPix2Pix
  & \multirow{3}{*}{ECP Night}
  & 64.11 & 31.8 \\

  CycleGAN-Turbo
  &
  & 38.36 & 37.4 \\

  Contrastive-SDXL
  &
  & \textbf{22.57} & \textbf{11.27} \\

  \addlinespace[0.6ex]
  \cdashline{1-4}
  \addlinespace[0.6ex]

  ECP Day
  & ECP Night
  & -- & 34.8 \\

  ECP Night subset 1
  & ECP Night subset 2
  & -- & 5.6 \\

  \bottomrule
\end{tabular*}
\end{table}

\begin{figure}[!t]
\centering
\pgfplotsset{
  umapPtsDay/.style={
    only marks, mark=*,
    mark size=1.35pt,
    mark options={fill=darkblue, opacity=0.82, line width=0.08pt, draw=black}
  },
  umapPtsNight/.style={
    only marks, mark=*,
    mark size=1.35pt,
    mark options={fill=mediumseagreen, opacity=0.82, line width=0.08pt, draw=black}
  },
  umapPtsSyn/.style={
    only marks, mark=*,
    mark size=1.35pt,
    mark options={fill=indianred, opacity=0.82, line width=0.08pt, draw=black}
  },
}

\makebox[\textwidth][l]{%
\begin{subfigure}[t]{0.45\textwidth}
\centering
\begin{tikzpicture}
\begin{axis}[
umapAxis,
legend pos=north west,
]
\addplot[umapPtsSyn]
  table [x index=0, y index=1, col sep=comma] {i2it_2d_3.csv};
\addlegendentry{night-synthetic}

\addplot[umapPtsDay]
  table [x index=0, y index=1, col sep=comma] {i2it_2d_1.csv};
\addlegendentry{day}

\addplot[umapPtsNight]
  table [x index=0, y index=1, col sep=comma] {i2it_2d_2.csv};
\addlegendentry{night}

\end{axis}
\end{tikzpicture}
\subcaption{CycleGAN-Turbo}
\end{subfigure}
\hspace{0.02\textwidth}
\begin{subfigure}[t]{0.45\textwidth}
\centering
\begin{tikzpicture}
\begin{axis}[
umapAxis,
legend pos=north west,
]

\addplot[umapPtsSyn]
  table [x index=0, y index=1, col sep=comma] {ip2p_2d_3.csv};
\addlegendentry{night-synthetic}

\addplot[umapPtsDay]
  table [x index=0, y index=1, col sep=comma] {ip2p_2d_1.csv};
\addlegendentry{day}

\addplot[umapPtsNight]
  table [x index=0, y index=1, col sep=comma] {ip2p_2d_2.csv};
\addlegendentry{night}

\end{axis}
\end{tikzpicture}
\subcaption{InstructPix2Pix}
\end{subfigure}%
}

\vspace{0.5cm}

\begin{subfigure}[t]{0.45\textwidth}
\centering
\begin{tikzpicture}
\begin{axis}[
umapAxis,
legend pos=north west,
]
\addplot[umapPtsSyn]
  table [x index=0, y index=1, col sep=comma] {csv_data_umap1.csv};
\addlegendentry{night-synthetic}

\addplot[umapPtsDay]
  table [x index=0, y index=1, col sep=comma] {csv_data_umap3.csv};
\addlegendentry{day}

\addplot[umapPtsNight]
  table [x index=0, y index=1, col sep=comma] {csv_data_umap2.csv};
\addlegendentry{night}

\end{axis}
\end{tikzpicture}
\subcaption{Contrastive-SDXL}
\end{subfigure}

\caption{Two-dimensional UMAP visualisation of real daytime, real night-time, and synthetic night-time images for three generative models. Each subplot corresponds to a different translation model: (a) CycleGAN-Turbo, (b) InstructPix2Pix, and (c) Contrastive-SDXL. Synthetic night-time images generated by Contrastive-SDXL cluster more closely with real night-time images and are better separated from daytime samples, indicating improved distributional alignment compared to the other models.}
\label{fig:umap}
\end{figure}


\begin{figure}[!t]
\centering
\setlength{\tabcolsep}{1pt}
\renewcommand{\arraystretch}{1.0}
\begin{tabular}{@{}ccc@{}}
{\footnotesize\textbf{Day}} &
{\footnotesize\textbf{Contrastive-SDXL}} &
{\footnotesize\textbf{CycleGAN-Turbo}} \\[3pt]

\raisebox{-0.5\height}{\includegraphics[width=0.325\textwidth]{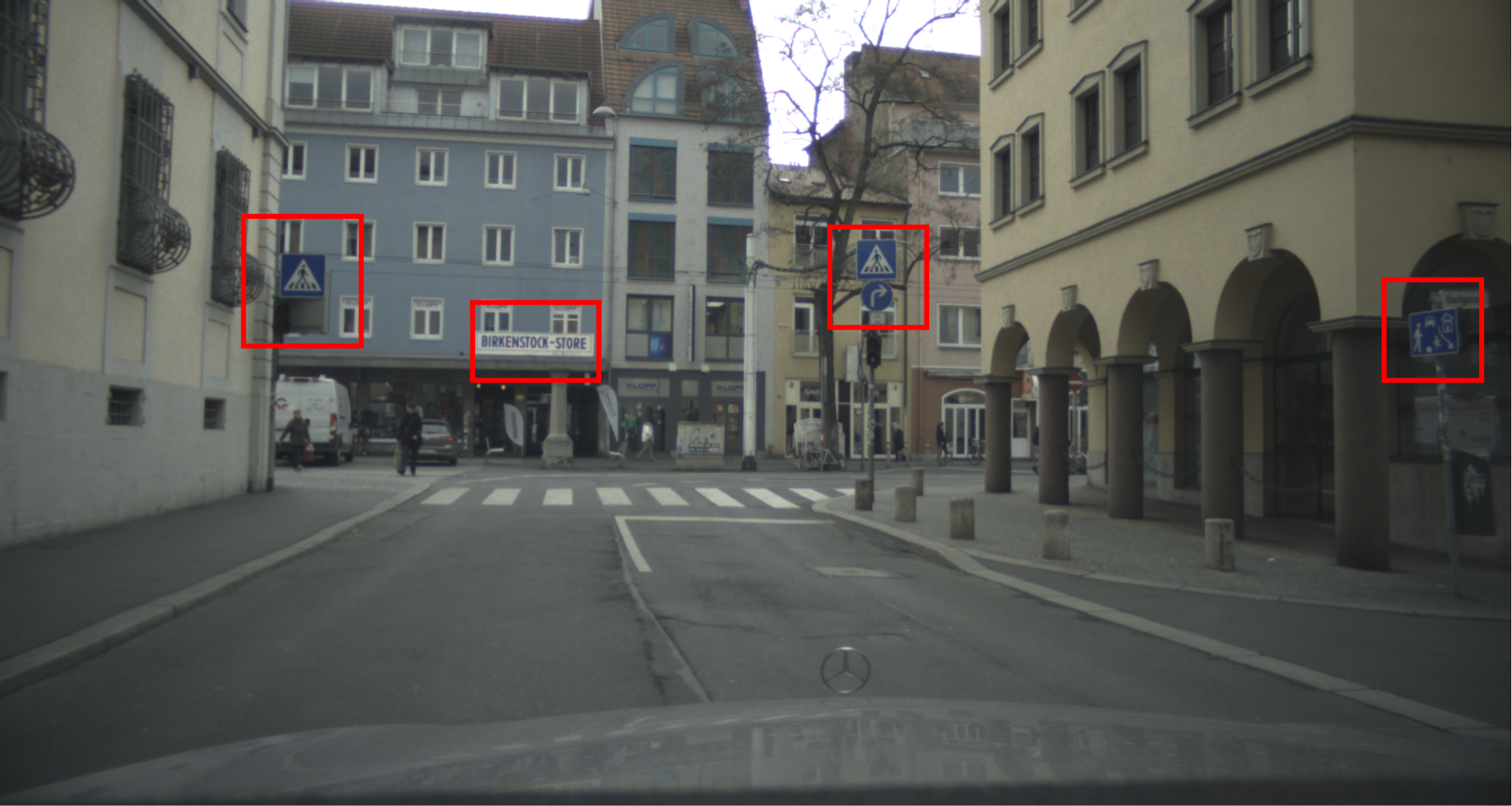}} &
\raisebox{-0.5\height}{\includegraphics[width=0.325\textwidth]{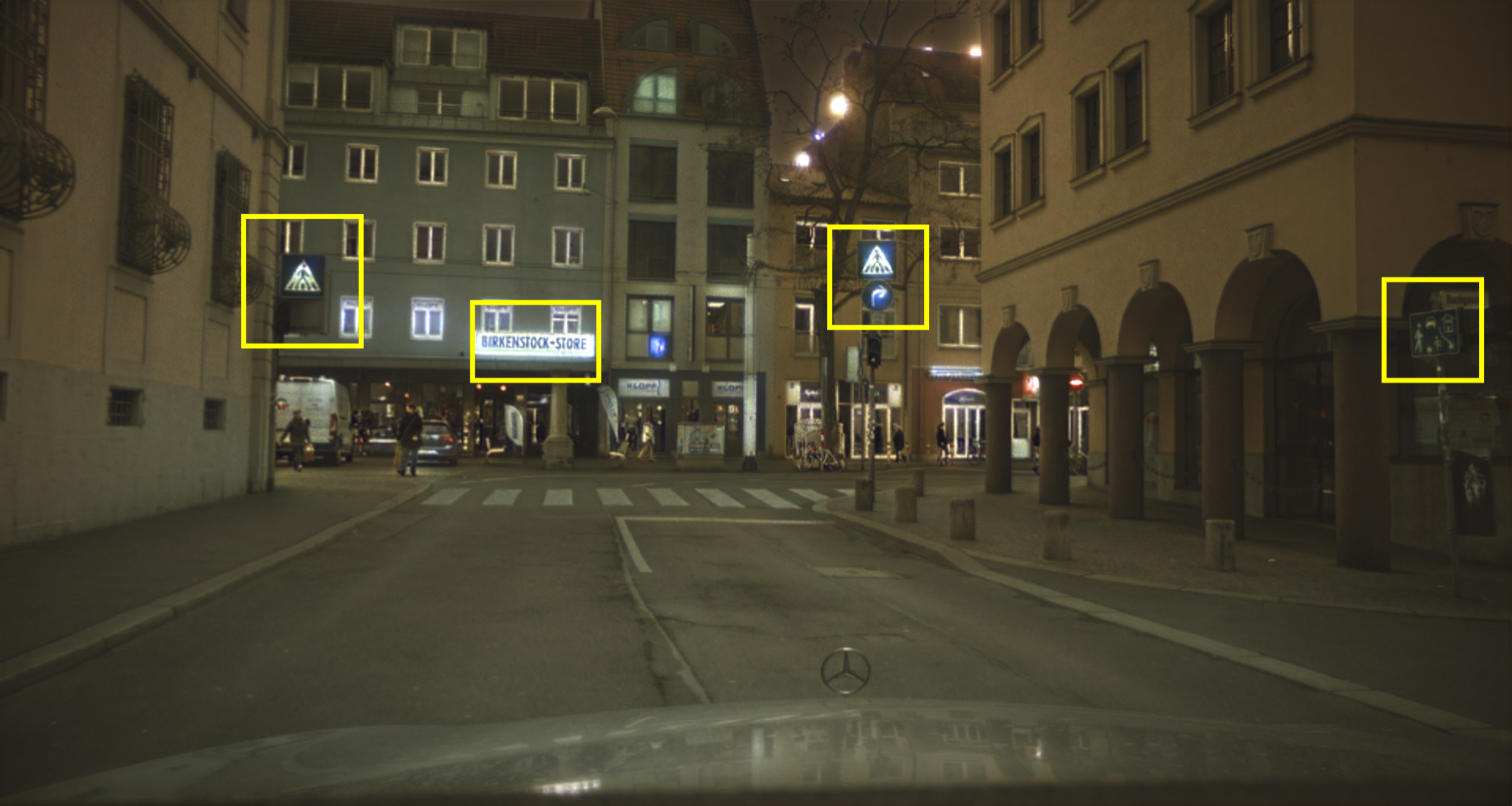}} &
\raisebox{-0.5\height}{\includegraphics[width=0.325\textwidth]{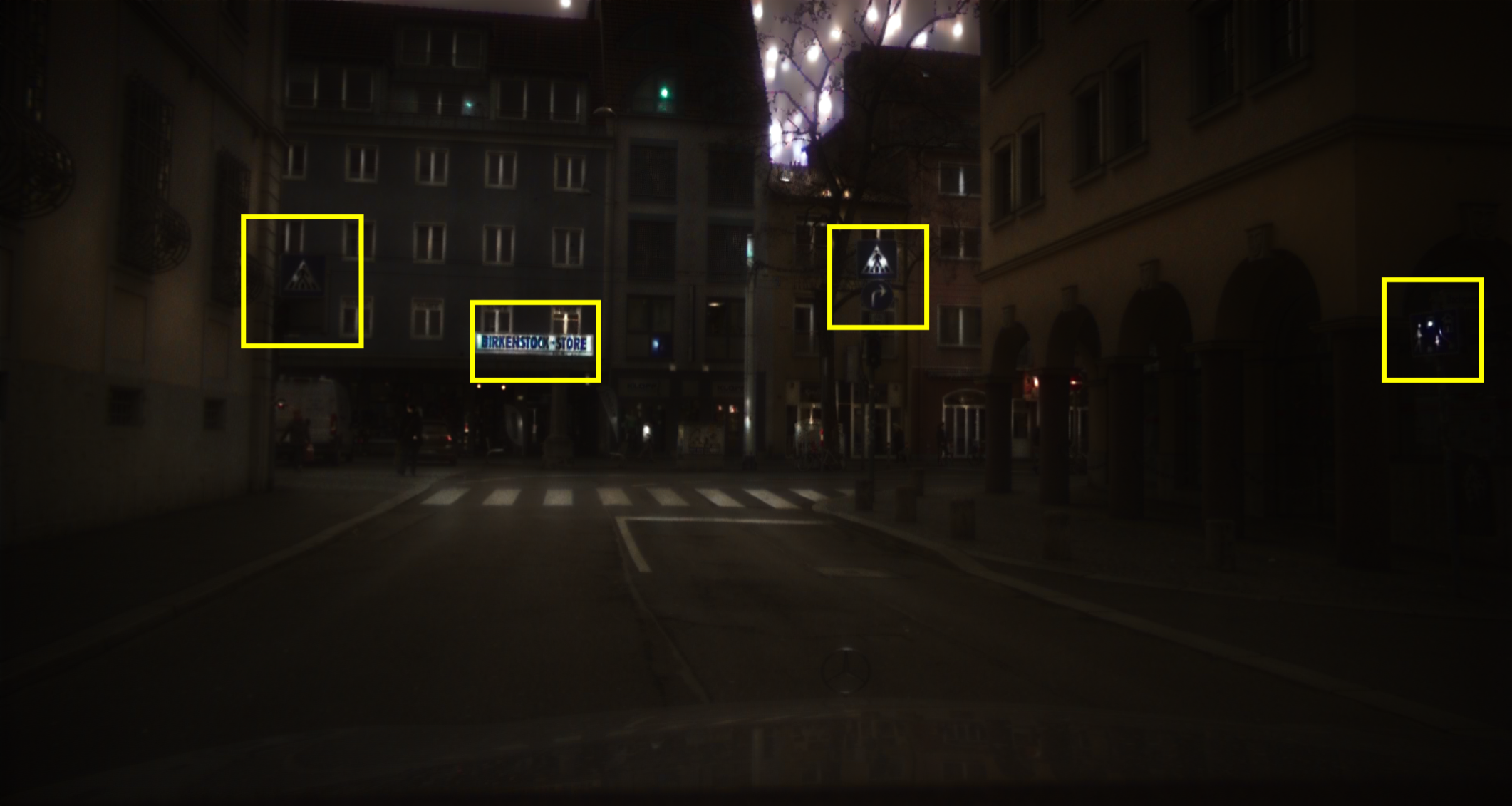}} \\
\noalign{\vskip 3pt}

\raisebox{-0.5\height}{\includegraphics[width=0.325\textwidth]{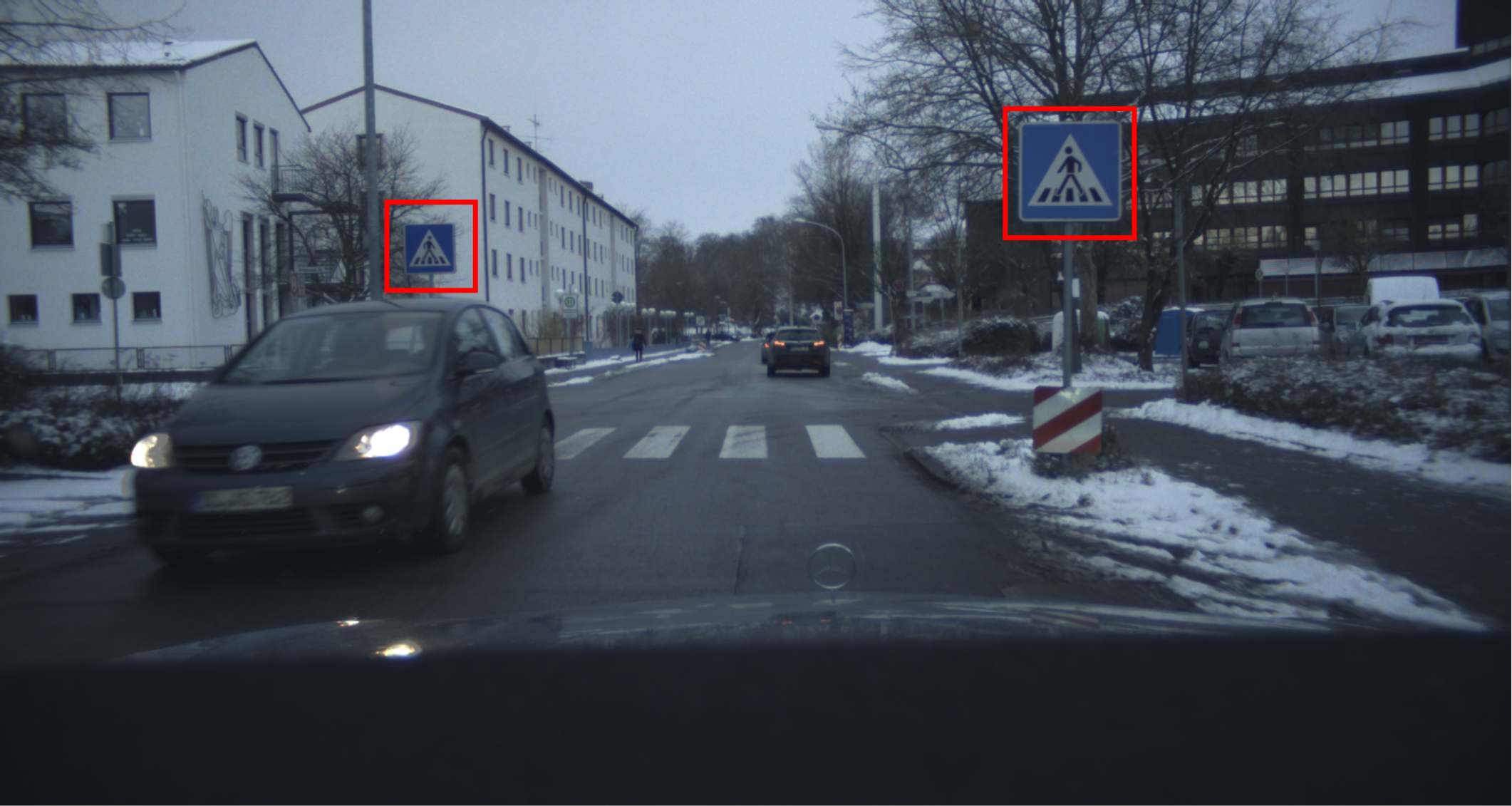}} &
\raisebox{-0.5\height}{\includegraphics[width=0.325\textwidth]{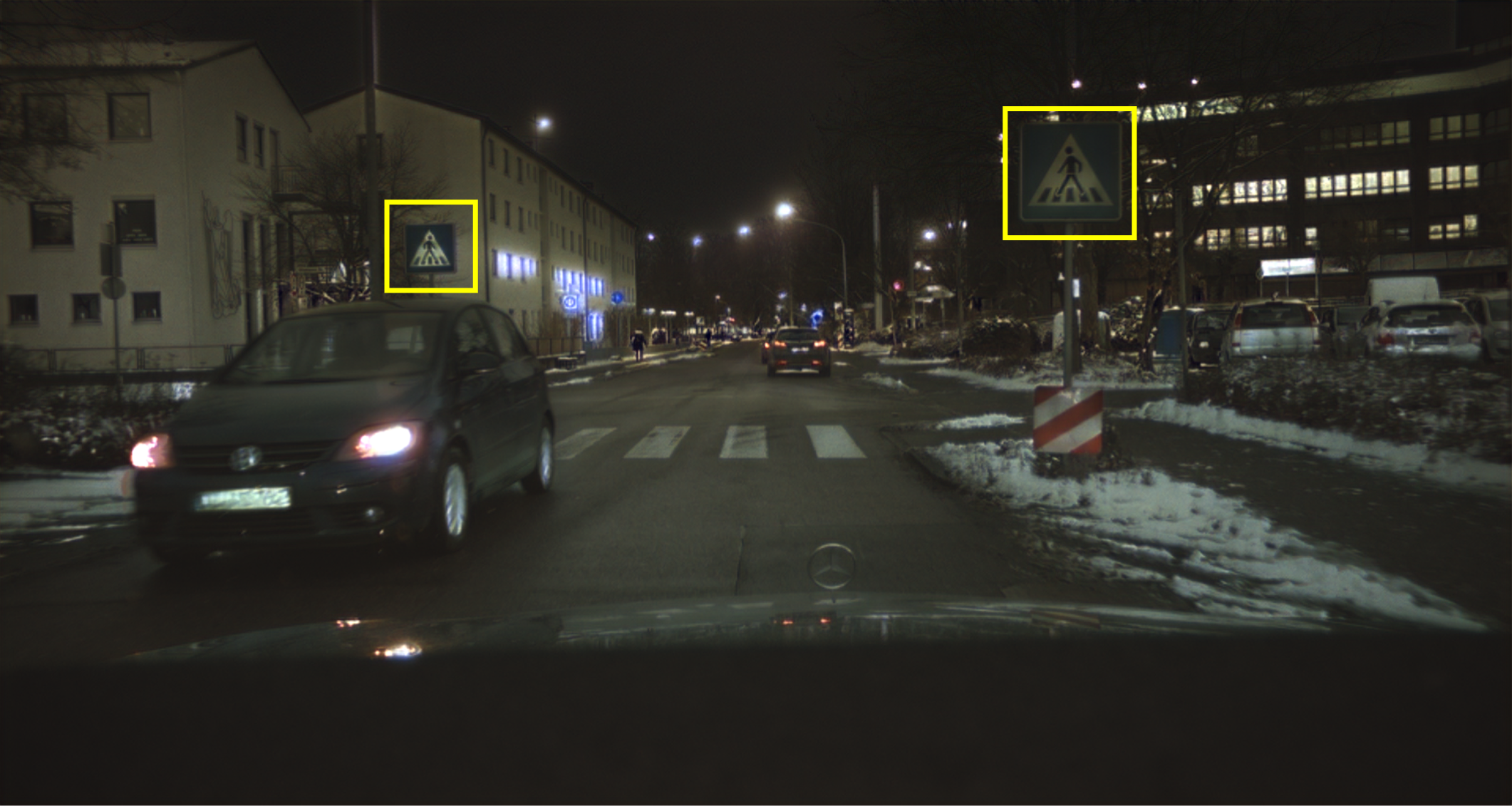}} &
\raisebox{-0.5\height}{\includegraphics[width=0.325\textwidth]{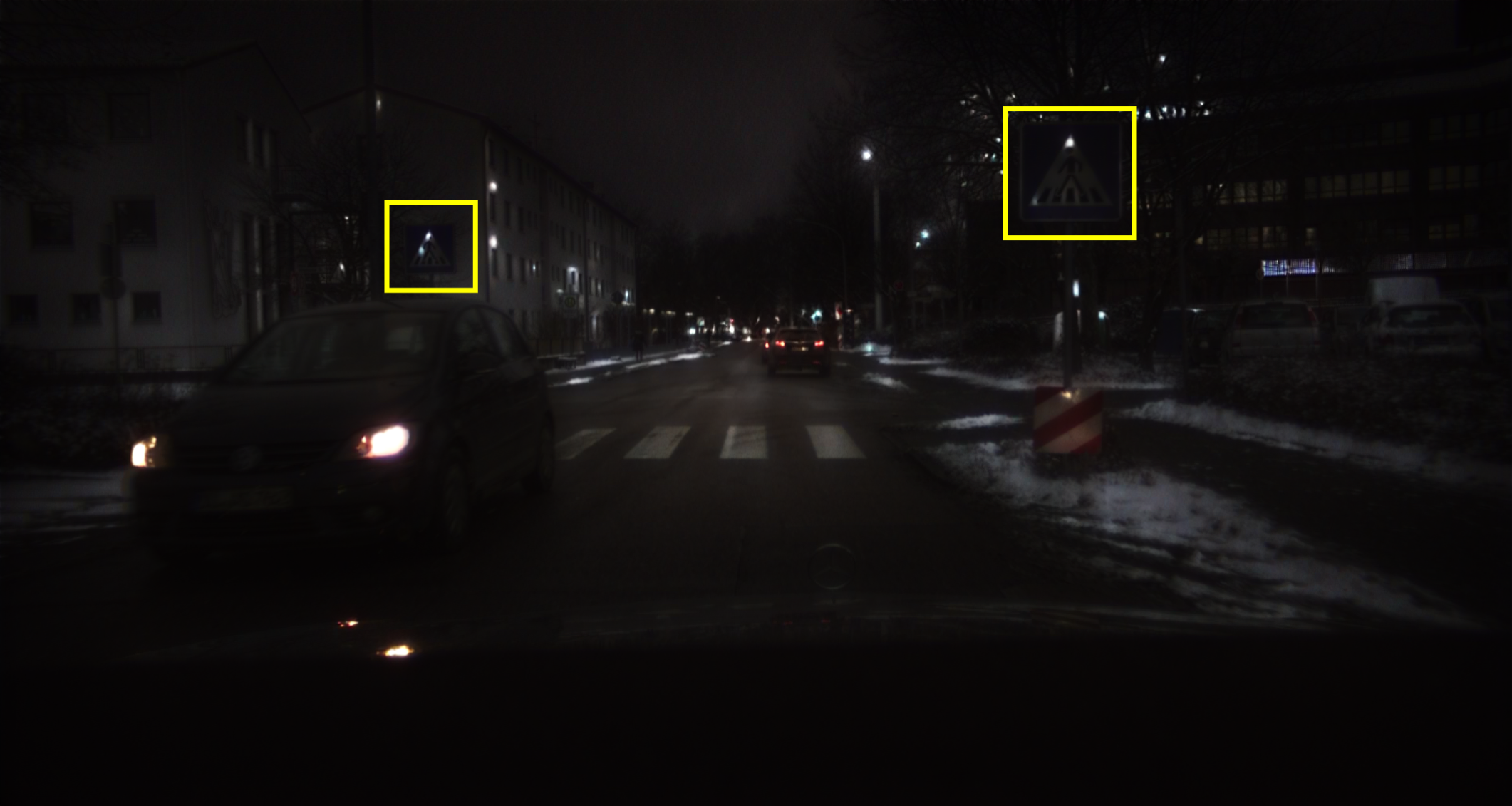}} \\
\noalign{\vskip 3pt}

\raisebox{-0.5\height}{\includegraphics[width=0.325\textwidth]{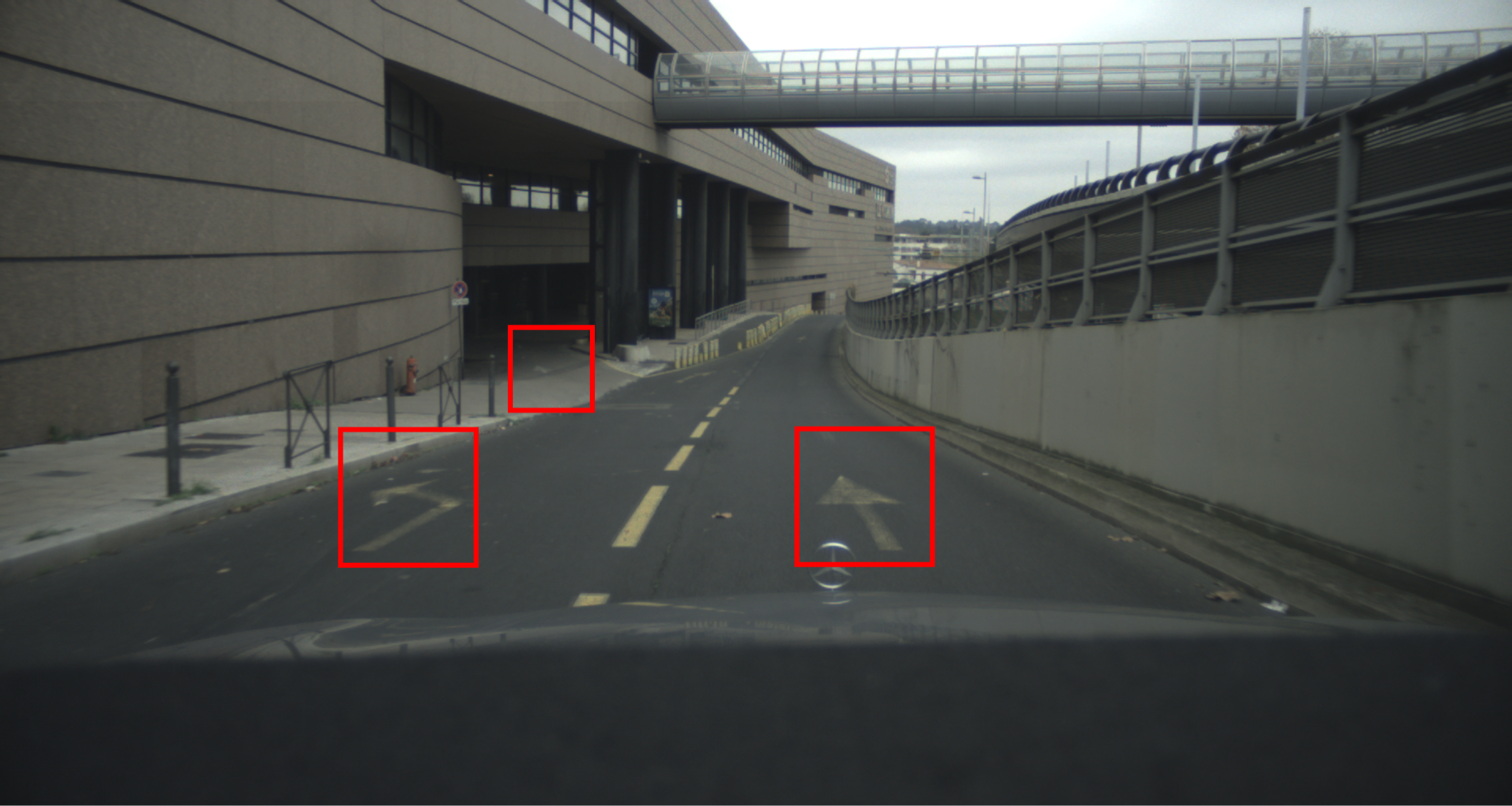}} &
\raisebox{-0.5\height}{\includegraphics[width=0.325\textwidth]{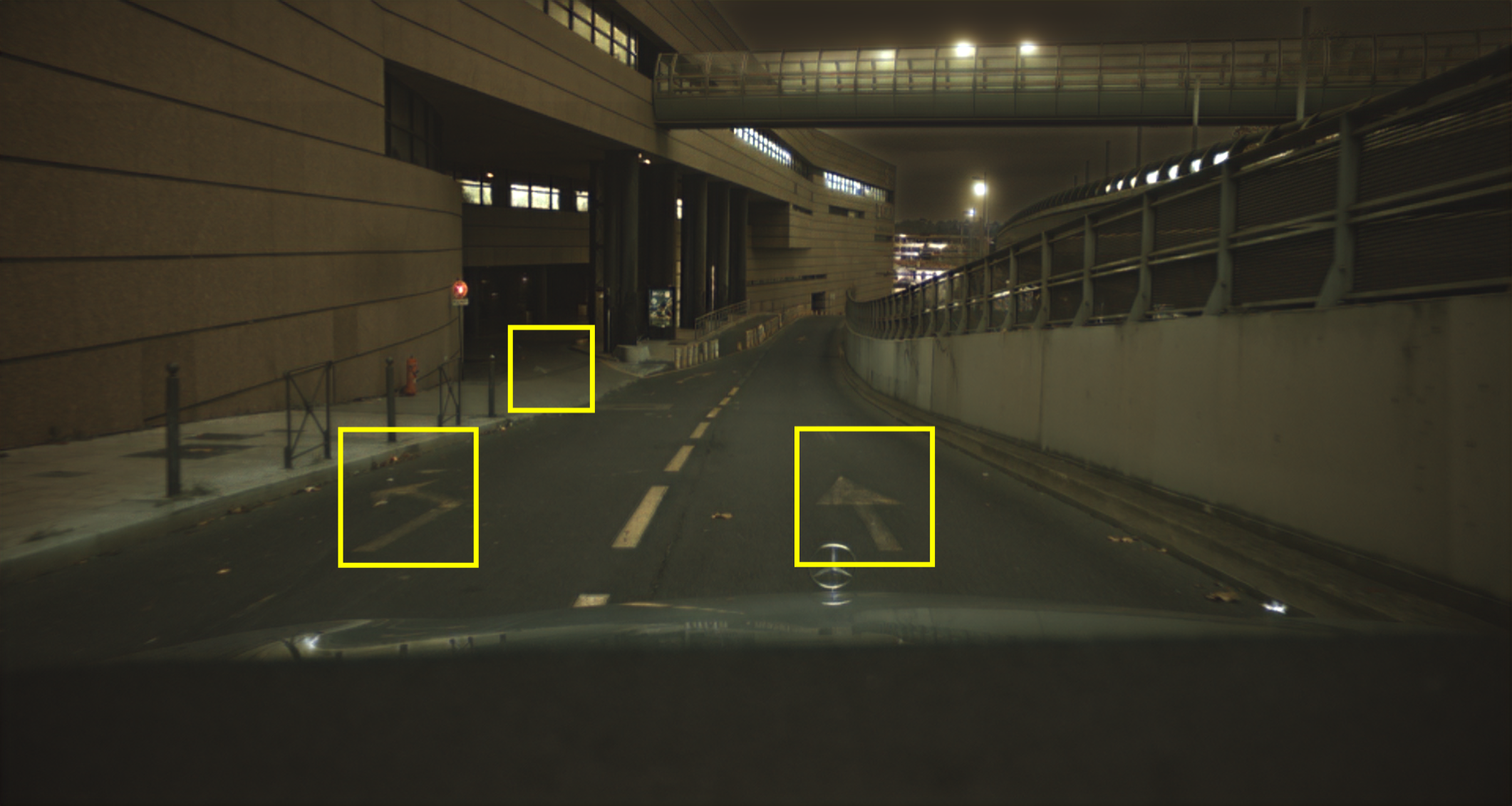}} &
\raisebox{-0.5\height}{\includegraphics[width=0.325\textwidth]{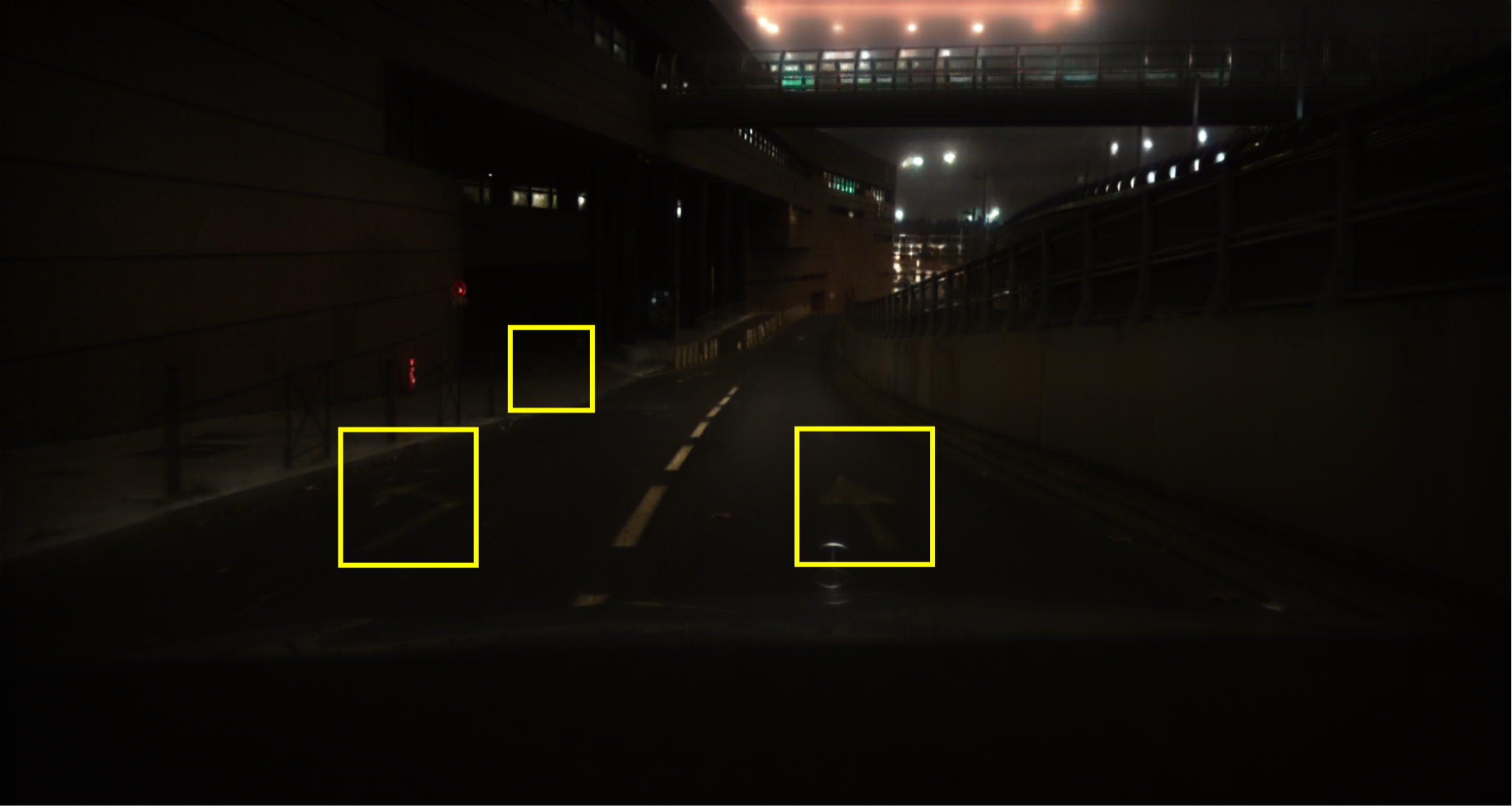}} \\
\noalign{\vskip 3pt}

\raisebox{-0.5\height}{\includegraphics[width=0.325\textwidth]{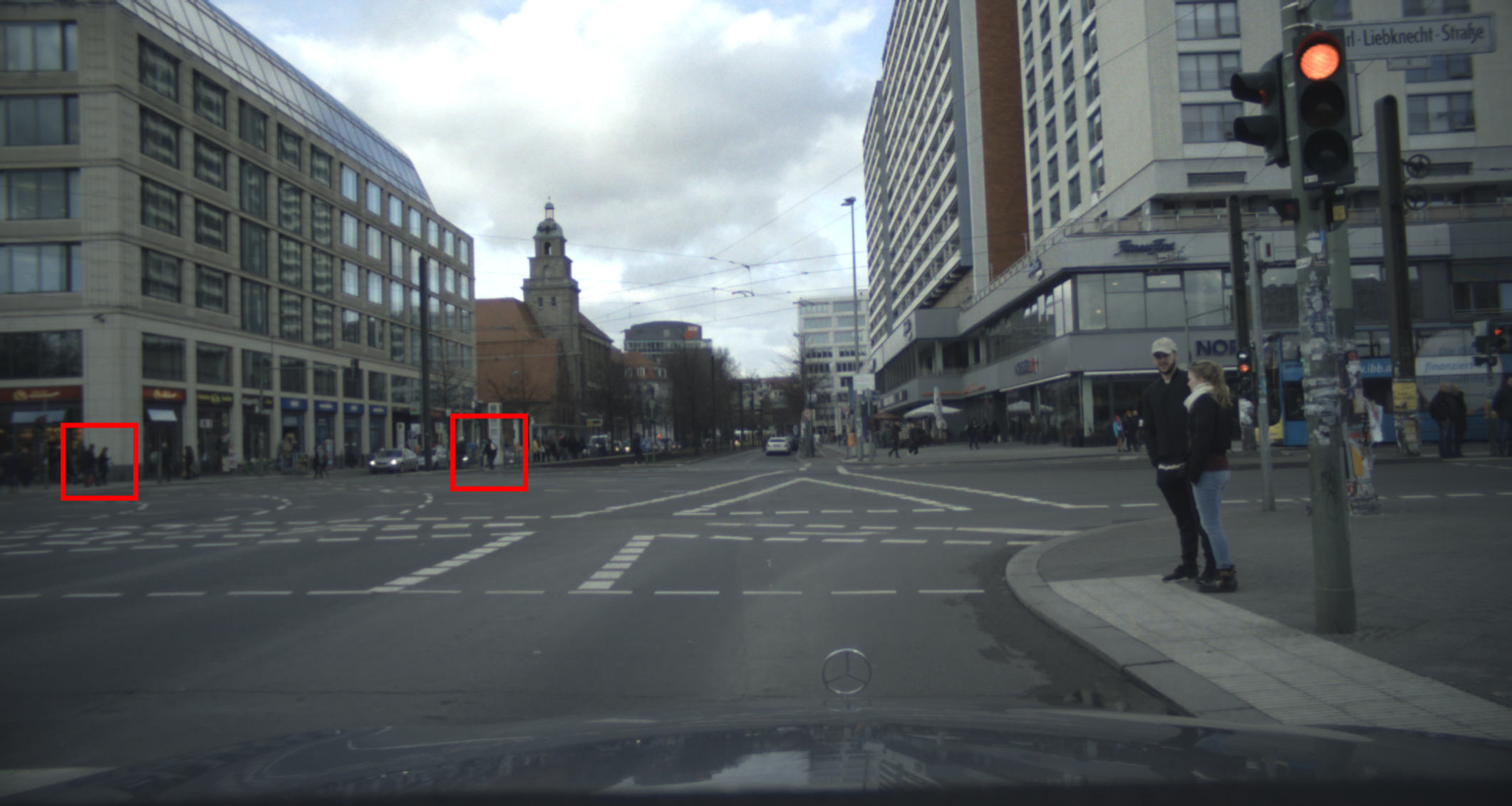}} &
\raisebox{-0.5\height}{\includegraphics[width=0.325\textwidth]{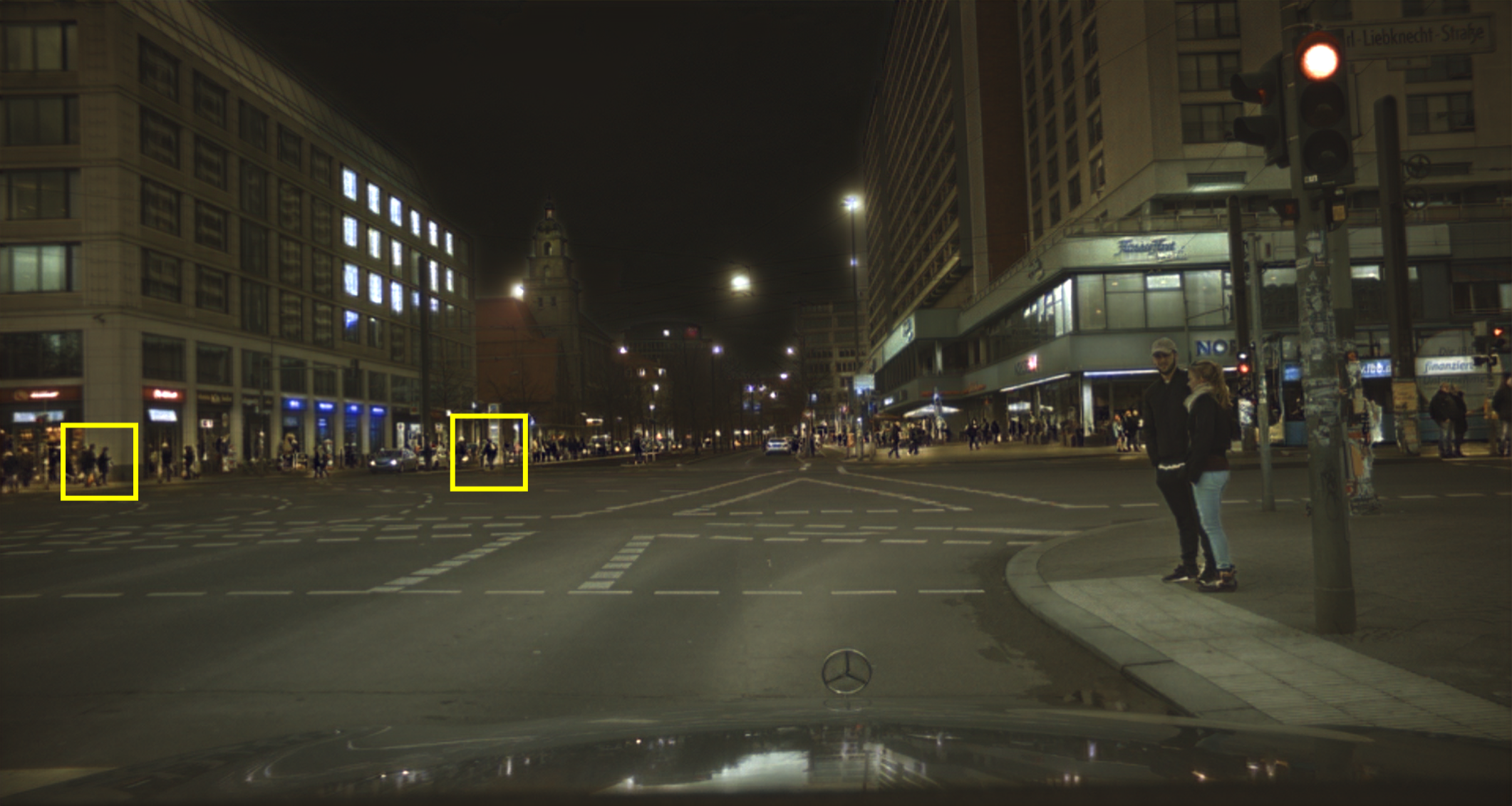}} &
\raisebox{-0.5\height}{\includegraphics[width=0.325\textwidth]{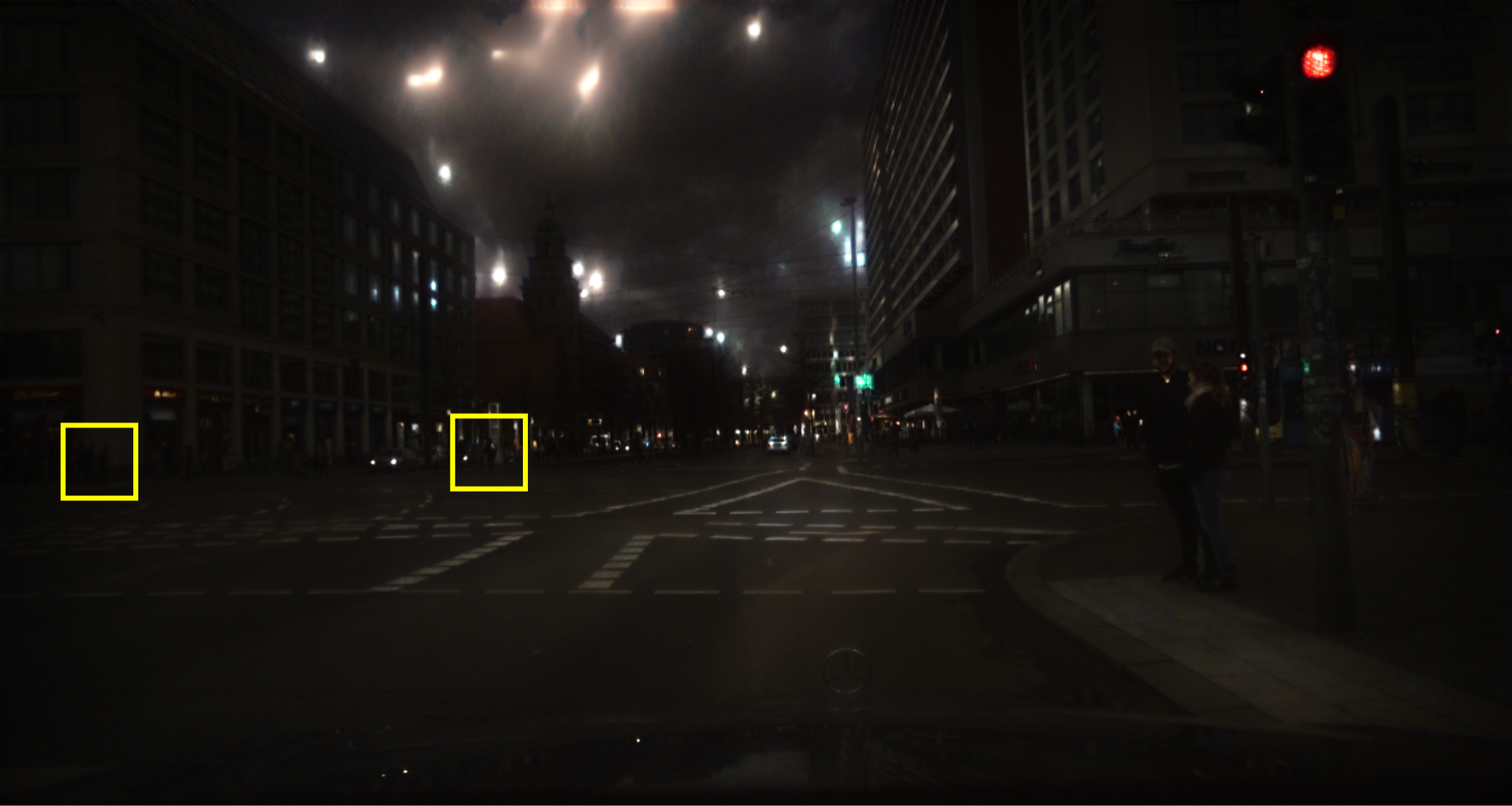}} \\
\end{tabular}
\caption{Qualitative comparison of local semantic consistency in day-to-night translation. Each row shows the daytime input, Contrastive-SDXL, and CycleGAN-Turbo from left to right. Highlighted regions focus on small or local structures, including traffic signs, road markings, and small or distant pedestrians. Contrastive-SDXL preserves visible correspondence with these source details, whereas CycleGAN-Turbo alters some structures and renders others too dark for reliable assessment.}
\label{fig:local_semantic_consistency}
\end{figure}

\subsection{Comparison to Baselines and Other Translation Models}

We compare Contrastive-SDXL with InstructPix2Pix and CycleGAN-Turbo in terms of synthetic image quality, night-domain distribution alignment and downstream detection performance. 

FID and WD are reported in Table~\ref{tab:fid_wd}, UMAP visualisations are shown in Figure~\ref{fig:umap}, and detector performance under synthetic-only and at different real-night injection ratios is reported in Table~\ref{tab:detector_aug_results}. For the detection experiments, the daytime-only detector serves as the baseline, while the model trained on full real night-time data serves as the target-domain reference.

\begin{table}[!t]
\centering
\caption{Pedestrian detection performance on the ECP day and night validation set under different training configurations. Results are reported as log-average miss rate across evaluation subsets. The baseline model (BM) highlights the day-to-night domain gap, while fine-tuned models use synthetic night-time images from different image-to-image diffusion models with varying real night augmentation ratios ($\rho$). The Target Model, fine-tuned on the full ECP night-time training set, serves as a benchmark. Models with overall best results are shown in bold.}

\label{tab:detector_aug_results}
\label{tab:pedestron_aug_results}
\small
\renewcommand{\arraystretch}{1.10}
\resizebox{1.0\textwidth}{!}{%
\begin{tabular}{p{3.6cm} p{2.2cm} c p{2.4cm} c c c c}
\toprule
\makecell{\textbf{Pedestrian}\\\textbf{Detector}} &
\textbf{I2I-Model} &
\makecell{\textbf{Real night}\\\textbf{injection ratio (\%)}} &
\textbf{Validation Dataset} &
\textbf{Reasonable $\downarrow$} &
\textbf{Small $\downarrow$} &
\textbf{Occluded $\downarrow$} &
\textbf{All $\downarrow$} \\
\midrule

\multirow{2}{*}{\textbf{Pedestron}-BM} &
-- & -- & ECP Day   & 4.0  & 10.0 & 24.0 & 14.0 \\
& -- & -- & ECP Night & 11.0 & 17.0 & 31.0 & 20.0 \\

\midrule

\multirow{12}{*}{Fine-tuned Models} &
\multirow{4}{*}{Contrastive-SDXL} &
0\%  & \multirow{4}{*}{ECP Night} & 9.0  & 14.0 & 25.0 & 16.0 \\
& & 5\%  &  & 8.0  & 12.0 & 25.0 & 15.0 \\
& & 10\% &  & 8.0  & 11.0 & 24.0 & 15.0 \\
& & 20\% &  & \textbf{8.0} & \textbf{11.0} & \textbf{23.0} & \textbf{14.0} \\

\noalign{\vskip 3pt}
\cdashline{2-8}
\noalign{\vskip 3pt}

& \multirow{4}{*}{CycleGAN-Turbo} &
0\%  & \multirow{4}{*}{ECP Night} & 14.0 & 18.0 & 31.0 & 21.0 \\
& & 5\%  &  & 11.0 & 17.0 & 29.0 & 18.0 \\
& & 10\% &  & 10.0 & 16.0 & 26.0 & 17.0 \\
& & 20\% &  & 9.0  & 15.0 & 26.0 & 16.0 \\

\noalign{\vskip 3pt}
\cdashline{2-8}
\noalign{\vskip 3pt}

& \multirow{4}{*}{InstructPix2Pix} &
0\%  & \multirow{4}{*}{ECP Night} & 28.0 & 31.0 & 46.0 & 37.0 \\
& & 5\%  &  & 16.0 & 22.0 & 41.0 & 25.0 \\
& & 10\% &  & 12.0 & 18.0 & 33.0 & 20.0 \\
& & 20\% &  & 12.0 & 15.0 & 31.0 & 19.0 \\

\midrule

\makecell{Target Model\\{\footnotesize (\textbf{Pedestron} fine-tuned}\\{\footnotesize on real night images)}} &
-- & -- & ECP Night & 7.0 & 11.0 & 22.0 & 14.0 \\

\specialrule{\heavyrulewidth}{3pt}{3pt}

\multirow{2}{*}{\textbf{YOLO}-BM} &
-- & -- & ECP Day   & 4.0  & 11.0 & 22.0 & 15.0 \\
& -- & -- & ECP Night & 10.0 & 17.0 & 36.0 & 19.0 \\

\midrule

\multirow{6}{*}{Fine-tuned Models} &
\multirow{2}{*}{Contrastive-SDXL} &
0\%  & \multirow{2}{*}{ECP Night} & 9.0  & 16.0 & 33.0 & 18.0 \\
& & 5\% &  & \textbf{9.0} & \textbf{14.0} & \textbf{31.0} & \textbf{17.0} \\

\noalign{\vskip 3pt}
\cdashline{2-8}
\noalign{\vskip 3pt}

& \multirow{2}{*}{CycleGAN-Turbo} &
0\%  & \multirow{2}{*}{ECP Night} & 12.0 & 20.0 & 35.0 & 21.0 \\
& & 5\% &  & 11.0  & 18.0 & 33.0 & 21.0 \\

\noalign{\vskip 3pt}
\cdashline{2-8}
\noalign{\vskip 3pt}

& \multirow{2}{*}{InstructPix2Pix} &
0\%  & \multirow{2}{*}{ECP Night} & 18.0 & 30.0 & 45.0 & 29.0 \\
& & 5\% &  & 15.0 & 22.0 & 40.0 & 25.0 \\

\midrule

\makecell{Target Model\\{\footnotesize (\textbf{YOLO} fine-tuned}\\{\footnotesize on real night images)}} &
-- & -- & ECP Night & 7.0 & 13.0 & 28.0 & 16.0 \\

\bottomrule
\end{tabular}
}
\end{table}


\subsubsection{Target-Domain Distribution Alignment}

Table~\ref{tab:fid_wd} shows that Contrastive-SDXL achieves the lowest FID of 22.57, compared with 38.36 for CycleGAN-Turbo and 64.11 for InstructPix2Pix. Contrastive-SDXL also obtains the lowest WD of 11.27, substantially below InstructPix2Pix at 31.8 and CycleGAN-Turbo at 37.4. For context, the WD between ECP daytime and night-time images is 34.8, whereas two disjoint subsets of real ECP night-time images have a WD of 5.6. The Contrastive-SDXL value therefore indicates substantially closer agreement with the real night-time distribution than either the original daytime domain or the other synthetic sets, which is also reflected qualitatively in the examples shown in Figure~\ref{fig:qualitative_comparison}. The UMAP visualisation in Figure~\ref{fig:umap} is qualitatively consistent with these quantitative results. Contrastive-SDXL embeddings exhibit greater overlap with the real night-time samples and clearer separation from the daytime distribution than the other generated sets. 

Figure~\ref{fig:local_semantic_consistency} provides a qualitative comparison of local structure preservation by Contrastive-SDXL and CycleGAN-Turbo across four daytime scenes and their translated night-time counterparts.

\subsubsection{Downstream Detection Performance}

Table~\ref{tab:detector_aug_results} reports detection performance under different augmentation settings. It first demonstrates the day-to-night domain gap for both detector families. Without night-time adaptation, the Pedestron \emph{All} LAMR increases from 14.0\% on ECP Day to 20.0\% on ECP Night, while the YOLO baseline increases from 15.0\% to 19.0\%.

In the synthetic-only setting, Contrastive-SDXL improves both detectors relative to their respective daytime-only baselines. For Pedestron, the \emph{All} LAMR decreases from 20.0\% to 16.0\%, while CycleGAN-Turbo obtains 21.0\% and InstructPix2Pix 37.0\%, neither improving over the daytime baseline. When labelled real night-time images equal to 20\% of the synthetic-set size are added, Contrastive-SDXL reaches an \emph{All} LAMR of 14.0\%. This matches the result of the model fine-tuned on the complete real night-time training set. 

The same overall pattern is observed with YOLO. Synthetic-only Contrastive-SDXL reduces the \emph{All} LAMR from the daytime baseline of 19.0\% to 18.0\%, whereas CycleGAN-Turbo and InstructPix2Pix obtain 21.0\% and 29.0\%, respectively. Adding labelled real night-time images equal to only 5\% of the synthetic-set size further reduces the Contrastive-SDXL result to 17.0\%, compared with 16.0\% when YOLO is fine-tuned using the complete real night-time training set. Unlike Pedestron, increasing the amount of labelled real night-time data beyond the 5\% injection ratio produced negligible or no further improvement in LAMR. Therefore, only the 0\% and 5\% real-night injection settings are reported for YOLO.

These results show that augmentation usefulness of the generated images depends not only on perceptual night-time realism but also on preserving source content that is relevant to downstream detection. Across the two evaluated detector families, Contrastive-SDXL is the only synthetic augmentation method that improves the LAMR over the corresponding daytime-only baseline without the addition of real night-time training images. Limited real-night injection provides further gains, with Pedestron matching the LAMR of full real-night training at a 20\% injection rate and YOLO reaching within one percentage point of its fully supervised reference at only a 5\% injection rate.

\subsection{Ablation Study}
\label{sec:ablation_study}

\begin{table}[htbp]
\centering
\caption{Ablation study of the proposed Contrastive-SDXL framework for day-to-night translation and downstream pedestrian detection. $L_{\mathrm{det}}$ denotes the detector-guided object consistency loss, while $L_{\mathrm{idt}}$ denotes the identity regularisation loss. Values in parentheses indicate the relative degradation with respect to the complete Contrastive-SDXL configuration. Lower values indicate better performance.}

\label{tab:ablation_study}

\small
\renewcommand{\arraystretch}{1.4}
\setlength{\tabcolsep}{3.5pt}

\newcommand{\reldelta}[1]{{\scriptsize\,(\,#1\%\,)}}

\resizebox{\textwidth}{!}{%
\begin{tabular}{p{2.8cm} c c c c c c c}
\toprule

\textbf{Ablation} &
\multicolumn{2}{c}{\textbf{Day $\rightarrow$ Night}} &
\textbf{Pedestrian Detector} &
\multicolumn{4}{c}{\textbf{Detector Performance}} \\

\cmidrule(lr){2-3}
\cmidrule(lr){5-8}

&
FID $\downarrow$ &
WD $\downarrow$ &
&
Reasonable $\downarrow$ &
Small $\downarrow$ &
Occluded $\downarrow$ &
All $\downarrow$ \\

\midrule

\multirow{2}{*}{\makecell[l]{Without $L_{\mathrm{det}}$}} &
\multirow{2}{*}{24.9 \reldelta{+10.3}} &
\multirow{2}{*}{11.7 \reldelta{+3.8}} &
Pedestron &
10.0 \reldelta{+11.1} &
14.2 \reldelta{+1.4} &
26.0 \reldelta{+4.0} &
16.0 \reldelta{0.0} \\

\cdashline{4-8}

&
&
&
YOLO26 &
11.0 \reldelta{+22.2} &
18.0 \reldelta{+12.5} &
35.0 \reldelta{+6.1} &
19.8 \reldelta{+10.0} \\

\specialrule{0.1pt}{5pt}{5pt}

\multirow{2}{*}{\makecell[l]{Without $L_{\mathrm{idt}}$}} &
\multirow{2}{*}{27.4 \reldelta{+21.4}} &
\multirow{2}{*}{12.0 \reldelta{+6.5}} &
Pedestron &
10.0 \reldelta{+11.1} &
15.0 \reldelta{+7.1} &
26.0 \reldelta{+4.0} &
17.0 \reldelta{+6.3} \\

\cdashline{4-8}

&
&
&
YOLO26 &
12.0 \reldelta{+33.3} &
19.2 \reldelta{+20.0} &
36.0 \reldelta{+9.1} &
21.9 \reldelta{+21.7} \\

\specialrule{0.1pt}{5pt}{5pt}

\multirow{2}{*}{\makecell[l]{Contrastive-SDXL\\(Ours)}} &
\multirow{2}{*}{\textbf{22.57}} &
\multirow{2}{*}{\textbf{11.27}} &
Pedestron &
\textbf{9.0} &
\textbf{14.0} &
\textbf{25.0} &
\textbf{16.0} \\

\cdashline{4-8}

&
&
&
YOLO26 &
\textbf{9.0} &
\textbf{16.0} &
\textbf{33.0} &
\textbf{18.0} \\

\bottomrule
\end{tabular}
}
\end{table}

We conduct an ablation study to examine the contributions of the detector-guided object consistency loss $L_{\mathrm{det}}$, identity regularisation $L_{\mathrm{idt}}$, and the use of DINOv2 as an external semantic encoder. For the loss ablations, each component is removed individually while the remaining training configuration is kept unchanged. We evaluate the resulting synthetic images using FID and Wasserstein distance (WD), and assess their downstream utility by fine-tuning Pedestron and YOLO26 using the corresponding synthetic datasets without real night-time injection. Table~\ref{tab:ablation_study} summarises the quantitative results. Values in parentheses denote the relative changes with respect to the full Contrastive-SDXL configuration. 

\paragraph{Effect of detector-guided object consistency loss}
The results in Table~\ref{tab:ablation_study} show that removing $L_{\mathrm{det}}$ increases FID from 22.57 to 24.9 and WD from 11.27 to 11.7, corresponding to relative increases of 10.3\% and 3.8\%, respectively. The effect is also reflected in downstream detection performance. For Pedestron, the \emph{All} LAMR remains the same at 16.0\%, while the \emph{Reasonable}, \emph{Small}, and \emph{Occluded} subsets increase from 9.0\%, 14.0\%, and 25.0\% to 10.0\%, 14.2\%, and 26.0\%, respectively. The effect is more pronounced for YOLO26, where removing $L_{\mathrm{det}}$ increases the \emph{All} LAMR from 18.0\% to 19.8\%. The corresponding LAMR values for the \emph{Reasonable}, \emph{Small}, and \emph{Occluded} subsets increase from 9.0\%, 16.0\%, and 33.0\% to 11.0\%, 18.0\%, and 35.0\%, respectively.

The DINOv2 representations used in the contrastive objectives encourage broader semantic correspondence between the source and translated images, whereas $L_{\mathrm{det}}$ explicitly penalises changes that affect pedestrian consistency. Therefore, removing this objective means a translated image can remain broadly similar to its source and visually aligned with the target domain while failing to preserve pedestrian-relevant features. This is particularly important when the source annotations are reused for detector training. Although less pronounced, the effect of removing $L_{\mathrm{det}}$ is also evident in the distributional alignment metrics. This suggests that the detector guidance provides an additional regularising effect on the translation process. 

\paragraph{Effect of identity regularisation}
Removing $L_{\mathrm{idt}}$ results in a larger deterioration in distributional alignment, increasing FID from 22.57 to 27.4 and WD from 11.27 to 12.0. These correspond to relative increases of 21.4\% and 6.5\%, respectively. A similar degradation is observed in downstream detector performance. Without identity regularisation, the \emph{Reasonable}, \emph{Small}, \emph{Occluded}, and \emph{All} LAMR values for YOLO26 increase from 9.0\%, 16.0\%, 33.0\%, and 18.0\% to 12.0\%, 19.2\%, 36.0\%, and 21.9\%, respectively. A similar trend is observed for Pedestron, where the corresponding LAMR values increase from 9.0\%, 14.0\%, 25.0\%, and 16.0\% to 10.0\%, 15.0\%, 26.0\%, and 17.0\%, respectively. For YOLO26, the \emph{All} LAMR therefore increases by 3.9 percentage points relative to the full Contrastive-SDXL model and also exceeds the 19.0\% daytime-only YOLO26 baseline on ECP Night.

These results are consistent with the intended role of identity regularisation, which is to discourage unnecessary modifications to real target-domain input. Its removal affects both distributional alignment and downstream detection performance, indicating that limiting unnecessary changes during translation is important for generating synthetic images that resemble the target domain while remaining useful for downstream detector adaptation.

\paragraph{Effect of the DINOv2 external encoder}
The source of the representations used for patch-wise semantic contrastive objectives was also examined during the development of Contrastive-SDXL. Initially, we evaluated variants in which the contrastive objectives were computed using self-attention representations extracted from different levels of the SDXL-Turbo U-Net encoder. As illustrated in Figure~\ref{fig:unet_vs_dinov2}, these generator-feature variants exhibit weaker correspondence between the source and translated images, including structural distortions and poorer preservation of fine pedestrian details. This motivated the use of DINOv2 as an external semantic encoder. DINOv2 is pre-trained on a large-scale dataset using a self-supervised objective and provides semantically meaningful representations that are suitable for the contrastive objectives. The qualitative comparison in Figure~\ref{fig:unet_vs_dinov2} shows that DINOv2 representations provide better preservation of source-scene structure, supporting their use in Contrastive-SDXL.

\begin{figure}[!t]
\centering

{\small \textbf{Day $\rightarrow$ Night}}\par\vspace{2mm}

\setlength{\tabcolsep}{2pt}

\begin{tabular}{@{}c@{\hspace{1mm}}c@{\hspace{1mm}}c@{}}
\multirow{3}{*}[-16mm]{\rotatebox{90}{\small\textbf{Input}}}
&
\includegraphics[width=0.45\textwidth]{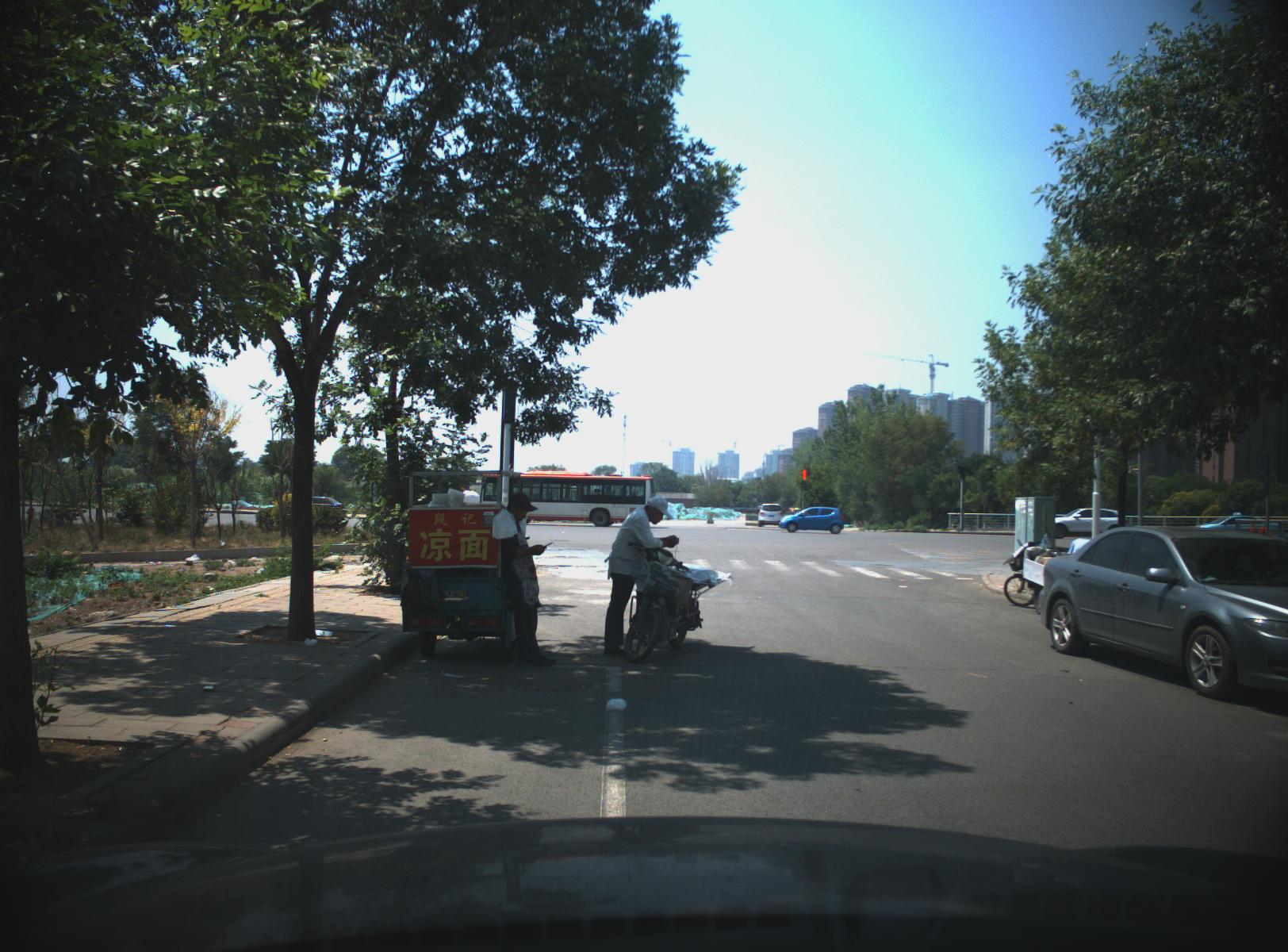}
&
\includegraphics[width=0.45\textwidth]{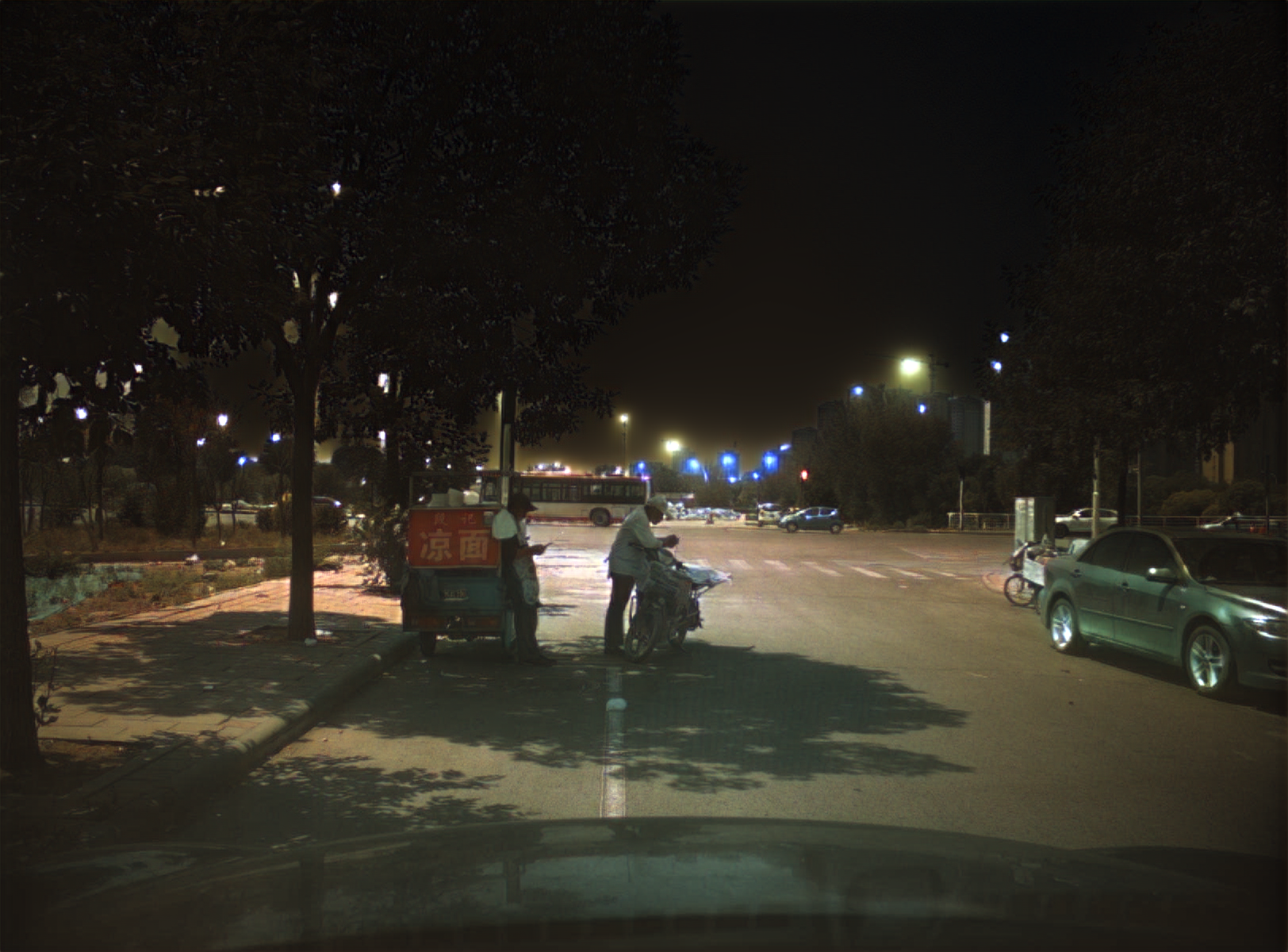}
\\[-0.5mm]

&
\includegraphics[width=0.45\textwidth]{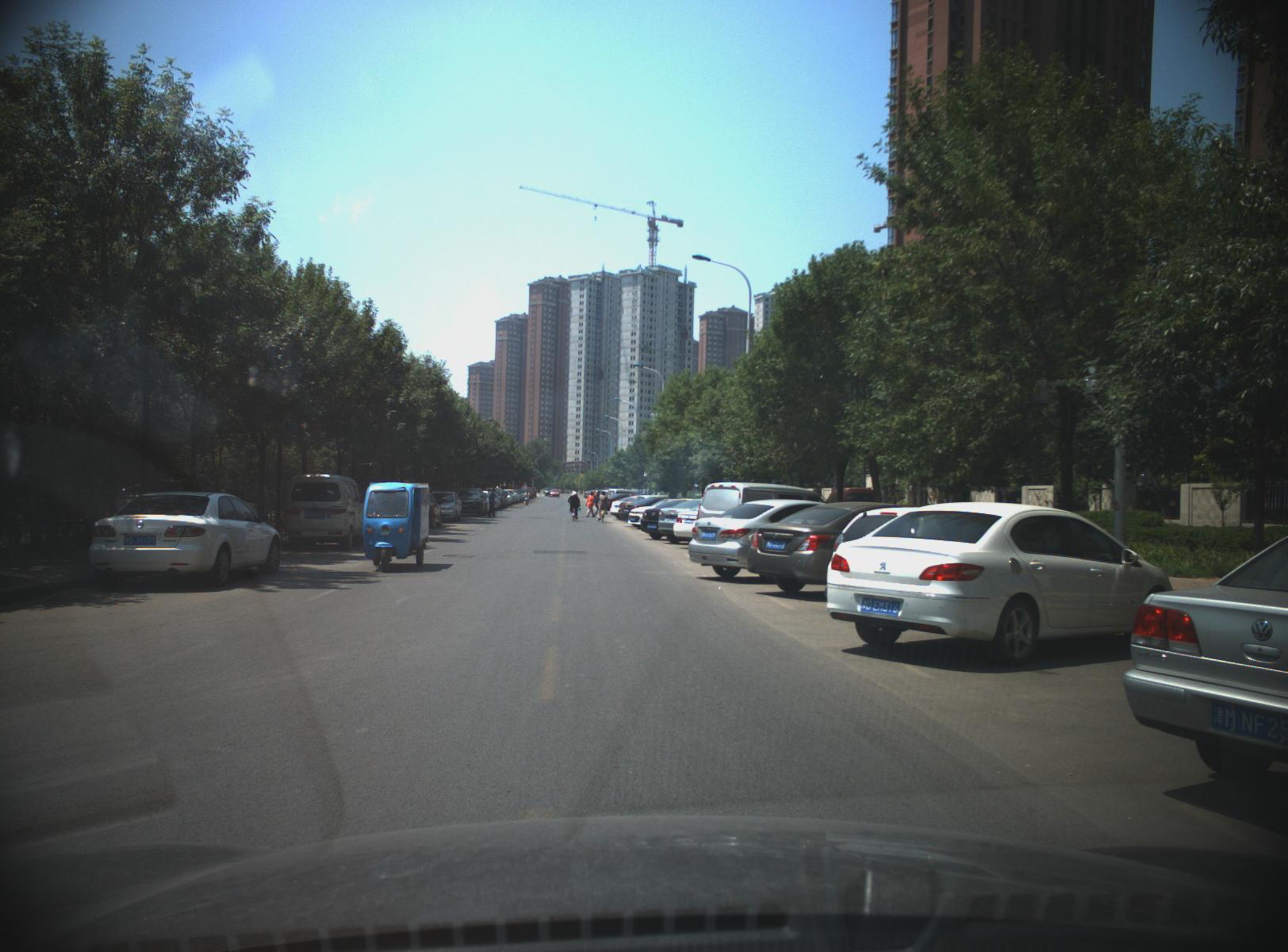}
&
\includegraphics[width=0.45\textwidth]{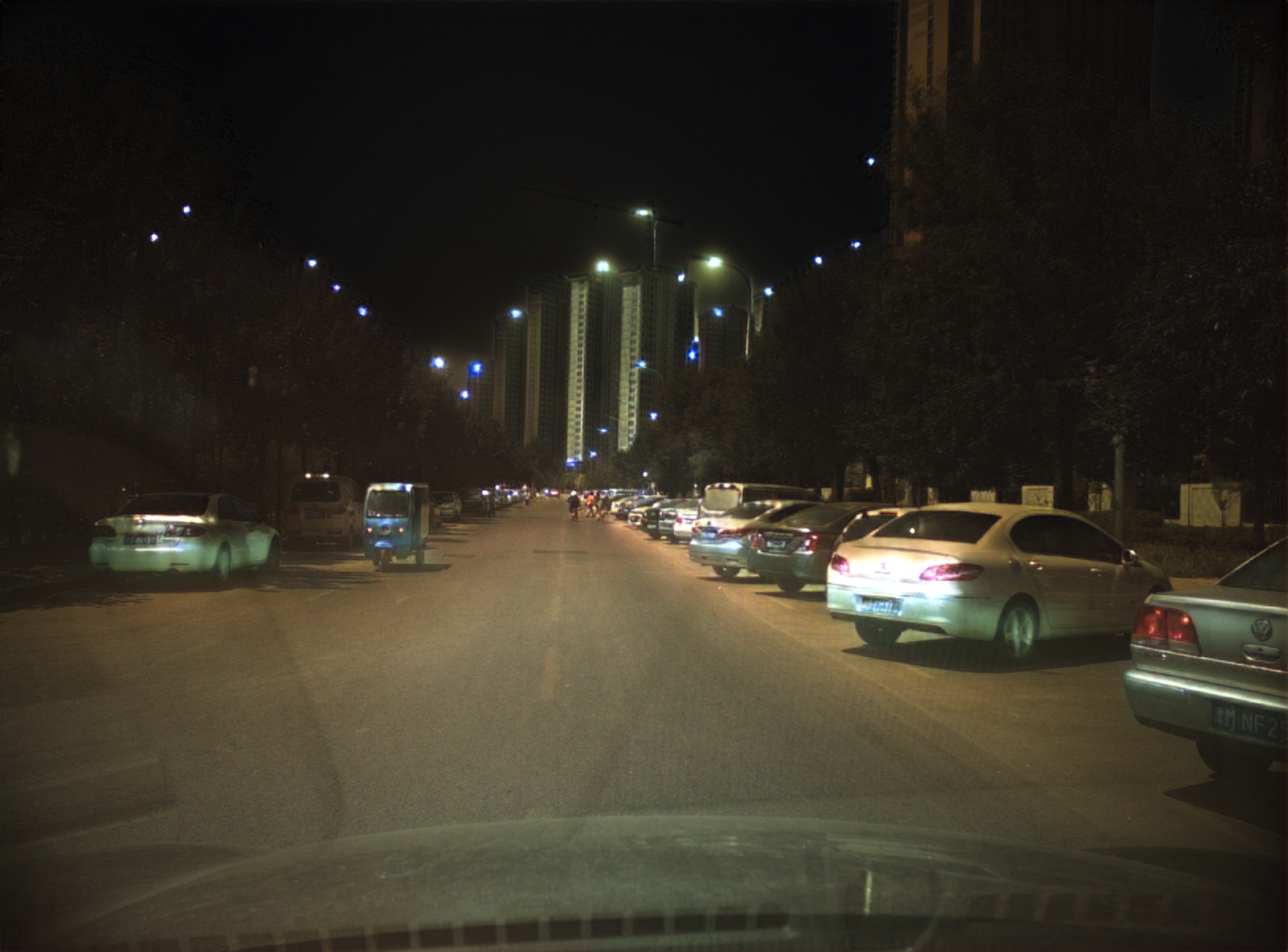}
\\[-0.5mm]

&
\includegraphics[width=0.45\textwidth]{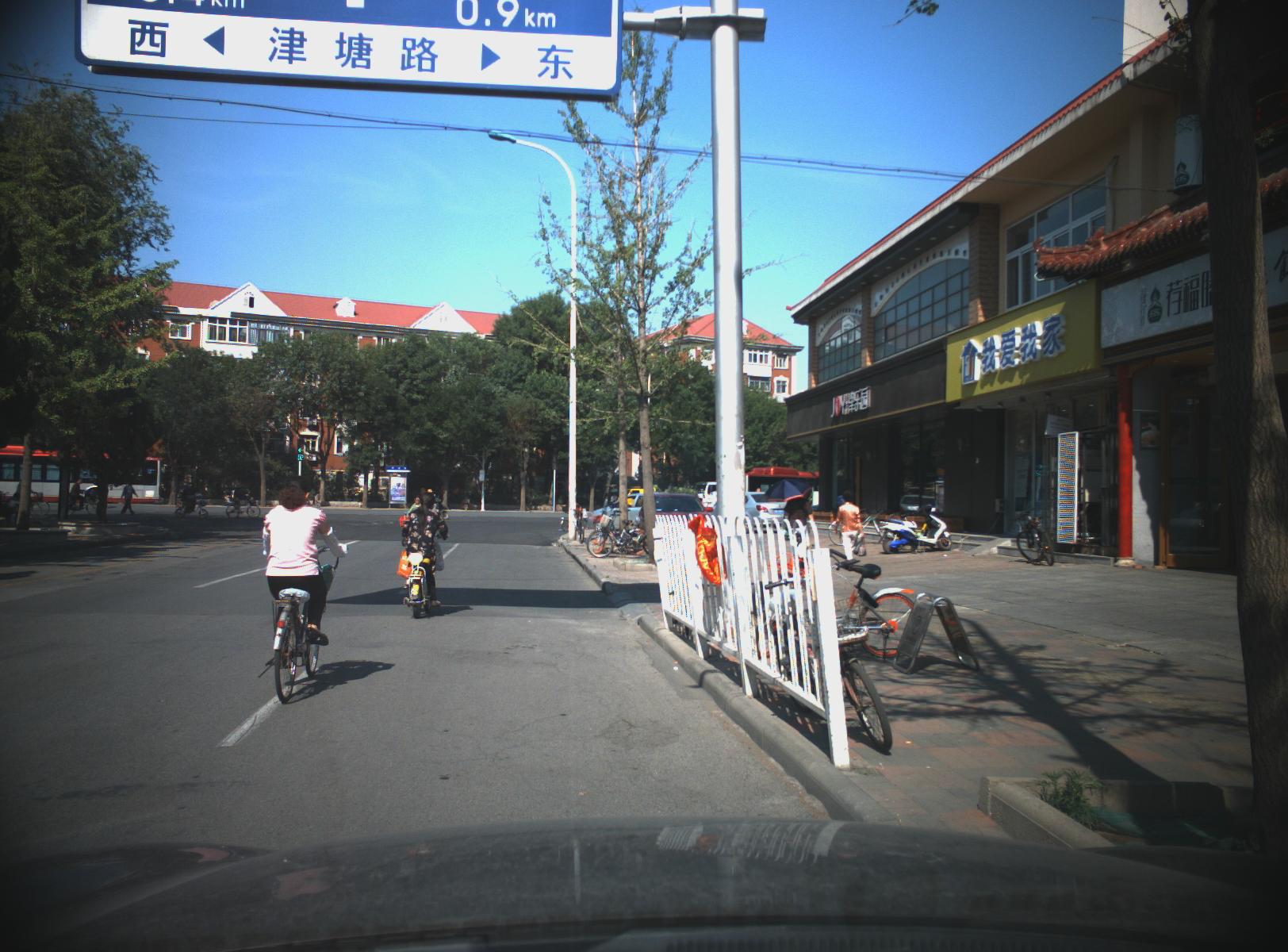}
&
\includegraphics[width=0.45\textwidth]{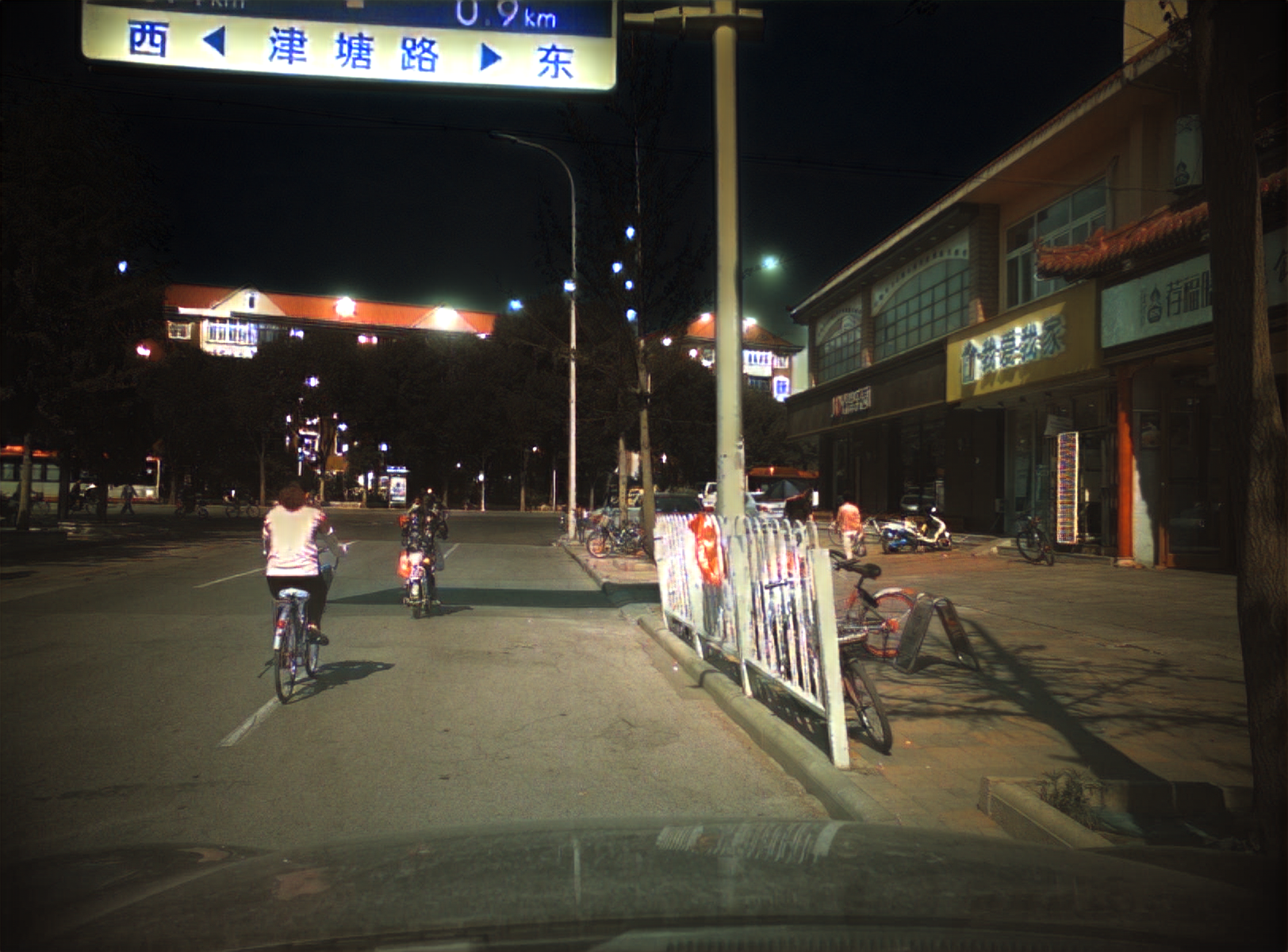}
\end{tabular}

\caption{Example daytime images from the TJU-DHD dataset (left column) and their corresponding night-time translations generated by Contrastive-SDXL (right column). The results demonstrate that Contrastive-SDXL effectively captures night-time lighting conditions while preserving pedestrian details, even in out-of-distribution urban scenes.}
\label{fig:ood_results}
\end{figure}

\subsection{Qualitative Generalisation to Out-of-Distribution Images}

To examine whether the day-to-night transformation learned by Contrastive-SDXL transfers beyond the ECP source domain, we apply the model to daytime images from the TJU-DHD dataset~\citep{pang2020tju}, which is not used during training. TJU-DHD contains diverse high-resolution scenes with substantial variation in illumination, environment, and pedestrian appearance. As shown in Figure~\ref{fig:ood_results}, Contrastive-SDXL produces visually realistic night-time appearances while preserving the structure of the source scenes and the visible pedestrians. These qualitative examples suggest that the learned transformation is not restricted to the visual characteristics of ECP and can generalise to previously unseen source imagery. However, this experiment evaluates only the cross-dataset generalisation of the translation model qualitatively; no downstream pedestrian-detection evaluation is performed on TJU-DHD.

\section{Discussion}
\label{sec:discussion}

The results show that the value of Contrastive-SDXL lies not only in producing images that resemble the night-time domain, but in producing synthetic training data whose inherited pedestrian annotations remain useful for downstream detection. Among the evaluated translation methods, Contrastive-SDXL achieves the closest alignment with the real night-time distribution according to the FID and Wasserstein distance, with the UMAP visualisation providing further evidence of this distributional alignment. The more important result from an engineering perspective, however, is the effect of these images on pedestrian detection under real night-time conditions. In the synthetic-only setting, Contrastive-SDXL reduces the LAMR across all four subsets (\emph{Reasonable}, \emph{Small}, \emph{Occluded}, \emph{All}) for both Pedestron and YOLO compared to their respective daytime-only baselines. Neither CycleGAN-Turbo nor InstructPix2Pix improves upon the corresponding daytime baseline. The improvement observed without real night-time images during downstream detector fine-tuning across two different detector architectures therefore provides evidence that the generated data contain information that is useful for the intended pedestrian-perception task rather than merely having greater visual similarity to the target domain.

These results also illustrate why target-domain realism and augmentation utility should be treated as related but distinct objectives. A translated image can exhibit convincing night-time characteristics while containing small structural changes that reduce the reliability of the annotations inherited from the daytime source. CycleGAN-Turbo, for example, achieves substantially better target-domain alignment than InstructPix2Pix according to FID, yet its generated images do not improve the LAMRs for either detector. The qualitative analysis of InstructPix2Pix similarly shows artefacts and content inconsistencies that can interfere with the correspondence between translated pedestrians and their source annotations. Distribution-level measures such as FID and Wasserstein distance are therefore useful for assessing whether the generated dataset approaches the target appearance distribution, but they cannot by themselves establish whether individual translated samples remain suitable for supervised detector training.

Contrastive-SDXL addresses these requirements through complementary constraints applied at different levels of the translation process. Adversarial supervision encourages alignment with the visual characteristics of the night-time domain, whereas the DINOv2-guided contrastive objectives constrain local and broader semantic changes relative to the daytime source. The detector-guided objective places additional emphasis on pedestrians, for which structural distortion or removal directly affects the validity of inherited bounding boxes, while identity regularisation discourages unnecessary modification of target-domain content. The downstream results suggest that jointly controlling these aspects of the translation is more useful for pedestrian-detection augmentation than optimising target-domain appearance alone. In this context, annotation preservation is not a secondary property of the generated images, rather a requirement that determines whether a visually plausible translation can serve as a valid labelled training sample.

The mixed synthetic--real experiments provide a second indication of the practical value of this formulation. For Pedestron, adding labelled real night-time images equal to 20\% of the synthetic-set size reduces the \emph{All} LAMR to 14.0\%, matching the result obtained when fine-tuning on the complete real night-time training set. For YOLO, adding real images equal to only 5\% of the synthetic-set size produces an \emph{All} LAMR of 17.0\%, compared with 16.0\% under full real-night training. The results show that the synthetic data can complement a comparatively small injection of labelled target-domain examples and substantially narrow the gap to full real-night training.

The difference in performance of Pedestron and YOLO is also informative. Pedestron obtains a larger improvement from synthetic-only augmentation and reaches the full-real result at a different real-data injection ratio from YOLO, despite both detectors being adapted using images generated from the same translation framework. The usefulness of synthetic augmentation therefore depends not only on the quality of the generated data, but also on how those data interact with the downstream detector. Differences in detector architecture, initialisation, and sensitivity to the composition of the adaptation set may contribute to this behaviour. These factors are not isolated individually in the present study, and attributing the difference to any one of them would therefore require further investigation. From an engineering perspective, the result also suggests that the appropriate balance between synthetic and real target-domain data should be validated for the detector being deployed rather than assumed to transfer unchanged between architectures.

The qualitative experiments on TJU-DHD provide further evidence that the learned transformation is not restricted to the visual appearance of ECP scenes. Contrastive-SDXL produces plausible night-time translations of previously unseen urban scenes while retaining visible pedestrian and scene structure. This result is encouraging for transfer beyond the training dataset, but it should be interpreted as qualitative cross-dataset evidence. Since downstream detection performance is not evaluated on translated TJU-DHD data, these experiments do not establish that the detection improvements observed on ECP will transfer unchanged to other datasets, camera systems, or geographic environments.

Overall, the experiments support a broader conclusion for diffusion-based data augmentation in vehicle perception. The suitability of synthetic data should be assessed according to the information required by the downstream engineering task rather than by target-domain realism alone. For night-time pedestrian detection, this requires the source images to be aligned with the night-time distribution while preserving the semantic content of the original scene and maintaining pedestrian structure so that the inherited annotations remain useful. Contrastive-SDXL addresses these requirements through target-domain alignment, DINOv2-guided semantic correspondence, detector-guided pedestrian consistency, and identity regularisation. The resulting synthetic data improve the performance of two downstream pedestrian detectors. Future work should extend the quantitative cross-dataset evaluation, investigate how the required balance of synthetic and real data varies across detector architectures, and examine whether the same annotation-preserving approach can be applied to other vehicle-perception tasks and more complex adverse-condition shifts.

\section{Conclusion}

This work presented Contrastive-SDXL, an annotation-preserving diffusion-based augmentation framework for night-time pedestrian detection in vehicle-mounted perception systems. The framework adapts SDXL-Turbo using LoRA and combines DINOv2-guided patch-wise semantic contrastive objectives, detector-guided pedestrian consistency, adversarial target-domain alignment, and identity regularisation to translate labelled daytime images towards the night-time domain while preserving detector-relevant scene content and the validity of inherited pedestrian annotations. On the EuroCity Persons dataset, Contrastive-SDXL achieves the closest target-domain alignment among the evaluated translation methods and, more importantly, produces synthetic data that improve downstream detection for both Pedestron and YOLO. In the synthetic-only setting, the \emph{All} LAMR decreases from 20.0\% to 16.0\% for Pedestron and from 19.0\% to 18.0\% for YOLO relative to their respective daytime-only baselines. When the synthetic data are combined with limited labelled night-time images, Pedestron reaches 14.0\%, matching the result obtained using the complete real night-time training set, while YOLO reaches 17.0\% compared with 16.0\% under full real-night training. These results show that the effectiveness of diffusion-based augmentation for intelligent-vehicle perception depends not only on reproducing the appearance of the target domain, but also on preserving the task-relevant information required for reliable supervised training. Contrastive-SDXL demonstrates that coupling target-domain adaptation with explicit semantic and object-level preservation can improve night-time pedestrian detection while reducing reliance on separately labelled night-time data.

\bibliographystyle{elsarticle-harv}
\bibliography{references}

@inproceedings{zhu2017unpaired,
  title={Unpaired image-to-image translation using cycle-consistent adversarial networks},
  author={Zhu, Jun-Yan and Park, Taesung and Isola, Phillip and Efros, Alexei A},
  booktitle={Proceedings of the IEEE international conference on computer vision},
  pages={2223--2232},
  year={2017}
}

@article{braun2018eurocity,
  title={The eurocity persons dataset: A novel benchmark for object detection},
  author={Braun, Markus and Krebs, Sebastian and Flohr, Fabian and Gavrila, Dariu M},
  journal={arXiv preprint arXiv:1805.07193},
  year={2018}
}

@inproceedings{brooks2023instructpix2pix,
  title = {{InstructPix2Pix}: Learning to follow image editing instructions},
  author={Brooks, Tim and Holynski, Aleksander and Efros, Alexei A},
  booktitle={Proceedings of the IEEE/CVF Conference on Computer Vision and Pattern Recognition},
  pages={18392--18402},
  year={2023}
}

@article{parmar2024one,
  title={One-step image translation with text-to-image models},
  author={Parmar, Gaurav and Park, Taesung and Narasimhan, Srinivasa and Zhu, Jun-Yan},
  journal={arXiv preprint arXiv:2403.12036},
  year={2024}
}

@article{hu2021lora,
  title={{LoRA}: Low-rank adaptation of large language models},
  author={Hu, Edward J and Shen, Yelong and Wallis, Phillip and Allen-Zhu, Zeyuan and Li, Yuanzhi and Wang, Shean and Wang, Lu and Chen, Weizhu},
  journal={arXiv preprint arXiv:2106.09685},
  year={2021}
}

@inproceedings{jung2022exploring,
  title={Exploring patch-wise semantic relation for contrastive learning in image-to-image translation tasks},
  author={Jung, Chanyong and Kwon, Gihyun and Ye, Jong Chul},
  booktitle={Proceedings of the IEEE/CVF conference on computer vision and pattern recognition},
  pages={18260--18269},
  year={2022}
}

@article{oquab2023dinov2,
  title={{DINOv2}: Learning robust visual features without supervision},
  author={Oquab, Maxime and Darcet, Timoth{\'e}e and Moutakanni, Th{\'e}o and Vo, Huy and Szafraniec, Marc and Khalidov, Vasil and Fernandez, Pierre and Haziza, Daniel and Massa, Francisco and El-Nouby, Alaaeldin and others},
  journal={arXiv preprint arXiv:2304.07193},
  year={2023}
}

@article{li2020generalized,
  title={Generalized focal loss: Learning qualified and distributed bounding boxes for dense object detection},
  author={Li, Xiang and Wang, Wenhai and Wu, Lijun and Chen, Shuo and Hu, Xiaolin and Li, Jun and Tang, Jinhui and Yang, Jian},
  journal={Advances in neural information processing systems},
  volume={33},
  pages={21002--21012},
  year={2020}
}

@inproceedings{zheng2020distance,
  title={Distance-IoU loss: Faster and better learning for bounding box regression},
  author={Zheng, Zhaohui and Wang, Ping and Liu, Wei and Li, Jinze and Ye, Rongguang and Ren, Dongwei},
  booktitle={Proceedings of the AAAI conference on artificial intelligence},
  volume={34},
  number={07},
  pages={12993--13000},
  year={2020}
}

@inproceedings{kumari2022ensembling,
  title = {Ensembling off-the-shelf models for {GAN} training},
  author={Kumari, Nupur and Zhang, Richard and Shechtman, Eli and Zhu, Jun-Yan},
  booktitle={Proceedings of the IEEE/CVF conference on computer vision and pattern recognition},
  pages={10651--10662},
  year={2022}
}

@inproceedings{neumann2019nightowls,
  author    = {Neumann, Lukáš and Karg, Michelle and Zhang, Shanshan and Scharfenberger, Christian and Piegert, Eric and Mistr, Sarah and Prokofyeva, Olga and Thiel, Robert and Vedaldi, Andrea and Zisserman, Andrew and Schiele, Bernt},
  title     = {NightOwls: A Pedestrians at Night Dataset},
  booktitle = {Computer Vision -- ACCV 2018, Lecture Notes in Computer Science},
  volume    = {11361},
  pages     = {691--705},
  year      = {2019},
  publisher = {Springer, Cham},
  doi       = {10.1007/978-3-030-20887-5_43},
}

@inproceedings{hasan2021generalizable,
  title={Generalizable pedestrian detection: The elephant in the room},
  author={Hasan, Irtiza and Liao, Shengcai and Li, Jinpeng and Akram, Saad Ullah and Shao, Ling},
  booktitle={Proceedings of the IEEE/CVF Conference on Computer Vision and Pattern Recognition},
  pages={11328--11337},
  year={2021}
}

@inproceedings{heusel2017gans,
  title = {{GANs} trained by a two time-scale update rule converge to a local {Nash} equilibrium},
  author={Heusel, Martin and Ramsauer, Hubert and Unterthiner, Thomas and Nessler, Bernhard and Hochreiter, Sepp},
  booktitle={Advances in neural information processing systems},
  volume={30},
  year={2017}
}

@article{mcinnes2018umap,
  title={{UMAP}: Uniform manifold approximation and projection for dimension reduction},
  author={McInnes, Leland and Healy, John and Melville, James},
  journal={arXiv preprint arXiv:1802.03426},
  year={2018}
}

@article{pang2020tju,
  title={{TJU-DHD}: A diverse high-resolution dataset for object detection},
  author={Pang, Yanwei and Cao, Jiale and Li, Yazhao and Xie, Jin and Sun, Hanqing and Gong, Jinfeng},
  journal={IEEE Transactions on Image Processing},
  volume={30},
  pages={207--219},
  year={2020},
  publisher={IEEE}
}

@inproceedings{saharia2022palette,
  title={Palette: Image-to-image diffusion models},
  author={Saharia, Chitwan and Chan, William and Chang, Huiwen and Lee, Chris and Ho, Jonathan and Salimans, Tim and Fleet, David and Norouzi, Mohammad},
  booktitle={ACM SIGGRAPH 2022 conference proceedings},
  pages={1--10},
  year={2022}
}

@article{meng2021sdedit,
  title={Sdedit: Guided image synthesis and editing with stochastic differential equations},
  author={Meng, Chenlin and He, Yutong and Song, Yang and Song, Jiaming and Wu, Jiajun and Zhu, Jun-Yan and Ermon, Stefano},
  journal={arXiv preprint arXiv:2108.01073},
  year={2021}
}

@article{greenberg2024seed,
  title={Seed-to-Seed: Image Translation in Diffusion Seed Space},
  author={Greenberg, Or and Kishon, Eran and Lischinski, Dani},
  journal={arXiv preprint arXiv:2409.00654},
  year={2024}
}

@Article{s24041339,
AUTHOR = {Bang, Geonkyu and Lee, Jinho and Endo, Yuki and Nishimori, Toshiaki and Nakao, Kenta and Kamijo, Shunsuke},
TITLE = {Semantic and Geometric-Aware Day-to-Night Image Translation Network},
JOURNAL = {Sensors},
VOLUME = {24},
YEAR = {2024},
NUMBER = {4},
ARTICLE-NUMBER = {1339},
URL = {https://www.mdpi.com/1424-8220/24/4/1339},
PubMedID = {38400497},
ISSN = {1424-8220},
DOI = {10.3390/s24041339}
}

@inproceedings{li2024light,
  title={Light the night: A multi-condition diffusion framework for unpaired low-light enhancement in autonomous driving},
  author={Li, Jinlong and Li, Baolu and Tu, Zhengzhong and Liu, Xinyu and Guo, Qing and Juefei-Xu, Felix and Xu, Runsheng and Yu, Hongkai},
  booktitle={Proceedings of the IEEE/CVF Conference on Computer Vision and Pattern Recognition},
  pages={15205--15215},
  year={2024}
}

@inproceedings{park2020contrastive,
  title={Contrastive learning for unpaired image-to-image translation},
  author={Park, Taesung and Efros, Alexei A and Zhang, Richard and Zhu, Jun-Yan},
  booktitle={European conference on computer vision},
  pages={319--345},
  year={2020},
  organization={Springer}
}

@inproceedings{zheng2021spatially,
  title={The spatially-correlative loss for various image translation tasks},
  author={Zheng, Chuanxia and Cham, Tat-Jen and Cai, Jianfei},
  booktitle={Proceedings of the IEEE/CVF conference on computer vision and pattern recognition},
  pages={16407--16417},
  year={2021}
}

@inproceedings{wang2021instance,
  title={Instance-wise hard negative example generation for contrastive learning in unpaired image-to-image translation},
  author={Wang, Weilun and Zhou, Wengang and Bao, Jianmin and Chen, Dong and Li, Houqiang},
  booktitle={Proceedings of the IEEE/CVF international conference on computer vision},
  pages={14020--14029},
  year={2021}
}

@inproceedings{lin2022exploring,
  title={Exploring negatives in contrastive learning for unpaired image-to-image translation},
  author={Lin, Yupei and Zhang, Sen and Chen, Tianshui and Lu, Yongyi and Li, Guangping and Shi, Yukai},
  booktitle={Proceedings of the 30th ACM international conference on multimedia},
  pages={1186--1194},
  year={2022}
}

@inproceedings{zhan2022modulated,
  title={Modulated contrast for versatile image synthesis},
  author={Zhan, Fangneng and Zhang, Jiahui and Yu, Yingchen and Wu, Rongliang and Lu, Shijian},
  booktitle={Proceedings of the IEEE/CVF Conference on Computer Vision and Pattern Recognition},
  pages={18280--18290},
  year={2022}
}

@inproceedings{caron2021emerging,
  title={Emerging properties in self-supervised vision transformers},
  author={Caron, Mathilde and Touvron, Hugo and Misra, Ishan and J{\'e}gou, Herv{\'e} and Mairal, Julien and Bojanowski, Piotr and Joulin, Armand},
  booktitle={Proceedings of the IEEE/CVF international conference on computer vision},
  pages={9650--9660},
  year={2021}
}

@misc{yolo26_ultralytics,
  author = {Glenn Jocher and Jing Qiu},
  title = {Ultralytics {YOLO26}},
  version = {26.0.0},
  year = {2026},
  url = {https://github.com/ultralytics/ultralytics},
  orcid = {0000-0001-5950-6979, 0000-0003-3783-7069},
  license = {AGPL-3.0}
}

@inproceedings{sauer2024adversarial,
  title={Adversarial diffusion distillation},
  author={Sauer, Axel and Lorenz, Dominik and Blattmann, Andreas and Rombach, Robin},
  booktitle={European Conference on Computer Vision},
  pages={87--103},
  year={2024},
  organization={Springer}
}

@inproceedings{zhang2017citypersons,
  title={Citypersons: A diverse dataset for pedestrian detection},
  author={Zhang, Shanshan and Benenson, Rodrigo and Schiele, Bernt},
  booktitle={Proceedings of the IEEE conference on computer vision and pattern recognition},
  pages={3213--3221},
  year={2017}
}

@article{lin2021gan,
  author  = {Lin, Che-Tsung and Huang, Sheng-Wei and Wu, Yen-Yi and Lai, Shang-Hong},
  title   = {{GAN}-Based Day-to-Night Image Style Transfer for Nighttime Vehicle Detection},
  journal = {IEEE Transactions on Intelligent Transportation Systems},
  year    = {2021},
  volume  = {22},
  number  = {2},
  pages   = {951--963},
  month   = feb,
  doi     = {10.1109/TITS.2019.2961679},
  issn    = {1524-9050}
}

@misc{yolov8_ultralytics,
  author  = {Glenn Jocher and Ayush Chaurasia and Jing Qiu},
  title   = {Ultralytics {YOLOv8}},
  version = {8.0.0},
  year    = {2023},
  url     = {https://github.com/ultralytics/ultralytics},
  orcid   = {0000-0001-5950-6979, 0000-0002-7603-6750, 0000-0003-3783-7069},
  license = {AGPL-3.0}
}

@article{lee2023daytonight,
  title   = {Day-to-night image translation via transfer learning to keep semantic information for driving simulator},
  author  = {Lee, Jinho and Shiotsuka, Daiki and Bang, Geonkyu and Endo, Yuki and Nishimori, Toshiaki and Nakao, Kenta and Kamijo, Shunsuke},
  journal = {IATSS Research},
  volume  = {47},
  number  = {2},
  pages   = {251--262},
  year    = {2023},
  doi     = {10.1016/j.iatssr.2023.04.001}
}

@article{lin2023cycleobject,
  title   = {Cycle-object consistency for image-to-image domain adaptation},
  author  = {Lin, Che-Tsung and Kew, Jie-Long and Chan, Chee Seng and Lai, Shang-Hong and Zach, Christopher},
  journal = {Pattern Recognition},
  volume  = {138},
  pages   = {109416},
  year    = {2023},
  doi     = {10.1016/j.patcog.2023.109416}
}

@article{cha2025styleprompt,
  title   = {Style prompt tuning for bridging visual gaps in autonomous driving},
  author  = {Cha, Suyeon and Choi, Giyun and Kwak, Minji and Choi, Jongwon},
  journal = {Engineering Applications of Artificial Intelligence},
  volume  = {161},
  pages   = {112105},
  year    = {2025},
  doi     = {
  10.1016/j.engappai.2025.112105}
}

@inproceedings{kim2022instaformer,
  title = {{InstaFormer}: Instance-aware image-to-image translation with transformer},
  author={Kim, Soohyun and Baek, Jongbeom and Park, Jihye and Kim, Gyeongnyeon and Kim, Seungryong},
  booktitle={2022 IEEE/CVF Conference on Computer Vision and Pattern Recognition (CVPR)},
  pages={18300--18310},
  year={2022},
  organization={IEEE}
}

\end{document}